\documentclass[10pt]{article}
\usepackage{graphicx}%
\usepackage{multirow}%
\usepackage{amsmath,amssymb,amsfonts}%
\usepackage{amsthm}%
\usepackage{mathrsfs}%
\usepackage[title]{appendix}%
\usepackage{xcolor}%
\usepackage{textcomp}%
\usepackage{manyfoot}%
\usepackage{booktabs}%
\usepackage{algorithm}%
\usepackage{algorithmicx}%
\usepackage{algpseudocode}%
\usepackage{listings}%
\usepackage{ulem}
\usepackage{url}
\usepackage{multirow}
\usepackage{booktabs}
\usepackage{dsfont}
\usepackage{tikz}
\usepackage{enumitem}
 \usepackage{amsmath}
 \usepackage{amsfonts}

\definecolor{SALMON}{rgb}{0.9725490, 0.4627451, 0.4274510}
\definecolor{TURQUOISE}{rgb}{0.0000000, 0.7490196, 0.7686275}

\newcommand{\Nd}{N_{\text{d}}}
\newcommand{\Bijk}{B_{i,j,k}}
\newcommand{\dij}{\bd_{i,j}}
\newcommand{\epsijk}{\varepsilon_{i,j,k}}
\newcommand{\Rij}{R_{i,j}}
\newcommand{\Rpij}{\Rij^{(p)}}
\newcommand{\Rbij}{\Rij^{(b)}}
\newcommand{\Uijk}{U_{i,j,k}}
\newcommand{\Wij}{W_{i,j}}
\newcommand{\Wpij}{\Wij^{(p)}}
\newcommand{\Wbij}{\Wij^{(b)}}
\newcommand{\Xijk}{X_{i,j,k}}
\newcommand{\Yijk}{Y_{i,j,k}}
\newcommand{\Zijk}{Z_{i,j,k}}
\newcommand{\bbeta}{\boldsymbol{\beta}}
\newcommand{\bB}{\boldsymbol{B}}
\newcommand{\bd}{\boldsymbol{d}}
\newcommand{\beps}{\boldsymbol{\varepsilon}}
\newcommand{\bR}{\boldsymbol{R}}
\newcommand{\bRb}{\bR^{(b)}}
\newcommand{\bRp}{\bR^{(p)}}
\newcommand{\bW}{\boldsymbol{W}}
\newcommand{\bWb}{\bW^{(b)}}
\newcommand{\bWp}{\bW^{(p)}}
\newcommand{\bX}{\boldsymbol{X}}
\newcommand{\bY}{\boldsymbol{Y}}
\newcommand{\bZ}{\boldsymbol{Z}}

\newcommand{\Xtijk}{\widetilde{X}_{i,j,k}}
\newcommand{\Xbarij}{\overline{X}_{i,j}}
\newcommand{\Xtbarij}{\overline{\widetilde{X}}_{i,j}}
\newcommand{\Ybarij}{\overline{Y}_{i,j}}
\newcommand{\bbetah}{\widehat{\bbeta}}
\newcommand{\Rhij}{\widehat{R}_{i,j}}

\newcommand{\Cov}[2]{\text{Cov}\left(#1,#2\right)}
\newcommand{\Esp}[1]{\mathbb{E}\left[#1\right]}
\newcommand{\Var}[1]{\mathbb{V}\left[#1\right]}
\newcommand{\Corre}[2]{\text{Cor}\left(#1,#2\right)}

\newcommand{\mcT}{\mathcal{T}}
\newcommand{\mcNew}{\mathcal{N}_{\text{ew}}}
\newcommand{\setN}{\mathbb{N}}
\newcommand{\mcJL}{\mathcal{J}_L}
\usepackage{threeparttable}
\usepackage{tabularx}
\definecolor{ForestGreen}{rgb}{0.07,0.62,0.34}
\usepackage[authoryear,longnamesfirst]{natbib}

\usepackage{hyperref}
\usepackage{authblk}
\usepackage{placeins}

\hypersetup{
    colorlinks=true,
    linkcolor=black,
    filecolor=magenta,      
    urlcolor=cyan,
    pdftitle={Wang2025study},
    pdfpagemode=FullScreen,
    citecolor=blue,
    breaklinks=true,
    }

\title{Study of the influence of a biased database on the prediction of standard algorithms for selecting the best candidate for an interview}

\author[1]{Shuyu Wang}

\author[1]{Angélique Saillet}

\author[1]{Philomène Le Gall}

\author[2]{Alain Lacroux}

\author[3]{Christelle Martin-Lacroux}

\author[1]{\setcounter{footnote}{1}Vincent Brault\footnote{Corresponding author: \url{vincent.brault@univ-grenoble-alpes.fr}}}

\affil[1]{\setcounter{footnote}{0}Univ. Grenoble Alpes, CNRS, Grenoble INP\footnote{Institute of Engineering Univ. Grenoble Alpes}, LJK, 38000 Grenoble, France}

\affil[2]{Université Paris 1 Panthéon-Sorbonne, PRISM, Paris, France}

\affil[3]{Univ. Grenoble Alpes, Grenoble INP$^{*}$, CERAG, 38000 Grenoble, France}

\begin{document}

\maketitle

\abstract{Artificial intelligence is used at various stages of the recruitment process to automatically select the best candidate for a position, with companies guaranteeing unbiased recruitment. However, the algorithms used are either trained by humans or are based on learning from past experiences that were biased. In this article, we propose to generate data mimicking external (discrimination) and internal biases (self-censorship) in order to train five classic algorithms and to study the extent to which they do or do not find the best candidates according to objective criteria. In addition, we study the influence of the anonymisation of files on the quality of predictions.}




\section{Introduction}

Artificial Intelligence (AI) is extensively used across various stages of the recruitment process, from automated candidate sourcing on social media platforms to asynchronous video recruitment methods. A study of Human Resources (HR) professionals representing 500 mid-sized organisations from diverse industries across five countries revealed that 24\% of businesses have already implemented AI for recruitment purposes, while 56\% of hiring managers plan to adopt it within the next year \citep{sage2020the}. AI is employed to augment human decision-making regarding job candidates (such as determining who should receive a job offer) and to support the actions of human decision-makers throughout the process (such as data collection and analysis; \citet{gonzalez2022allying}).

Some applications incorporating AI algorithms are widely accepted and relatively uncontroversial. For instance, automated CV (curriculum vitae) analysis, which involves the identification and automatic extraction of relevant information from CVs, is commonly used to build pre-recruitment databases with suitable candidates for specific jobs. However, other applications present more ambitious promises and pose greater challenges. This is particularly true for matching solutions based on machine learning algorithms, which analyze job specifications and applicants’ data (sourced from various channels) to identify the most promising matches. These matching solutions are classified as high-risk AI systems under the EU Artificial Intelligence Act. Providers of such high-risk systems are required to ensure their systems achieve appropriate levels of accuracy, robustness, and cybersecurity, while also implementing a quality management system to ensure regulatory compliance.

Providers of AI-based solutions assert that AI technology enhances efficiency while improving both the quality and objectivity of the recruitment process. Efficiency is achieved through time savings, as these systems are capable of processing large volumes of applications in a short period, relieving recruiters from labour-intensive tasks. Providers of algorithmic matching solutions claim that their systems ensure objective decision-making, as they are not influenced by the biases and stereotypes associated with human judgement. Consequently, algorithmic matching is expected to reduce discriminatory bias in the recruitment process.

The promise of non-discrimination is appealing, as recruiters have a legal obligation to ensure fair and unbiased hiring practices. However, discrimination remains a significant and persistent problem across the labour market in nearly all countries where it has been studied \citep{lippens2023state}, with particular prominence in France \citep{challe2024cyclical}. Gender\footnote{Sex generally refers to a set of biological attributes that are associated with physical and physiological features such as chromosomal genotype, hormonal levels, internal and external anatomy. A binary sex categorization (male/female) is usually designated at birth ("sex assigned at birth") and is in most cases based solely on the visible external anatomy of a newborn. In reality, sex categorizations include people who are intersex/have differences of sex development (DSD).}, ethnicity, age, and disability continue to be widespread grounds for discrimination \citep{Rouhban2022dix}. Discrimination can also manifest in more subtle and indirect ways for marginalized groups. One well-known effect is stereotype threat, where members of groups subjected to negative stereotypes may censor themselves in certain applications \citep{ramsoomair2019sources} or under-perform in recruitment tests due to internalizing stereotypes of inferior performance. Stereotype threat effects have been observed across various social groups and stereotypes, including girls in mathematics, science, and technology \citep{cadaret2017stereotype,shapiro2012role}.

At first glance, the use of AI systems to reduce biases seems to present a persuasive commercial and ethical argument for clients, and perhaps even a strategy to attract applicants. However, matching systems rely on supervised learning algorithms (either machine learning or deep learning), which introduces two major concerns: 
\begin{itemize}
    \item The "black box effect," which makes it sometimes difficult to explain the outcome produced by the AI \citep{hildebrandt2010challenges}, persists despite recent advancements in the field of explainable AI (XAI) applied to recruitment \citep{lee2023fat}.
    \item The sensitivity of learning algorithms to the data they are trained on. \citet{favaretto2019big} highlight various algorithmic sources of discrimination, particularly the risk that the training data may be contaminated by past discriminatory cases or that certain categories of candidates may be over- or under-represented. If training datasets are biased, particularly due to the under-representation of specific groups, these biases may influence the matching process.  
\end{itemize}
Evaluating the performance of different learning algorithms used in recruitment is therefore crucial. On one hand, designers of these systems promise objective, bias-free hiring, but on the other hand, documented cases of algorithmic discrimination have emerged, such as Amazon's algorithm, which favoured predominantly male candidates for technical roles \citep{kochling2023better}.

In this article, we aim to assess the extent to which these algorithms can mitigate, sustain, or even exacerbate the biases present in the training datasets. The most straightforward approach to achieving this would be to analyze the algorithms employed by providers of matching solutions. However, this approach is impractical due to both commercial and technical constraints: developers are generally unwilling to disclose the characteristics and performance of their systems, citing industrial secrecy and the need to maintain a competitive advantage in a highly competitive market \citep{hildebrandt2010challenges}.

To overcome this challenge, we propose using simulations. Given that the algorithms used by developers are likely to belong to the classes of algorithms currently identified and in use, this article seeks to test the impact of various types of algorithms on simulated recruitment processes using a biased training database. In particular, we are interested in describing two types of bias (external and self-censorship) and then testing five common algorithms (those that give the best prediction results when we know the perfect ranking). We conclude with a discussion of the results and suggestions for improvements.

\section{Recruitment methods simulation}\label{sec:simulation}

Companies that want to recruit new staff have various selection methods at their disposal. Taking France as an example, the CV is the main selection tool used (in 88\% of recruitments), followed by telephone interviews (in 30\% of recruitments) and knowledge or intelligence tests (in 28\% of recruitments). Shortlisted candidates undergo a recruitment interview in 88\% of cases \citep{bergeat2017comment,remy2024efforts}.

Recruiters base their decisions on objective variables (related to the job requirements) and on their subjective perceptions of the candidate's strengths and weaknesses. 
The objective variables correspond, for example, to the content of the CV related to the job (skills, experience, education), as well as to the candidate's performance observed in technical or cognitive aptitude tests. Subjective perceptions can be based on visible characteristics of the candidate as well as impressions made during the interview. Examples of visible characteristics include gender, physical appearance, and ethnic/racial background. Research has consistently shown that hiring managers' subjective perceptions are often influenced by cognitive biases related to the activation of stereotypes \citep{whysall2018cognitive}. This can lead to discriminatory hiring behaviour, defined as the unequal treatment of an individual on the basis of a legally prohibited criterion (e.g. origin, age, etc.), which can occur during the hiring process. A meta-analysis found that ethnic and racial discrimination remains widespread in the Organization for Economic Cooperation and Development (OECD) countries \citep{zschirnt2016ethnic}.

The presence of stereotypes and discrimination can also have adverse effects on job applicants from stigmatized populations, by undermining their self-esteem. This can lead to self-censorship mechanisms (e.g., applicants dropping out) and a risk of lower performance in some selection tests: a mechanism known in social psychology as "stereotype threat". Stereotype threat can explain situations in which certain applications are unsuccessful despite the candidates' suitability for the job. This bias refers to the ‘difficult situation’ in which members of a social group (e.g. African-Americans, women) face the possibility of being judged or treated in a stereotypical way, or doing something that would confirm the stereotype. For example, when members of a stereotyped group take standardised aptitude tests, as in job selection contexts, their performance may be partially compromised when they encounter evidence of a strong negative stereotype in the test environment (e.g. women are not good at maths or ethnic minorities have lower intellectual ability; \cite{steele2002contending}).

In order to simulate a recruitment process, it is therefore necessary to consider both objective and subjective variables that may lead to discriminatory behaviour. It is also interesting to consider the fact that discrimination can be caused by the recruiter (external censorship), but also by the candidate who engages in self-censoring behaviour.

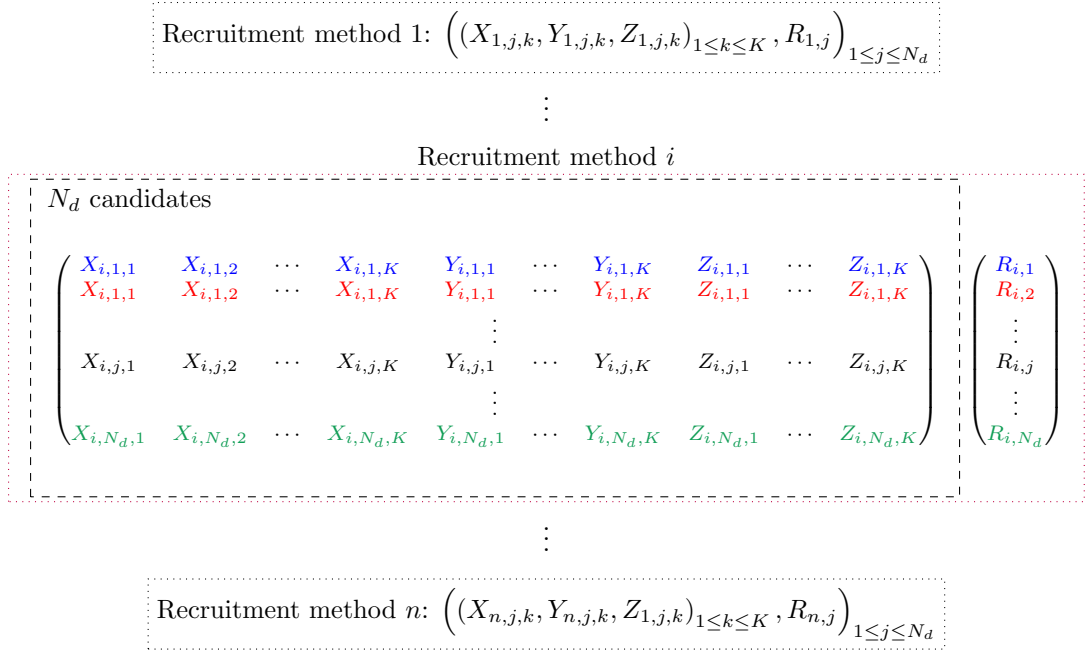
\begin{figure}[!ht]
\begin{center}
\begin{tikzpicture}[scale=1.3]
%%%% 
\node (class) at (-0.775,0.75) {\footnotesize$\begin{pmatrix}
\textcolor{blue}{X_{i,1,1}}&\textcolor{blue}{X_{i,1,2}}&\cdots&\textcolor{blue}{X_{i,1,K}}&\textcolor{blue}{Y_{i,1,1}}&\cdots&\textcolor{blue}{Y_{i,1,K}}&\textcolor{blue}{Z_{i,1,1}}&\cdots&\textcolor{blue}{Z_{i,1,K}}\\
\textcolor{red}{X_{i,1,1}}&\textcolor{red}{X_{i,1,2}}&\cdots&\textcolor{red}{X_{i,1,K}}&\textcolor{red}{Y_{i,1,1}}&\cdots&\textcolor{red}{Y_{i,1,K}}&\textcolor{red}{Z_{i,1,1}}&\cdots&\textcolor{red}{Z_{i,1,K}}\\
\multicolumn{10}{c}{\vdots}\\
X_{i,j,1}&X_{i,j,2}&\cdots&X_{i,j,K}&Y_{i,j,1}&\cdots&Y_{i,j,K}&Z_{i,j,1}&\cdots&Z_{i,j,K}\\
\multicolumn{10}{c}{\vdots}\\
\textcolor{ForestGreen}{X_{i,N_d,1}}&\textcolor{ForestGreen}{X_{i,N_d,2}}&\cdots&\textcolor{ForestGreen}{X_{i,N_d,K}}&\textcolor{ForestGreen}{Y_{i,N_d,1}}&\cdots&\textcolor{ForestGreen}{Y_{i,N_d,K}}&\textcolor{ForestGreen}{Z_{i,N_d,1}}&\cdots&\textcolor{ForestGreen}{Z_{i,N_d,K}}\\
\end{pmatrix}$};

%%% Nd Dossiers
\draw[dashed] (-5.525,-0.75) rectangle (3.975,2.5);
\draw (-5.45,2.5) node[below right]{$N_d$ candidates};
%%% Classement
\node (rank) at (4.55,0.75) {\footnotesize$\begin{pmatrix}
\textcolor{blue}{R_{i,1}}\\ \textcolor{red}{R_{i,2}}\\\vdots\\ R_{i,j}\\\vdots\\\textcolor{ForestGreen}{R_{i,N_d}}\\
\end{pmatrix}$};

%%% Entretien
\draw[dotted,purple] (-5.75,-0.8) rectangle (5.25,2.55);
\draw (-0.25,2.55) node[above]{Recruitment method $i$};

%%% Liste des entretiens
\node[draw,dotted] (inter1) at (-0.25,3.95) {Recruitment method 1: $\left(\left(X_{1,j,k},Y_{1,j,k},Z_{1,j,k}\right)_{1\leq k\leq K},R_{1,j}\right)_{1\leq j\leq N_d}$};
\node[draw,dotted] (intern) at (-0.25,-1.95) {Recruitment method $n$: $\left(\left(X_{n,j,k},Y_{n,j,k},Z_{1,j,k}\right)_{1\leq k\leq K},R_{n,j}\right)_{1\leq j\leq N_d}$};
\draw (-0.25,3.3) node{$\vdots$};
\draw (-0.25,-1.1) node{$\vdots$};
\end{tikzpicture}
\end{center}
    \caption{Schematic representation of the notations for $n$ recruitment methods. Each recruitment method $i$ is composed of $N_d$ profiles (one by row in the central box) and each profile is associated at an answer to know if it has the job or not (center right matrix).}
    \label{fig:notation:schema}
\end{figure}

We assume that $n$ \textbf{recruitment methods} are simulated with $\Nd$ \textbf{profiles} $\dij$ each with $i\in\{1,\ldots,n\}$ and $j\in\left\{1,\ldots,\Nd\right\}$. Each profile is characterized by (see Figure~\ref{fig:notation:schema} for a schematic representation):
\begin{itemize}
	\item $K$ \textbf{objective} variables $\left(\Xijk\right)_{1\leq k\leq K}$,
	\item $K$ \textbf{discriminatory } variables $\left(\Yijk\right)_{1\leq k\leq K}$,
	\item $K$ variables \textbf{correlated} with discriminatory variables $\left(\Zijk\right)_{1\leq k\leq K}$,
	\item the \textbf{rank} $\Rij\in\{1,\ldots,\Nd\}$ within the recruitment method $i$,
	\item the \textbf{success} $\Wij\in\{0,1\}$ of getting the job $i$. It is assumed that only one person will be caught in the end then $\Wij$ is equal to $1$ if and only if $\Rij$ is also equal to $1$.
\end{itemize}

For the following, the variables $\Xijk$ are assumed to be independents and to follow the same Gaussian distribution $\mathcal{N}\left(0,1\right)$. A recruitment method is called \textbf{perfect} if the ranking is only based on the objective variable; in this article, the perfect ranking is made from the means $\Xbarij$. In this case, $\Rpij$ represents the perfect rank of the profile $\dij$ and $\Wpij$ the associated success variable. In particular, it is assumed that there exist a permutation $\sigma_i\in\mathfrak{S}\left(\left\{1,\ldots,\Nd\right\}\right)$ such that:
\[\overline{X}_{i,\sigma_i^{-1}(1)}>\overline{X}_{i,\sigma_i^{-1}(2)}>\cdots>\overline{X}_{i,\sigma_i^{-1}\left(\Nd\right)}\]
and $\Rpij$ is defined by $\Rpij=\sigma_i(j)$. As the laws of $\Xbarij$ are continuous, $\sigma_i$ is almost surely unique for the Lebesgue measure.

Objective variables can represent what is expected in a CV or the results of supposedly objective tests.

For the \textbf{biased} rankings, a first proposal for bias by discrimination is presented, followed by a second by stereotype threat. In the both cases, $\Rbij$ represents the biased ranking and $\Wbij$ the associated success.

\subsection{Recruitment methods biased by external censorship (discrimination)}

Discrimination from a recruiter corresponds to external censorship which can manifest itself in two ways: either by rejection, or by downgrading the candidate. 

To simulate the influence of an external censorship, it is assumed that there exists a \textbf{threshold} $S\in\mathbb{R}$ below which a profile $\dij$ will not be studied. Given a correlation $\alpha\in[0,1[$ two cases are studied:

\begin{itemize}
    \item The \textbf{binary case} assumes that the variables $\Yijk$ are independent, independent of the variables $\Xijk$ and follow the same Bernoulli distribution $\mathcal{B}\left(1/2\right)$. Moreover, for each triplet $(i,j,k)$ $\Zijk$ is generated by:
    \begin{equation}\label{Eq:Z:binary}
        \Zijk=\Uijk\Yijk+(1-\Uijk)\Bijk
    \end{equation}
    with the variables $\Bijk$ are independent, independent of the variables $\Xijk$ and $\Yijk$ and follow the same Bernoulli distribution $\mathcal{B}\left(1/2\right)$ and $\Uijk$ are independent, independent of the variables $\Xijk$, $\Yijk$ and $\Bijk$ and follow the same Bernoulli distribution $\mathcal{B}\left(\alpha\right)$.
    \item The \textbf{continuous case} assumes that the variables $\Yijk$ are independent, independent of the variables $\Xijk$ and follow the same Gaussian distribution $\mathcal{N}\left(0,1\right)$. Moreover, for each triplet $(i,j,k)$ $\Zijk$ is generated by:
    \begin{equation}\label{Eq:Z:continuous}
        \Zijk=\frac{\alpha}{\sqrt{1-\alpha^2}}\Yijk+\epsijk
    \end{equation}
    with the variables $\epsijk$ are independent, independent of the variables $\Xijk$ and $\Yijk$ and follow the same Bernoulli distribution $\mathcal{N}\left(0,1\right)$.
\end{itemize}

In the both cases~\eqref{Eq:Z:binary} and~\eqref{Eq:Z:continuous}, for each triplet $(i,j,k)$, the correlation between $\Yijk$ and $\Zijk$ is $\alpha$ (see Supplementary Information).\\

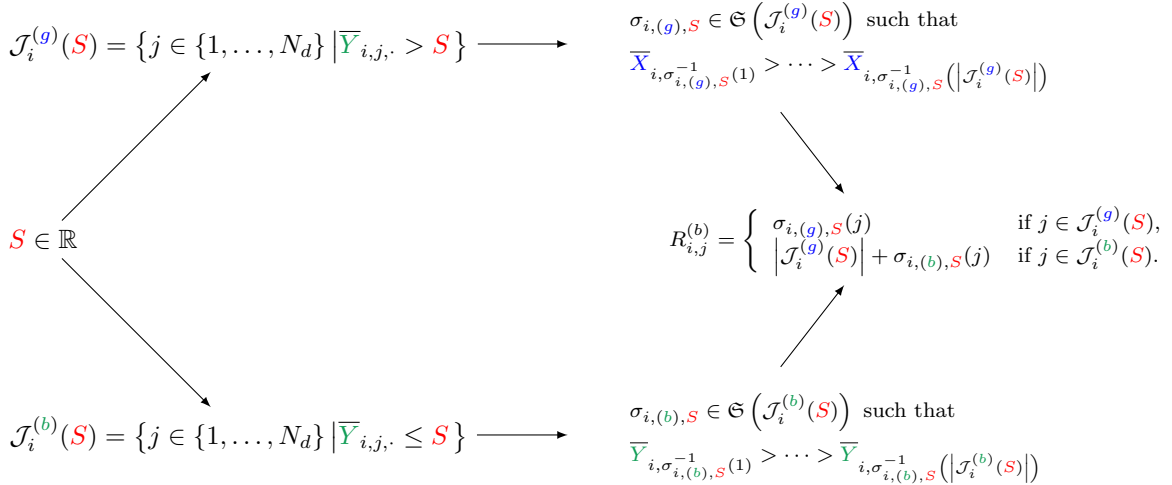
\begin{figure}[!ht]
    \centering
    \begin{tikzpicture}[scale=1.3]
\node (S) at (-2,0) {$\textcolor{red}{S}\in\mathbb{R}$};

\node (pos) at (0,2) {$\mathcal{J}_i^{(\textcolor{blue}{g})}(\textcolor{red}{S})=\left\{j\in\{1,\ldots,N_d\}\left|\overline{\textcolor{ForestGreen}{Y}}_{i,j,\cdot}> \textcolor{red}{S}\right.\right\}$};
\draw[->,>=latex] (S) -- (pos);

\node (neg) at (0,-2) {$\mathcal{J}_i^{(\textcolor{ForestGreen}{b})}(\textcolor{red}{S})=\left\{j\in\{1,\ldots,N_d\}\left|\overline{\textcolor{ForestGreen}{Y}}_{i,j,\cdot}\leq \textcolor{red}{S}\right.\right\}$};
\draw[->,>=latex] (S) -- (neg);

\node (sig_pos) at (5,2) {\begin{minipage}{0.25\textwidth}\footnotesize
\begin{eqnarray*}
&&\sigma_{i,(\textcolor{blue}{g}),\textcolor{red}{S}}\in\mathfrak{S}\left(\mathcal{J}_i^{(\textcolor{blue}{g})}(\textcolor{red}{S})\right)\text{ such that}\\
&&\overline{\textcolor{blue}{X}}_{i,\sigma_{i,(\textcolor{blue}{g}),\textcolor{red}{S}}^{-1}(1)}>\cdots>\overline{\textcolor{blue}{X}}_{i,\sigma_{i,(\textcolor{blue}{g}),\textcolor{red}{S}}^{-1}\left(\left|\mathcal{J}_i^{(\textcolor{blue}{g})}(\textcolor{red}{S})\right|\right)}\\
\end{eqnarray*}
\end{minipage}};
\draw[->,>=latex] (pos) -- (sig_pos);

\node (rank) at (7,0) {\begin{minipage}{0.5\textwidth}
\footnotesize
\[R_{i,j}^{(b)}=\left\{\begin{array}{ll}
        \sigma_{i,(\textcolor{blue}{g}),\textcolor{red}{S}}(j)&\text{if }j\in  \mathcal{J}_i^{(\textcolor{blue}{g})}(\textcolor{red}{S}),\\
        \left|\mathcal{J}_i^{(\textcolor{blue}{g})}(\textcolor{red}{S})\right|+ \sigma_{i,(\textcolor{ForestGreen}{b}),\textcolor{red}{S}}(j)&\text{if }j\in  \mathcal{J}_i^{(\textcolor{ForestGreen}{b})}(\textcolor{red}{S}).\\ 
    \end{array}\right.\]
\end{minipage}};
\draw[->,>=latex] (sig_pos) -- (rank.150);

\node (sig_neg) at (5,-2) {\begin{minipage}{0.25\textwidth}\footnotesize
\begin{eqnarray*}
&&\sigma_{i,(\textcolor{ForestGreen}{b}),\textcolor{red}{S}}\in\mathfrak{S}\left(\mathcal{J}_i^{(\textcolor{ForestGreen}{b})}(\textcolor{red}{S})\right)\text{ such that}\\
&&\overline{\textcolor{ForestGreen}{Y}}_{i,\sigma_{i,(\textcolor{ForestGreen}{b}),\textcolor{red}{S}}^{-1}(1)}>\cdots>\overline{\textcolor{ForestGreen}{Y}}_{i,\sigma_{i,(\textcolor{ForestGreen}{b}),\textcolor{red}{S}}^{-1}\left(\left|\mathcal{J}_i^{(\textcolor{ForestGreen}{b})}(\textcolor{red}{S})\right|\right)}\\
\end{eqnarray*}
\end{minipage}};
\draw[->,>=latex] (neg) -- (sig_neg);

\draw[->,>=latex] (sig_neg) -- (rank.210);
\end{tikzpicture}
    \caption{Schematic representation of the biased ranking. Starting from the threshold $S$, profiles are separated into two categories: discriminated (bottom) and non-discriminated (top). The \textit{good} profiles and variables $\Xbarij$ are used to start the ranking, while the \textit{bad} profiles and variables $\Ybarij$ are used to end the ranking.}
    \label{fig:biasedrankingschema}
\end{figure}

Given a threshold $S\in\mathbb{R}$, for each case and each recruitment method $i$, the biased ranking is defined as follows (see Figure~\ref{fig:biasedrankingschema} for a schematic representation):
\begin{enumerate}
    \item the profiles are divided into two groups: those whose average $\Ybarij$ exceeds the $S$ threshold (and who will be studied), i.e. the set denoted $\mathcal{J}_i^{(g)}(S)=\left\{j\in\{1,\ldots,\Nd\}\left|\Ybarij> S\right.\right\}$; and the complementary group\linebreak$\mathcal{J}_i^{(b)}(S)=\left\{j\in\{1,\ldots,\Nd\}\left|\Ybarij\leq S\right.\right\}$, the set of profiles that won't be studied.
    \item the biased ranking begins by the profiles in the \textit{good} set $\mathcal{J}_i^{(g)}(S)$ following the objective variables $\Xbarij$ as the perfect ranking. In particular, it is assumed that there exists a permutation $\sigma_{i,(g),S}\in\mathfrak{S}\left(\mathcal{J}_i^{(g)}(S)\right)$ such that:
    \[\overline{X}_{i,\sigma_{i,(g),S}^{-1}(1)}>\overline{X}_{i,\sigma_{i,(g),S}^{-1}(2)}>\cdots>\overline{X}_{i,\sigma_{i,(g),S}^{-1}\left(\left|\mathcal{J}_i^{(g)}(S)\right|\right)}\]
    \item the ending of the biased ranking is made up by the profiles in the \textit{bad} set $\mathcal{J}_i^{(g)}(S)$ following the means $\Ybarij$ then it is assumed that there exists a permutation $\sigma_{i,(b),S}\in\mathfrak{S}\left(\mathcal{J}_i^{(b)}(S)\right)$ such that:
    \[\overline{Y}_{i,\sigma_{i,(b),S}^{-1}(1)}\geq\overline{Y}_{i,\sigma_{i,(b),S}^{-1}(2)}\geq\cdots\geq\overline{Y}_{i,\sigma_{i,(b),S}^{-1}\left(\left|\mathcal{J}_i^{(b)}(S)\right|\right)}.\]
    In the event of a equality (only possible in the binary case), the order is chosen randomly.
    \item Finally, the ranking is defined by:
    \begin{equation}\label{eq:rankingbiased}
        \Rbij=\left\{\begin{array}{ll}
        \sigma_{i,(g),S}(j) &\text{if }j\in  \mathcal{J}_i^{(g)}(S),\\
        \left|\mathcal{J}_i^{(g)}(S)\right|+ \sigma_{i,(b),S}(j)&\text{if }j\in  \mathcal{J}_i^{(b)}(S).\\ 
    \end{array}\right.
    \end{equation}
\end{enumerate}

Thus, it is assumed that non-discriminated profiles are ranked according to their skills, while discriminated profiles are ranked according to their levels of discrimination.

\begin{table}[!ht]
\begin{center}
    \caption{Example of perfect rankings ($R^{(p)}_{i,j}$) and biased rankings ($R^{(b)}_{i,j}$) according to different thresholds. The blue numbers correspond to rankings that are improved by censorship, and the red ones to those that are downgraded. The black rankings correspond to those that remain unchanged by the previous censorship.}
    \label{tab:example:threshold}
    %\begin{threeparttable}
        \begin{tabularx}{\linewidth}{>{\centering\arraybackslash}X>{\centering\arraybackslash}X>{\centering\arraybackslash}X||>{\centering\arraybackslash}X>{\centering\arraybackslash}X>{\centering\arraybackslash}X>{\centering\arraybackslash}X}
    %    \toprule
        \multicolumn{3}{c||}{Values}&\multicolumn{4}{c}{$R^{(b)}_{i,j}$ according to threshold S}\\
        $\overline{X}_{i,j,\cdot}$&$\overline{Y}_{i,j,\cdot}$& $R^{(p)}_{i,j}$&S=-0.5&S=0&S=0.5&S=1\\\hline\hline
    %    \midrule
	       -0.15&0.33&3&\textcolor{blue}{2}&2&2&2\\
            0.36& -0.52&1&\textcolor{red}{5}&5&5&5\\
            -1.10&0.56&5&\textcolor{blue}{4}&\textcolor{blue}{3}&\textcolor{blue}{1}&1\\
            0.28&0.32&2&\textcolor{blue}{1}&1&\textcolor{red}{3}&3\\
            -0.48&-0.32&4&\textcolor{blue}{3}&\textcolor{red}{4}&4&4\\
    %    \bottomrule
        \end{tabularx}
    %    \begin{tablenotes}
            \item $\overline{X}_{i,j,\cdot}$ objective variables means, $\overline{Y}_{i,j,\cdot}$ discriminatory variables means, $R^{(p)}_{i,j}$ perfect ranks and $R^{(b)}_{i,j}$.
    %    \end{tablenotes}
    %\end{threeparttable}
\end{center}
\end{table}

The choice of $\alpha_i$ parameters is complicated. In organisational psychology, a correlation of 0.2 is fairly common, a correlation of 0.4 is beginning to be noticeable, a correlation of 0.6 is considered strong and a correlation of 0.8 is really very rare (more often than not these are poorly discriminated variables). For example, in the article \cite{lovakov2021empirically} less than 5\% of the correlations found are greater than 0.67. In the following, we will study $\alpha$ correlations of $0.2$, $0.5$ and $0.8$.

\subsection{Recruitment methods biased by self-censorship}

One way of avoiding recruiter bias is to propose tests that are judged to be \textit{objective}, and to select the best candidate on the results of  cognitive ability tests or specific abilities tests \citep{nye2020more}. However, it appears that bias can still occur depending on how the tests were presented. This type of internal bias is called stereotype threat. For example, in their study, \cite{steele2006math} show that women could experience stress related to the activation of a gendered depreciation stereotype when an aptitude test mentioned that it was a math test (compared to an identical test without this mention). To mimic this type of censoring in learning data, we again take the binary case~\eqref{Eq:Z:binary} and propose to consider a parameter $\mu>0$ of \textbf{depreciation} and to calculate for any triplet $(i,j,k)$ the depreciated $\Xtijk$ variables as follows:
\[\Xtijk=\Xijk-\mu\left(1-\Yijk\right).\]
In this case, the biased ranking repeats the perfect ranking technique of section~\ref{sec:simulation} but with the $\Xtbarij$ variables.

\subsection{Anonymising files}

Another way to avoid discrimination by recruiters (external censorship) is to remove personal information from the application file that is unrelated to the job requirements and likely to be used for discrimination: this mainly concerns name, photo, and sociodemographic information \citep{foley2018does}.  

Finally, for the training bases, we provide the \textbf{full databases}, i.e. with the three variables $\left(\bX,\bY,\bZ\right)$ and the \textbf{anonymized databases} with only $\left(\bX,\bZ\right)$. In addition, we compare the algorithms having learned on perfect databases (i.e. with $\bRp$ or $\bWp$) and biased databases ($\bRb$ or $\bWb$). In this way, the four types of training data supplied to the algorithms can be summarized in table~\ref{tab:résumé_config}.
\begin{table}[ht]
    \centering
    \caption{Summary table of the six configurations for algorithm training bases.}
    \label{tab:résumé_config}
    \begin{tabular}{c||c|c|c}
    \multicolumn{4}{c}{}\\
    \multicolumn{1}{c}{}&\multicolumn{3}{c}{Rank}\\\cmidrule{2-4}
    Databases&Perfect&Biased by discrimination&Biased self-censorship\\\cmidrule{2-4}\morecmidrules\cmidrule{2-4}
    Full&$\left(\bX,\bY,\bZ,\bRp,\bWp\right)$&$\left(\bX,\bY,\bZ,\bRb,\bWb\right)$&$\left(\widetilde{\bX},\bY,\bZ,\bRb,\bWb\right)$\\\hline
    Anonymous&$\left(\bX,\bZ,\bRp,\bWp\right)$&$\left(\bX,\bZ,\bRb,\bWb\right)$&$\left(\widetilde{\bX},\bZ,\bRb,\bWb\right)$\\
    \end{tabular}
\end{table}

To assess the quality of the procedures, we simulate $m$ recruitment methods $i\in\{n+1,\ldots,n+m\}$ of $\Nd$ files equally and provide the values $\left(\bX,\bY,\bZ\right)$ and $\left(\bX,\bZ\right)$ depending on whether the files are complete or anonymous and estimate the quality of the results on perfect rankings.

\section{Studied algorithms}

In this section, we present the five algorithms studied in this project:
\begin{itemize}[label=$\bullet$]
    \item Logistic regression (see \cite{berkson1944application}),
    \item Logistic regression with variable selection using the \textit{AIC} criterion (\textit{Akaike Information Criterion}; see \cite{akaike1973information}),
    \item $L$-nearest neighbors (see \cite{fix1951discriminatory,cover1967nearest}),
    \item Multilayer perceptron (see \cite{rosenblatt1957perceptron}),
    \item Support Vector Machine (or \textit{SVM}; see \cite{cortes1995support}).
\end{itemize}
We chose to study only these five algorithms for several reasons:
\begin{enumerate}
    \item as will be shown later, when perfect rankings are used for learning, these methods give good ranking rates in excess of 80\%, unlike, for example, the \textit{CART} (\textit{Classification And Regression Trees}) algorithm from \citet{breiman1984classification}.
    \item methods vary in complexity and use, and are used by different scientific communities.
\end{enumerate}

In the remainder of this section, a few notions for each of the methods are recalled. For learning purposes, it is assumed that the methods only have access to the fact that the files have had the job or not (and not their ranking). In addition, when hyperparameters have to be calibrated, a study is proposed to explain the choices made for the final comparison.

\subsection{Logistic regression}\label{sec:reglogistique}

Logistic regression assumes that the variable $W$ results from a Bernoulli distribution with probability dependent on the variables $(\bX,\bY,\bZ)$. In particular, it assumes that there exists a vector $\beta\in\mathbb{R}^{3K+1}$ such that:
\begin{eqnarray*}
    \mathbb{P}\left(W=1\left|\bX,\bY,\bZ;\bbeta\right.\right)&=&\mathbb{P}\left(W=1\left|X_1,\ldots,X_K,Y_1,\ldots,Y_K,Z_1,\ldots,Z_K;\bbeta\right.\right)\\
    &=&\Phi\left[\beta_0+\sum_{k=1}^K\left(\beta_kX_k+\beta_{K+k}Y_k+\beta_{2K+k}Z_k\right)\right]
\end{eqnarray*}
where $\Phi$ is the logistic function defined for all $x\in\mathbb{R}$ by:
\[\Phi\left(x\right)=\frac{1}{1+e^{-x}}=\frac{e^x}{1+e^{x}}.\]
Thus, given an estimate $\bbetah$ of the vector $\bbeta$ and new recruitment methods $i\in\{n+1,\ldots,n+m\}$, the probability of the file $\dij$ having the recruitment method can be calculated and files can be ranked from highest to lowest probability to have the interview. We assume there is a permutation $\sigma_i\in\mathfrak{S}\left(\left\{1,\ldots,\Nd\right\}\right)$ such that:
\[\mathbb{P}\left(W=1\left|\bd_{i,\sigma_i^{-1}(1)};\bbetah\right.\right)>\mathbb{P}\left(W=1\left|\bd_{i,\sigma_i^{-1}(2)};\bbetah\right.\right)>\cdots>\mathbb{P}\left(W=1\left|\bd_{i,\sigma_i^{-1}(\Nd)};\bbetah\right.\right)\]
with $\Rhij=\sigma_i(j)$. For the estimation, the function \texttt{glm} in the package \texttt{stats} of \textbf{R} software (see~\cite{r2022r}) is used.

\subsection{Logistic regression and the \textit{AIC} criterion}\label{sec:AIC}

Taking all the variables as in section~\ref{sec:reglogistique} can lead to over-learning, and it is therefore advisable to use a variable selection method. As this is a prediction problem, the \textit{AIC} criterion was chosen. The aim is to select the variables that maximise the log-likelihood minus the number of parameters considered to be non-zero~:
\[\bbetah_{\text{AIC}}\in\underset{\bbeta\in\mathbb{R}^{3K+1}}{\arg\!\max}\left\{\mathcal{L}\left(\bbeta\right)-\sum_{k=0}^{3K+1}\mathds{1}_{\{\beta_k\neq0\}}\right\}\]
where $\mathcal{L}\left(\bbeta\right)$ is the loglikelihood and $\mathds{1}$ is the indicator function.

As this represents $2^{3K+1}$ possible models to compare, it quickly becomes time-consuming to test all the models ($65536$ for $K=5$) or even impossible to achieve in less than a year for the supercomputer \textit{Summit}\footnote{148.6 petaflops~; assuming it processes one model per operation} as soon as $K$ is greater than 27. There are several ways of getting around this time problem:
\begin{itemize}[label=$\bullet$]
    \item the \textit{forward} method, which starts with the null vector and progressively allows the coordinates of the $\bbeta$ vector to be non-zero, each time selecting the one that increases the \textit{AIC} criterion the most until it stabilizes;
    \item on the other hand, the \textit{backward} method, which starts with the complete vector and progressively sets to zero the coordinates of the $\bbeta$ vector, each time selecting the one that increases the \textit{AIC} criterion the most until the latter stabilizes.
\end{itemize}
For the procedure tested, the combination of the two implemented in the \texttt{stepAIC} function of the \texttt{MASS} package of \texttt{R} (see \citet{venables2002modern}) was used.

\subsection{$L$-nearest neighbors}\label{sec:Lnn}

Given a train set $\mcT$ of candidates which $\bW$ is known, a set $\mcNew$ of $\Nd$ new candidates and an integer $L\in\setN^\star$, the $L$-nearest neighbor principle is, for each candidate $j\in\mcNew$, to calculate the distance $d\left(j,j'\right)$ at each candidate $j'\in\mcT$ (for example, with the euclidean distance between $\left(X_{j,k},Y_{j,k},Z_{j,k}\right)_{1\leq k\leq K}$ and $\left(X_{j',k},Y_{j',k},Z_{j',k}\right)_{1\leq k\leq K}$) and to find the $L$-nearest candidates $\mcJL(j)$:
\[\mcJL(j)=\left\{j'\in\mcT\left|R'_{j'}(j)\leq L\right.\right\}\]
where $R'_{j'}(j)$ is the rank of the candidate $j'$ following the distances $d\left(j,j'\right)$ in ascending order. At the end, the score of the candidate $j$ is obtain by the mean of the values $\left(W_{j'}\right)_{j'\in\mcJL}$:
\[\frac{1}{L}\sum_{j'\in\mcJL}W_{j'}\]
and the prediction of the best candidate is the one with the highest score (in the event of a tie, the choice is random among the best scores).

For this method, the choice of $L$ is decisive and there is no reason that this choice should be the same for all datasets in all configurations. For this reason, the method of the $10$-fold is used (see \cite{arlot2010survey}). For the choice of the range of possible $L$, an initial study in the perfect case has been carried out (see Appendix B). In the end, it was decided to test all values between $1$ and $70$.

\subsection{Multilayer perceptron}

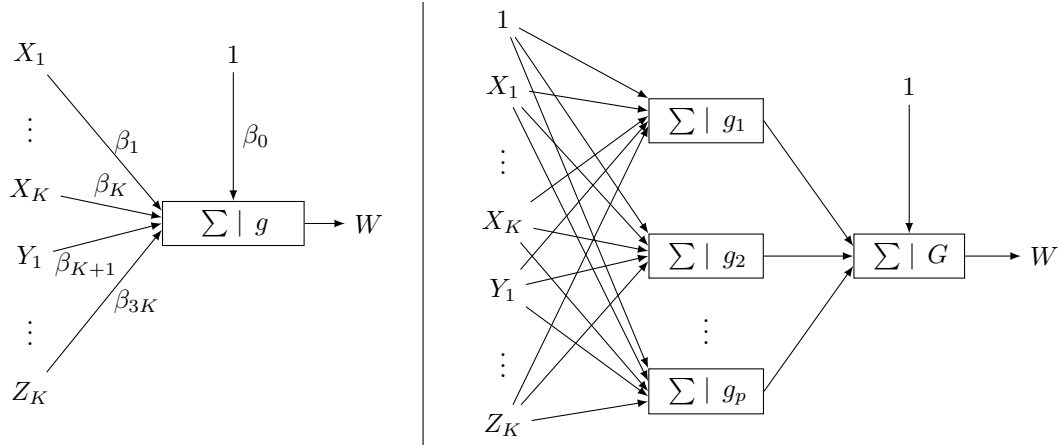
\begin{figure}[ht]
    \centering
    \begin{tabular}{c|c}
        \begin{minipage}[c]{0.35\textwidth}
            \begin{center}
            \begin{tikzpicture}[scale=0.9]
                \node (x1) at (0,5) {$X_1$};
                \node (xj) at (0,4) {$\vdots$};
                \node (xK) at (0,3) {$X_{K}$};
                \node (y1) at (0,2) {$Y_1$};
                \node (yj) at (0,1) {$\vdots$};
                \node (zK) at (0,0) {$Z_{K}$};
                \node (1) at (3,5) {$1$};

                \node[draw] (bloc) at (3,2.5) {$\quad\left.\sum\;\right|\; g\quad$};
                \draw[->,>=latex] (x1) -- (bloc.170) node[midway,right]{$\beta_1$};
                \draw[->,>=latex] (xK) -- (bloc.175) node[midway,above]{$\beta_K$};
                \draw[->,>=latex] (y1) -- (bloc.west) node[pos=0.3,below]{$\beta_{K+1}$};
                \draw[->,>=latex] (zK) -- (bloc.185) node[midway,right]{$\beta_{3K}$};
                \draw[->,>=latex] (1) -- (bloc.north) node[midway,right]{$\beta_{0}$};

                \node (W) at (5,2.5) {$W$};
                \draw[->,>=latex] (bloc.east) -- (W.west);
            \end{tikzpicture}
            \end{center}
        \end{minipage}&
        \begin{minipage}[c]{0.55\textwidth}
            \begin{center}
            \begin{tikzpicture}[scale=0.9]
                \node (1) at (0,6) {$1$};
                \node (x1) at (0,5) {$X_1$};
                \node (xj) at (0,4) {$\vdots$};
                \node (xK) at (0,3) {$X_{K}$};
                \node (y1) at (0,2) {$Y_1$};
                \node (yj) at (0,1) {$\vdots$};
                \node (zK) at (0,0) {$Z_{K}$};
                %%% Bloc 1
                \node[draw] (bloc1) at (3,4.5) {$\;\left.\sum\;\right|\; g_1\;$};
                \draw[->,>=latex] (1) -- (bloc1.north west);
                \draw[->,>=latex] (x1) -- (bloc1.170);
                \draw[->,>=latex] (xK) -- (bloc1.175);
                \draw[->,>=latex] (y1) -- (bloc1.west);
                \draw[->,>=latex] (zK) -- (bloc1.185);
                %%% Bloc 2
                \node[draw] (bloc2) at (3,2.5) {$\;\left.\sum\;\right|\; g_2\;$};
                \draw[->,>=latex] (1) -- (bloc2.north west);
                \draw[->,>=latex] (x1) -- (bloc2.170);
                \draw[->,>=latex] (xK) -- (bloc2.175);
                \draw[->,>=latex] (y1) -- (bloc2.west);
                \draw[->,>=latex] (zK) -- (bloc2.185);
                %%% Bloc p
                \node (p3) at (3,1.5) {$\vdots$};
                \node[draw] (bloc3) at (3,0.5) {$\;\left.\sum\;\right|\; g_p\;$};
                \draw[->,>=latex] (1) -- (bloc3.north west);
                \draw[->,>=latex] (x1) -- (bloc3.170);
                \draw[->,>=latex] (xK) -- (bloc3.west);
                \draw[->,>=latex] (y1) -- (bloc3.185);
                \draw[->,>=latex] (zK) -- (bloc3.190);
                %%% Bloc 4
                \node (1bis) at (6,5) {$1$};
                \node[draw] (bloc4) at (6,2.5) {$\;\left.\sum\;\right|\; G\;$};
                \draw[->,>=latex] (1bis) -- (bloc4.north);
                \draw[->,>=latex] (bloc1.east) -- (bloc4.170);
                \draw[->,>=latex] (bloc2.east) -- (bloc4.west);
                \draw[->,>=latex] (bloc3.east) -- (bloc4.190);
                %%% Output
                \node (W) at (8,2.5)  {$W$};
                \draw[->,>=latex] (bloc4.east) -- (W.west);

            \end{tikzpicture}
            \end{center}
        \end{minipage}
    \end{tabular}
    \caption{Schematic representation of a neuron (left) and a neural network with a hidden layer containing $p$ neurons (right).}
    \label{fig:perceptron}
\end{figure}

The following procedure is a single-layer neural network. The principle of a neural network is to assume that there is a function $g$ and parameters $\beta\in\mathbb{R}^{3K+1}$ such that
\[W=g\left[\beta_0+\sum_{k=1}^K\left(\beta_kX_k+\beta_{K+k}Y_k+\beta_{2K+k}Z_k\right)\right].\]
In particular, if $g=\Phi$ the logistic function, we find the logistic model from section~\ref{sec:reglogistique} (see a schematic on the left-hand side of figure~\ref{fig:perceptron}).

A multi-layer perceptron is the combination of several layers made up of several neurons: each layer contains a certain number of neurons taking as input either initial data or transformations from previous layers. In the diagram on the right of figure~\ref{fig:perceptron}, we've represented the case where there's just one hidden layer containing $p$ neurons. If each function $g_k$ is the function worth $0$ or the input $k$ ($X$, $Y$ or $Z$) and $G=\Phi$ is the logistic function, then we find the objective of the procedure in section~\ref{sec:AIC}.

Given the number of possibilities, and with a view to generalizing previous methods, the model studied was that of the single-layer perceptron with logistic functions (the function \texttt{nnet} from the package of the same name proposed by \citet{venables2002modern} of the \texttt{R} software was used). In this context, it is necessary to choose the number of nodes and the regularization parameter (\texttt{decay} in the function). An experimental design has been proposed in hardware supplementary to calibrate a range of values using the \texttt{tune.nnet} function in the \texttt{e1071} package (see \citet{meyer2023e1071}).

For the choice of the number of nodes and the parameter \texttt{decay}, an initial study in the perfect case has been carried out (see Appendix C). In the end, it was decided to test all number of nodes between $1$ and $10$ and a parameter \texttt{decay} equals to 0.

\subsection{Support Vector Machine}

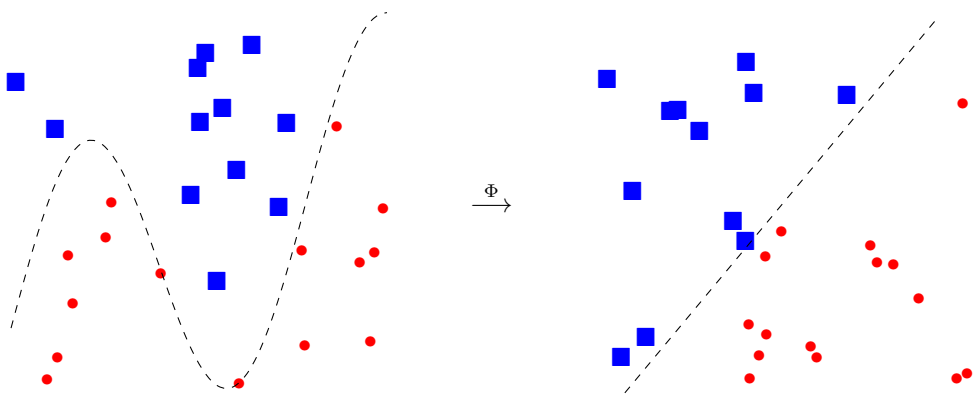
\begin{figure}[!ht]
    \centering
    \begin{tabular}{ccc}
         \begin{minipage}{0.4\linewidth}
             \begin{center}
                 \begin{tikzpicture}[scale=5]
\draw (0.866158169461414,0.210997556477277) node{$\textcolor{red}{\bullet}$};
\draw (0.251578350085765,-0.0851121918284758) node{$\textcolor{red}{\bullet}$};
\draw (0.772582617588341,-0.121014168344946) node{$\textcolor{red}{\bullet}$};
\draw (0.122940408764407,-0.404123518707739) node{$\textcolor{red}{\bullet}$};
\draw (0.781065883580595,-0.373733643068509) node{$\textcolor{red}{\bullet}$};
\draw (0.955462690442801,-0.361447427513586) node{$\textcolor{red}{\bullet}$};
\draw (0.0956014848779887,-0.463497958936305) node{$\textcolor{red}{\bullet}$};
\draw (0.966130643617362,-0.125022311553867) node{$\textcolor{red}{\bullet}$};
\draw (0.398261459777132,-0.18072826005383) node{$\textcolor{red}{\bullet}$};
\draw (0.606519517954439,-0.473564657707275) node{$\textcolor{red}{\bullet}$};
\draw (0.163744637276977,-0.262107541876422) node{$\textcolor{red}{\bullet}$};
\draw (0.927604985656217,-0.153276446367191) node{$\textcolor{red}{\bullet}$};
\draw (0.151547014480457,-0.132648264848204) node{$\textcolor{red}{\bullet}$};
\draw (0.989422131096944,-0.0087401737445457) node{$\textcolor{red}{\bullet}$};
\draw (0.267216322710738,0.00624237107248469) node{$\textcolor{red}{\bullet}$};
\draw (0.732246579602361,0.219825534989549) node{$\textcolor{blue}{\blacksquare}$};
\draw (0.546987550333142,-0.200442151956768) node{$\textcolor{blue}{\blacksquare}$};
\draw (0.711836699163541,-0.00384866432151337) node{$\textcolor{blue}{\blacksquare}$};
\draw (0.516574223292992,0.406165024639531) node{$\textcolor{blue}{\blacksquare}$};
\draw (0.502617597114295,0.223244485353543) node{$\textcolor{blue}{\blacksquare}$};
\draw (0.56252154847607,0.259534074833109) node{$\textcolor{blue}{\blacksquare}$};
\draw (0.496236823732033,0.366841734760864) node{$\textcolor{blue}{\blacksquare}$};
\draw (0.599398671183735,0.0964766637435691) node{$\textcolor{blue}{\blacksquare}$};
\draw (0.0129468676168472,0.329907132453709) node{$\textcolor{blue}{\blacksquare}$};
\draw (0.63985182903707,0.42798149879226) node{$\textcolor{blue}{\blacksquare}$};
\draw (0.117264235392213,0.203988688207655) node{$\textcolor{blue}{\blacksquare}$};
\draw (0.477749483892694,0.0282345739206808) node{$\textcolor{blue}{\blacksquare}$};
\draw[dashed] plot[domain=0:1,samples=50,smooth] (\x,{(-0.2+-0.589911251939131+1.05*\x+sin(8*\x r))/2.42926949856251});
                 \end{tikzpicture}
             \end{center}
         \end{minipage}&
         $\overset{\Phi}{\longrightarrow}$&
         \begin{minipage}{0.4\linewidth}
             \begin{center}
                 \begin{tikzpicture}[scale=5]
\draw (0.864428689470515,0.248827275587246) node{$\textcolor{red}{\bullet}$};
\draw (0.753866600804031,0.344338626135141) node{$\textcolor{red}{\bullet}$};
\draw (0.103293992346153,0.536747969221324) node{$\textcolor{blue}{\blacksquare}$};
\draw (0.457427569432184,0.360918447840959) node{$\textcolor{red}{\bullet}$};
\draw (0.965505853062496,0.0356355728581548) node{$\textcolor{red}{\bullet}$};
\draw (0.673757296754047,0.793690179474652) node{$\textcolor{blue}{\blacksquare}$};
\draw (0.459587272955105,0.153047098079696) node{$\textcolor{red}{\bullet}$};
\draw (0.415222968673334,0.0358382104896009) node{$\textcolor{red}{\bullet}$};
\draw (0.736211410723627,0.391340614529327) node{$\textcolor{red}{\bullet}$};
\draw (0.139318131143227,0.14731619367376) node{$\textcolor{blue}{\blacksquare}$};
\draw (0.403825178975239,0.402713200310245) node{$\textcolor{blue}{\blacksquare}$};
\draw (0.223787900526077,0.75113808689639) node{$\textcolor{blue}{\blacksquare}$};
\draw (0.0358489686623216,0.834802182624117) node{$\textcolor{blue}{\blacksquare}$};
\draw (0.0735068782232702,0.094307190971449) node{$\textcolor{blue}{\blacksquare}$};
\draw (0.797147535718977,0.340391848469153) node{$\textcolor{red}{\bullet}$};
\draw (0.281495467061177,0.696612007683143) node{$\textcolor{blue}{\blacksquare}$};
\draw (0.405708539299667,0.878904439043254) node{$\textcolor{blue}{\blacksquare}$};
\draw (0.425954640144482,0.798800712917) node{$\textcolor{blue}{\blacksquare}$};
\draw (0.413104725070298,0.179787595756352) node{$\textcolor{red}{\bullet}$};
\draw (0.203375915298238,0.749600438168272) node{$\textcolor{blue}{\blacksquare}$};
\draw (0.992976029869169,0.0487192748114467) node{$\textcolor{red}{\bullet}$};
\draw (0.370794901391491,0.456629368942231) node{$\textcolor{blue}{\blacksquare}$};
\draw (0.440226106438786,0.0974791666958481) node{$\textcolor{red}{\bullet}$};
\draw (0.577677784720436,0.122010142775252) node{$\textcolor{red}{\bullet}$};
\draw (0.593819778878242,0.0929843881167471) node{$\textcolor{red}{\bullet}$};
\draw (0.981573664583266,0.767224124167114) node{$\textcolor{red}{\bullet}$};
\draw (0.499858013587072,0.427582624135539) node{$\textcolor{red}{\bullet}$};
\draw[dashed] (0.0833333333333333,0) -- (0.916666666666667,1);
                 \end{tikzpicture}
             \end{center}
         \end{minipage}  \\
    \end{tabular}
    \caption{Schematic representation of a scatter plot with two labels ($\textcolor{blue}{\blacksquare}$ and $\textcolor{red}{\bullet}$) where the boundary is along a derivable curve (left) and its projection by a $\Phi$ function so that the separation is according to a hyperplane (right). The boundary is symbolised by a dashed line.}
    \label{fig:svm:schematic}
\end{figure}

The \textit{support vector machine} (or \textit{SVM}) method is a generalization of the classification problem where a hyperplane is used to separate points belonging to two classes with, if it is possible, a safety band. To understand this, we present an example in the two-dimensional case. On the left of the Figure~\ref{fig:svm:schematic}, a scatter plot is represented with two labels ($\textcolor{blue}{\blacksquare}$~and~$\textcolor{red}{\bullet}$); the dashed line symbolizes the boundary between the two labels, and is not linear but follows a derivable curve. The principle consists in finding a projection $\Phi$ in a space where the boundary between the two labels is a hyperplane (on the right of the Figure~\ref{fig:svm:schematic}).

The choice of transformation is therefore important. There are several basic kernels available, each with its own hyperparameters. For this method, and in order to save time, we estimated the best method in two steps: first, the search for a kernel giving better results than the others, and then the optimisation of this kernel. The kernel selected almost every time is the linear model. We have therefore decided to keep it for the procedure.

\section{Comparisons}

In this section, the experimental design is presented first, followed by the results.

\subsection{Experimental design}

For each scenario (threshold with binary or continuous values for $Y$ and self-censorship case), $n=5000$ recruitment methods were simulated for the training base and $m=n/10$ for the test base each composed of $\Nd=5$ records with $\alpha\in\{0.2,0.5,0.8\}$ for the correlation. In management studies (see the meta-analysis by \cite{paterson2016assessment}), $0.2$ is considered as a correlation corresponding to a medium effect size, $0.5$ as a correlation corresponding to a large effect and $0.8$ as a correlation corresponding to a very large effect, almost never encountered.

In order to have the same probability that two files are discriminated in the quantitative and qualitative plans, we have set for the second the thresholds $S\in\{0,\ldots,4/5\}$ and, for the first, thresholds in $S\in\{q_{\mathbb{P}(C\leq0)},\ldots,q_{\mathbb{P}(C\leq4)}\}$ where $q$ is the quantile of the centered Gaussian distribution of variance $1/\Nd$ and $C$ follows a binomial distribution of parameters $\mathcal{B}in\left(\Nd; 1/2\right)$~; thus we obtain rejection probabilities of around $\{3\%,19\%,50\%,81\%,97\%\}$ in both cases. $\mu\in\{0.4,0.8,\ldots,2\}$ were chosen for values of $\mu$.

Moreover, each case is made up of $K=5$ variables (15 in all). To limit the amount of randomness, we simulate the intrinsic values of each scenario (for example, $(\bX,\bY,\beps)$ for quantitative plans and $(\bX,\bY,\bB)$ for qualitative plans) and then apply the biases.

Finally, for each configuration, $100$ datasets are simulated and the files (only) as well as the result of the recruitment method (taken or not) are given to the algorithms to train them (without saying which files were put in competition). For each recruitment method in the test database, the error is calculated by comparing the prediction of the file selected by the algorithm and the file that would have been chosen by the \textit{perfect} ranking.

\subsection{Results}

To compare results, a graph is plotted for each scenario, each algorithm, each correlation and each censoring type, where each point represents the average predictions obtained on the test set when the algorithm was trained on the perfect dataset ($y$-axis) and the prediction on the biased dataset ($x$-axis). Each graph represents the 100 complete datasets (in turquoise) and the 100 anonymized datasets (in salmon). The graphs are then grouped by algorithm and scenario (see, for example, the figure~\ref{fig:binom_Auto_Reg_Log} for logistic regression  the self-censorship scenario), where the rows represent the correlation between $Y$ and $Z$ (from least correlated on the left to most correlated on the right) and the columns the importance of the bias (ranging from 3\% censoring on the first row to 97\% on the last). To see the influence of bias on predictions, a straight line $y=x$ is added (in black): if a point is below the straight line (resp. above it), this means that the prediction from the biased base is worse than that from the base with perfect ranking (resp. better). Finally, the ellipses represent the 95\% region, provided that the distribution of points follows a bivariate Gaussian distribution.

\begin{figure}[!ht]
    \centering
    \includegraphics[width=\linewidth]{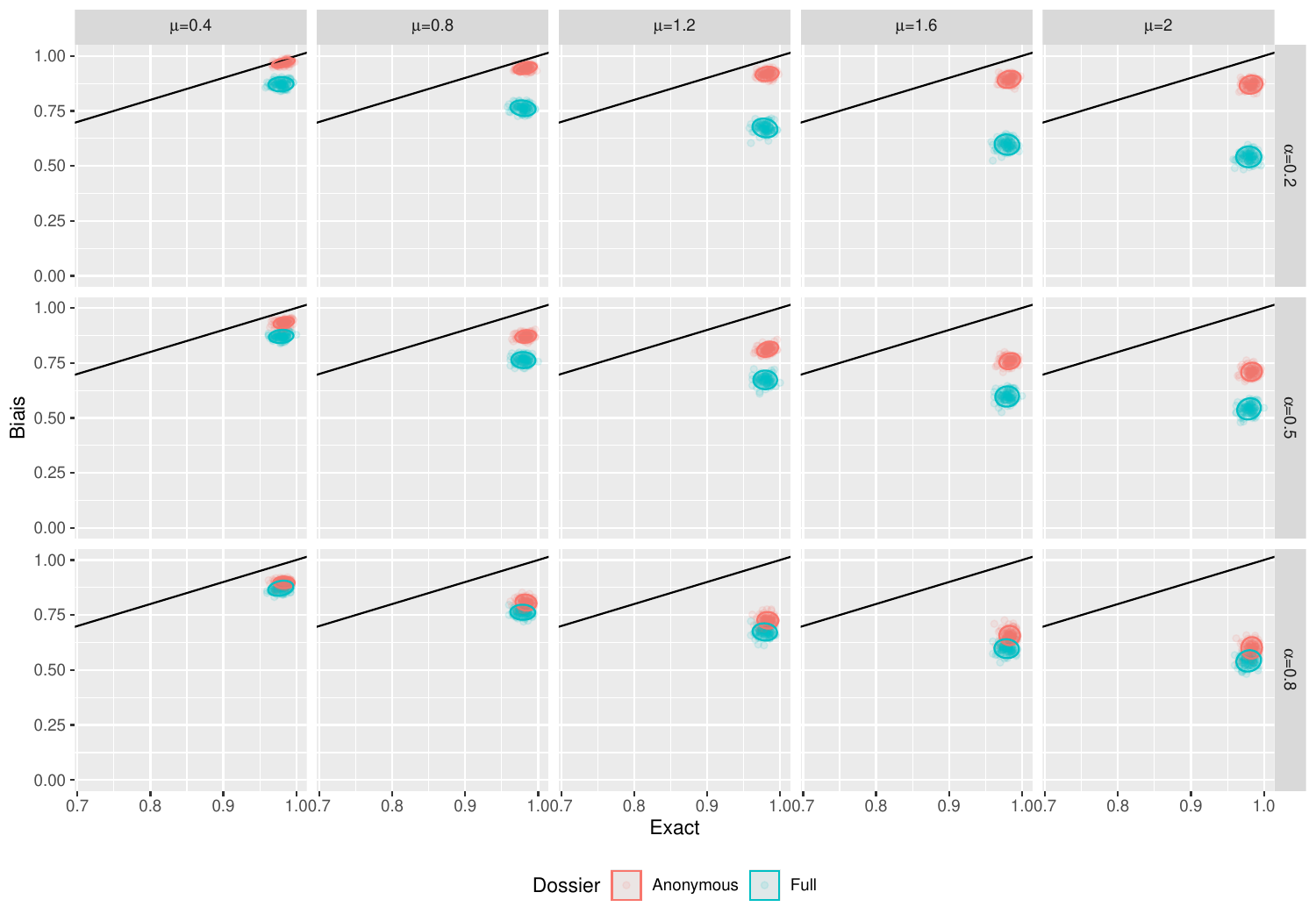}
    \caption{Representation of logistic regression results for the self-censorship scenario, where each point has as its $x$-axis the average rate of correct classifications if the algorithm were to train on perfect classification, and as its $y$-axis the biased classification according to discrimination case (columns), $\alpha$ correlation (rows) and file type (in \textcolor{TURQUOISE}{turquoise} for complete files and \textcolor{SALMON}{saumon} for anonymized files). The black line represents $y=x$ and ellipses at 95\% have been added.}
    \label{fig:binom_Auto_Reg_Log}
\end{figure}

For an overview of all the results, Table~\ref{tab:sumarise} groups together all the figures; enlarged versions are available in section~E of the supplementary information.

\begin{table}
    \centering
    \caption{Summary table of all figures by method (in rows) and scenario (in columns). For each figure, each point has as its $x$-axis the average rate of correct classifications if the algorithm were to train on perfect classification, and as its $y$-axis the biased classification according to discrimination case (columns), $\alpha$ correlation (rows) and file type (in \textcolor{TURQUOISE}{turquoise} for complete files and \textcolor{SALMON}{saumon} for anonymized files). The black line represents $y=x$ and ellipses at 95\% have been added.}
    \label{tab:sumarise}
    \begin{tabular}{c||c||c|c}
    \multicolumn{1}{c}{}&\multicolumn{3}{c}{Scenario}\\\cmidrule{2-4}
    &Self-censorship&Threshold and binary&Threshold and continuous\\\hline\hline
    \rotatebox{90}{Logistic regression}&\includegraphics[width=0.28\linewidth]{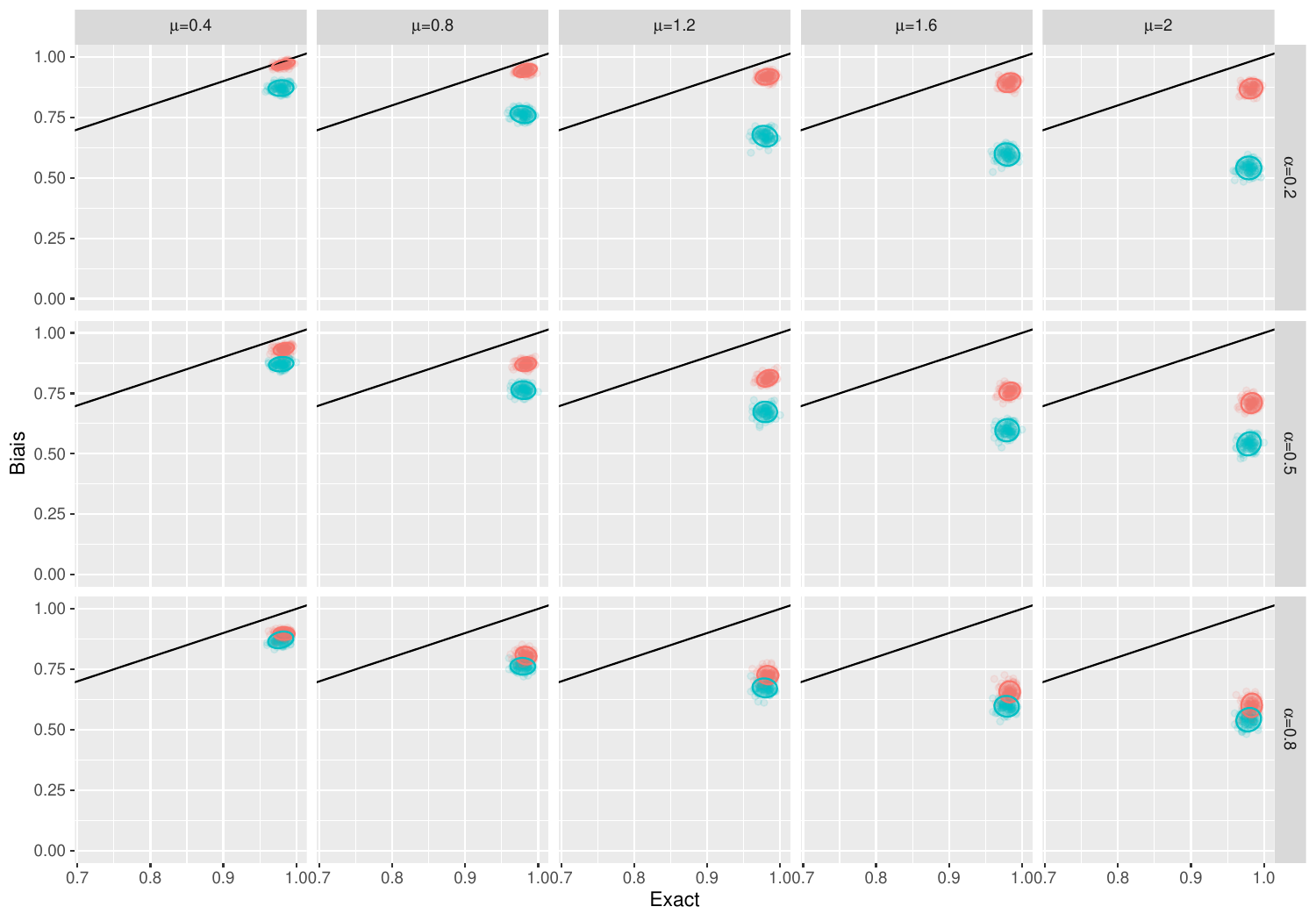}&\includegraphics[width=0.28\linewidth]{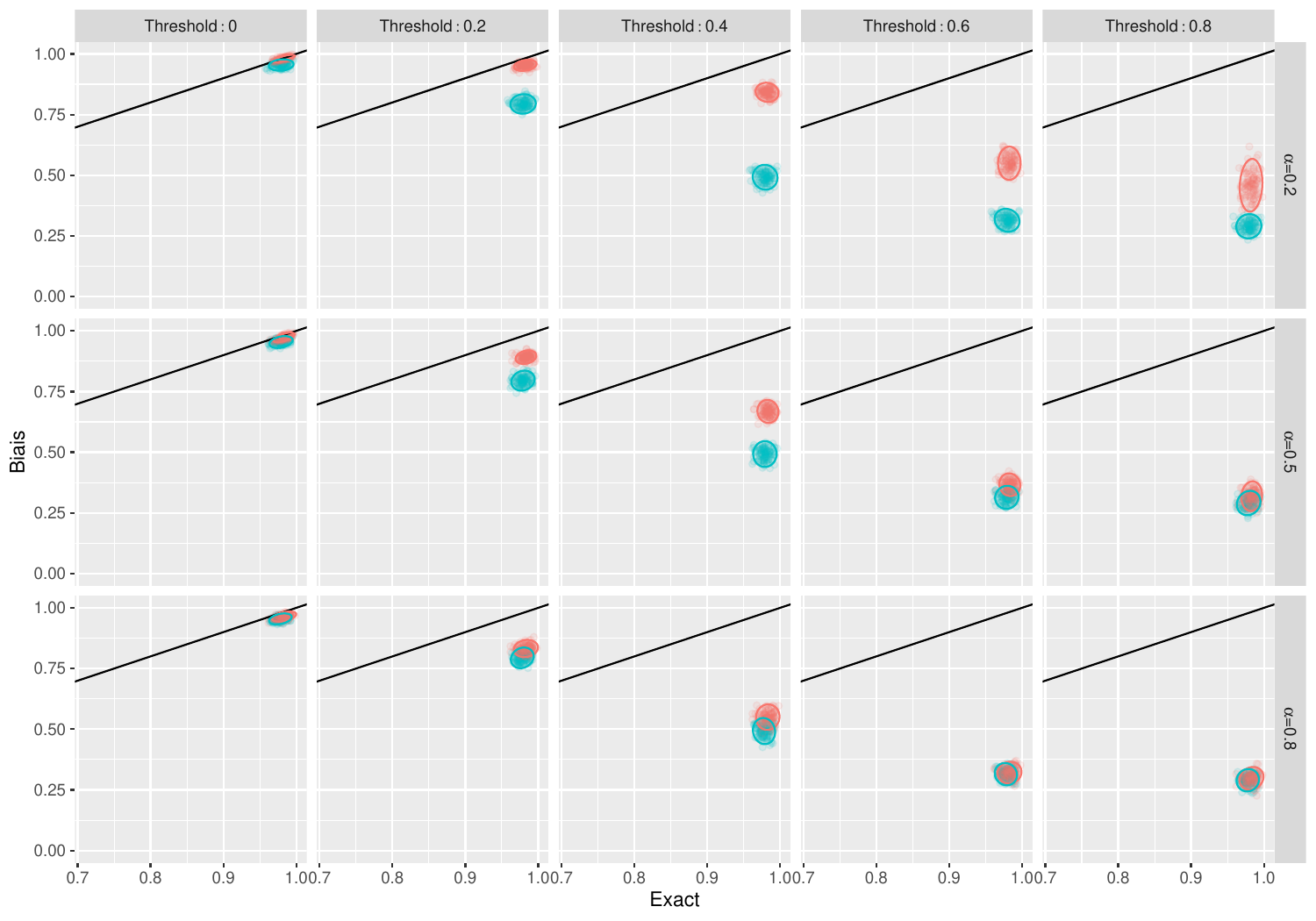}&\includegraphics[width=0.28\linewidth]{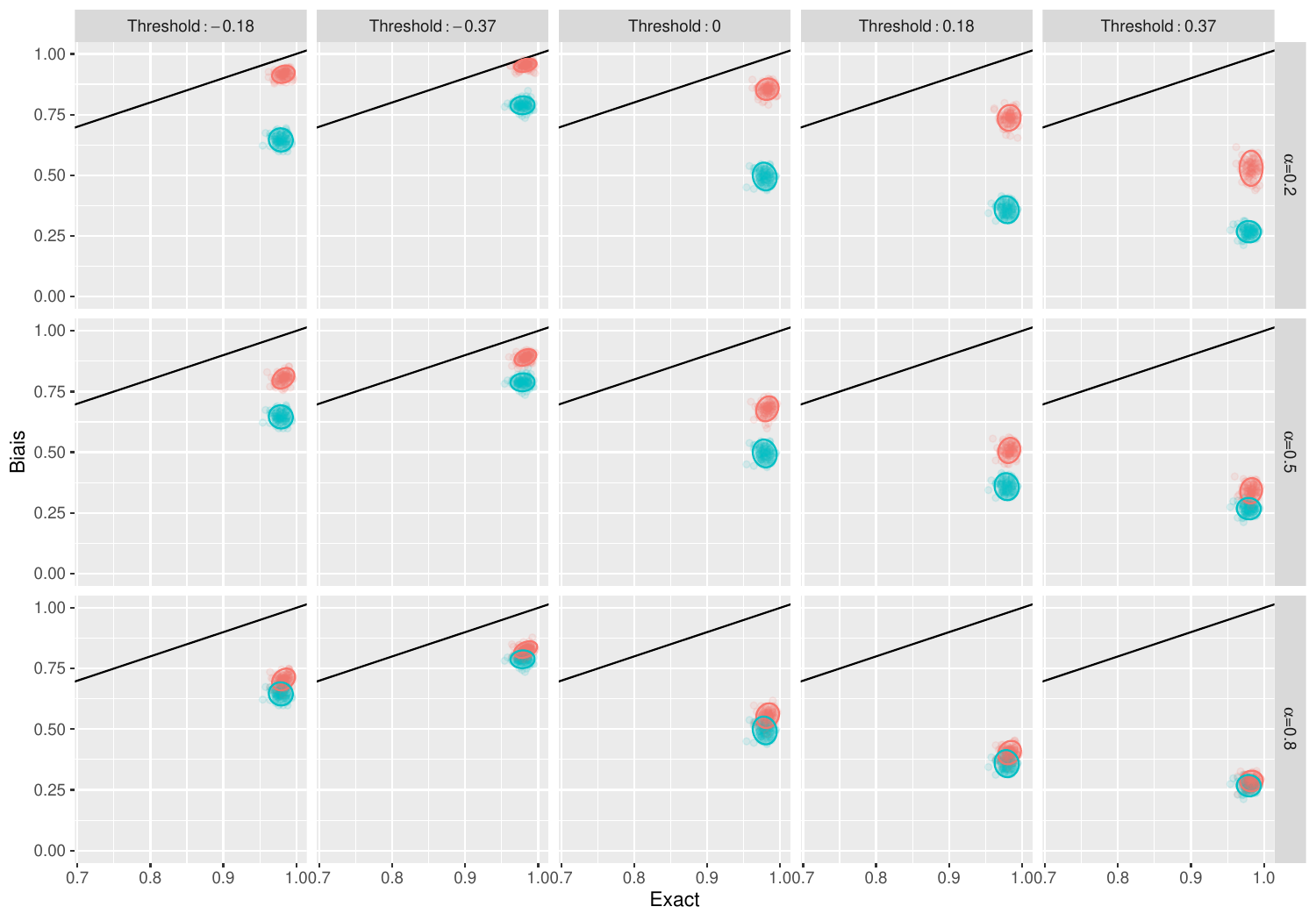}\\\hline
    \rotatebox{90}{Logistic reg + AIC}&\includegraphics[width=0.28\linewidth]{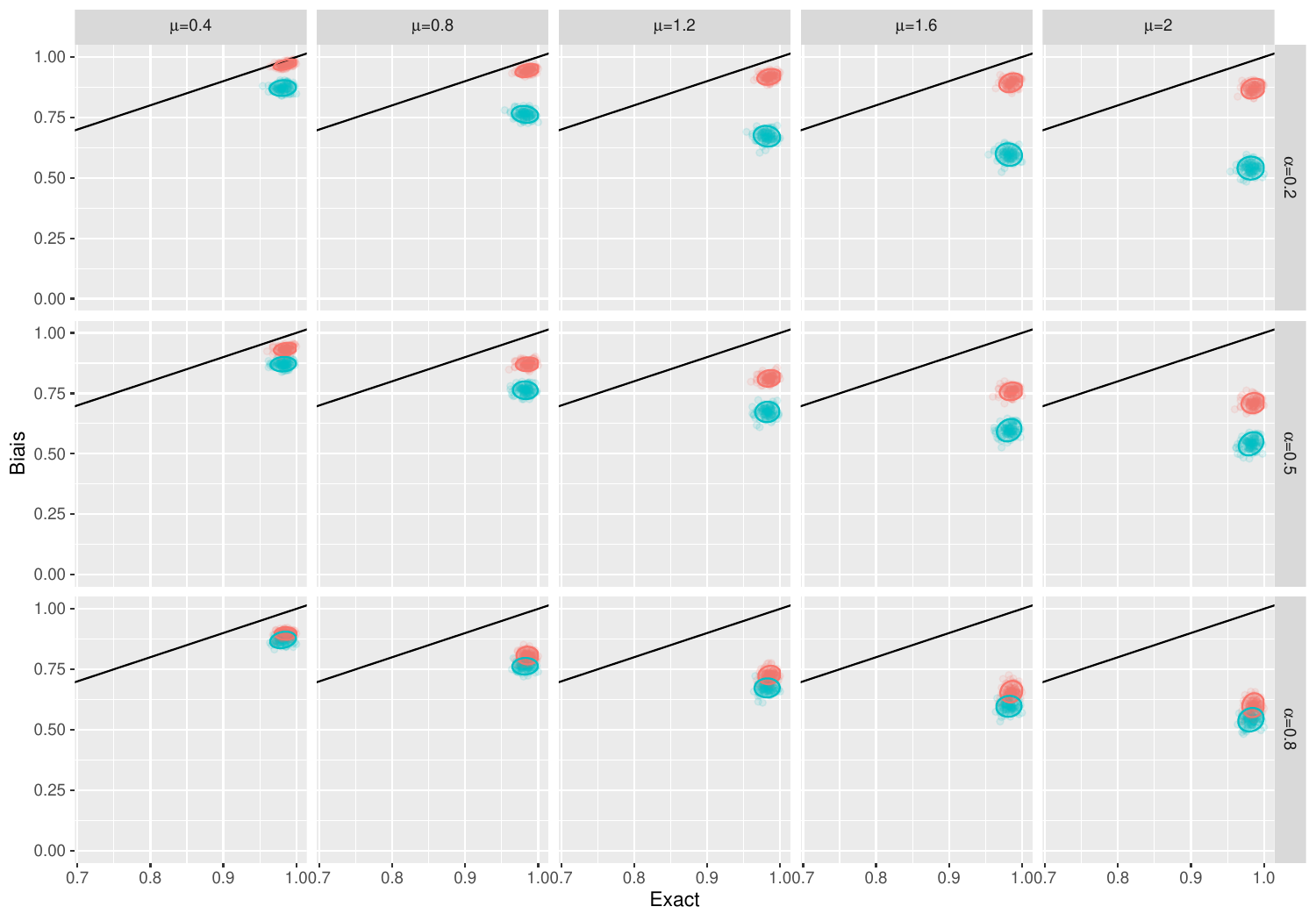}&\includegraphics[width=0.28\linewidth]{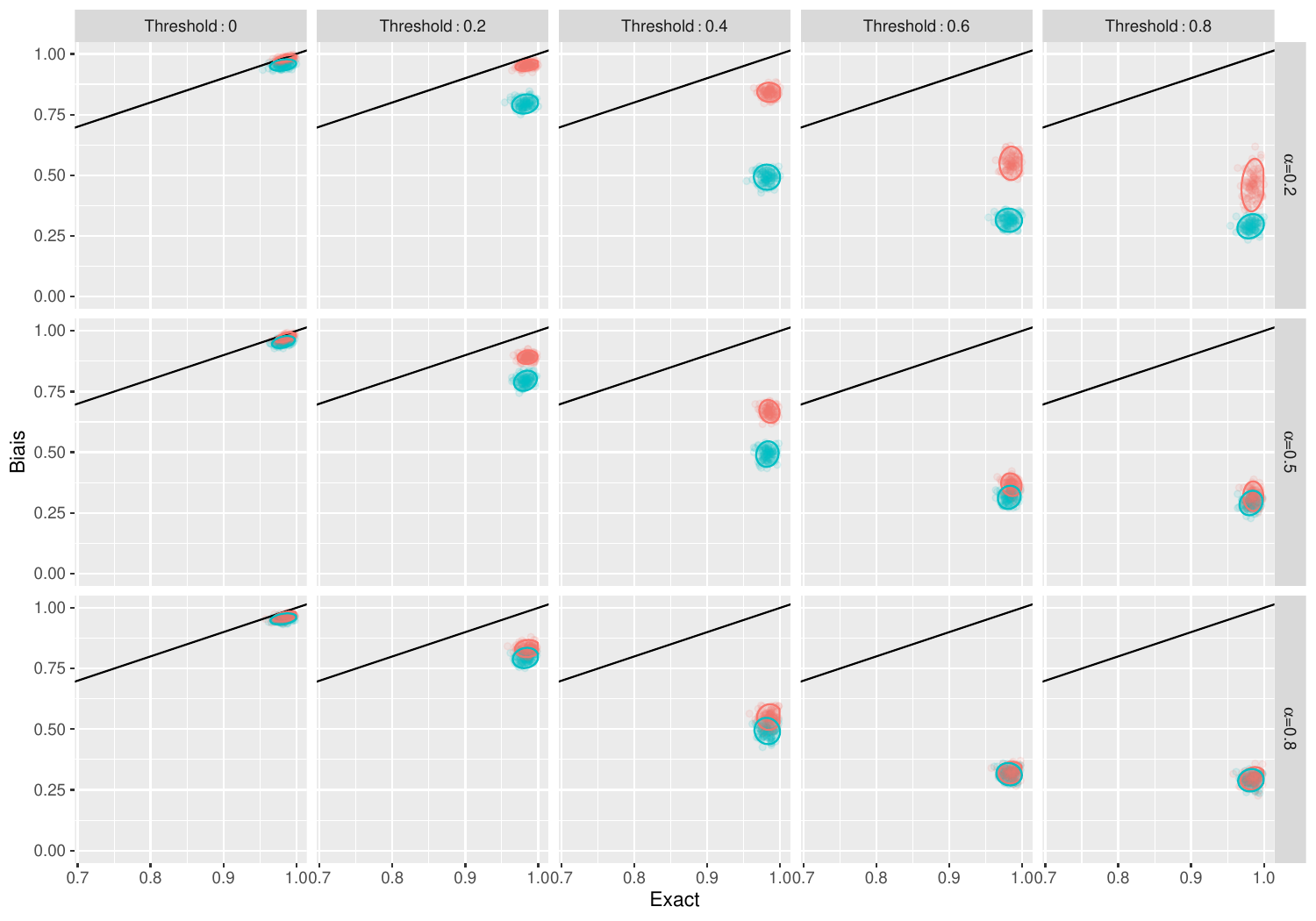}&\includegraphics[width=0.28\linewidth]{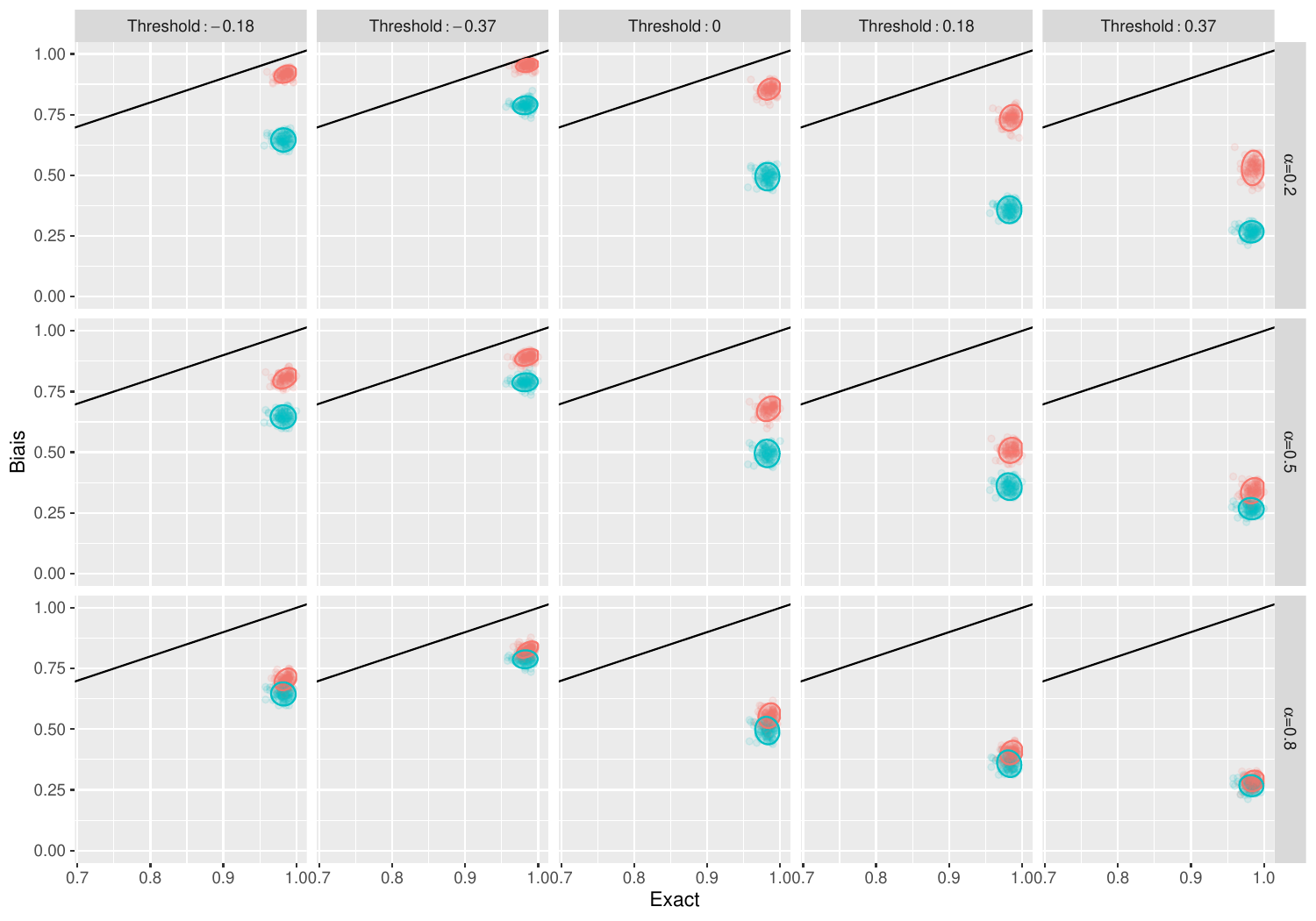}\\\hline
    \rotatebox{90}{\textit{SVM} method}&\includegraphics[width=0.28\linewidth]{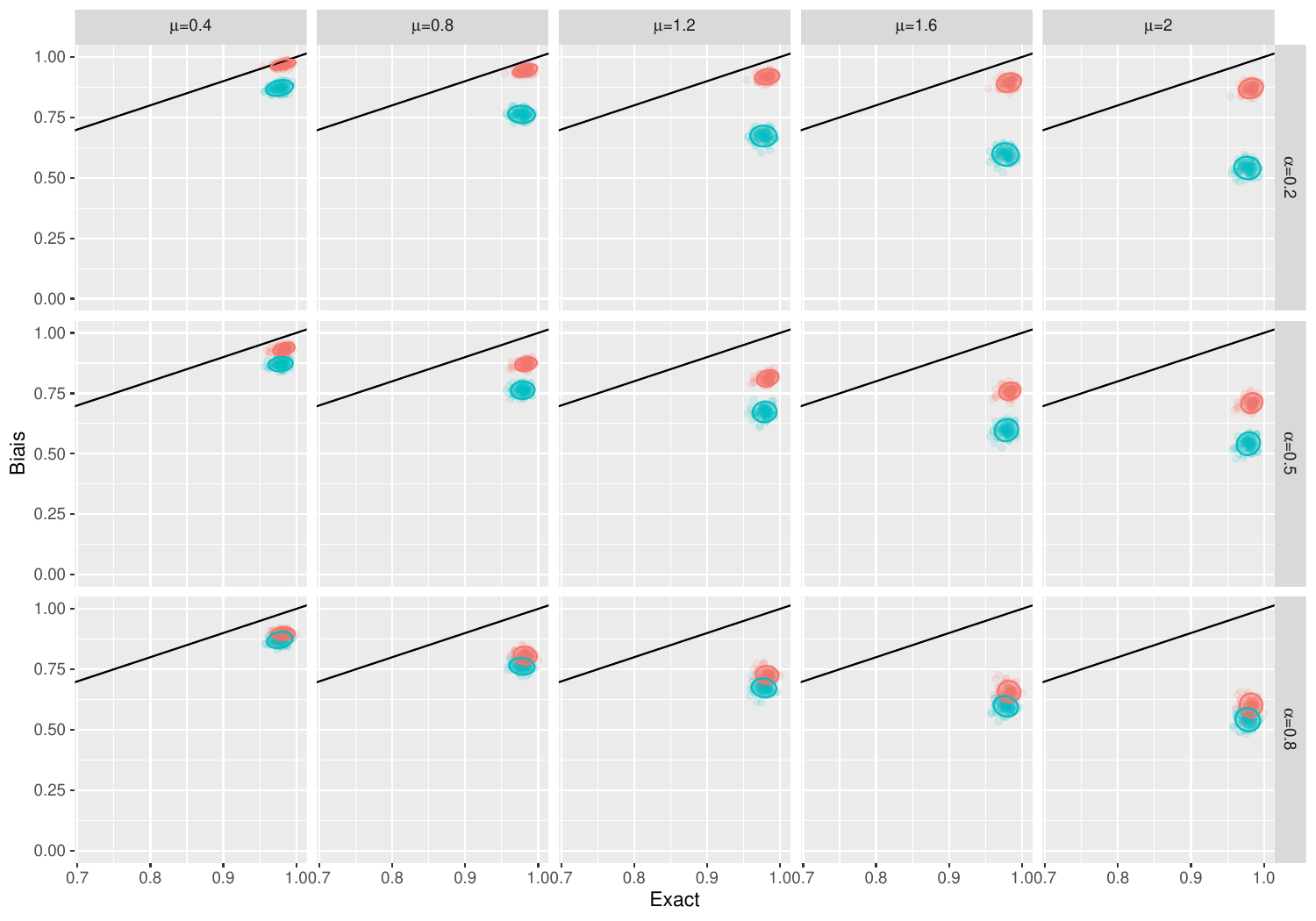}&\includegraphics[width=0.28\linewidth]{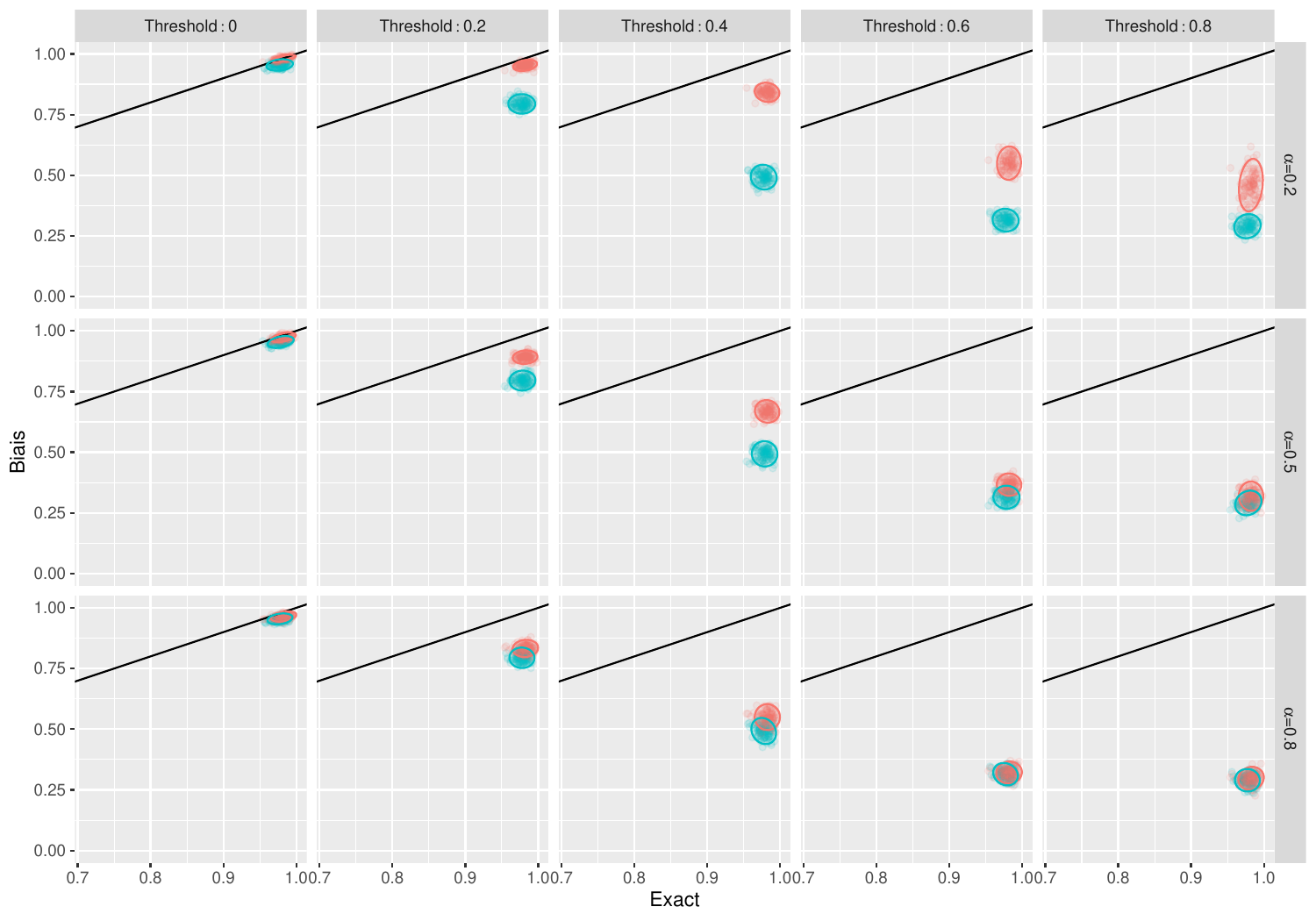}&\includegraphics[width=0.28\linewidth]{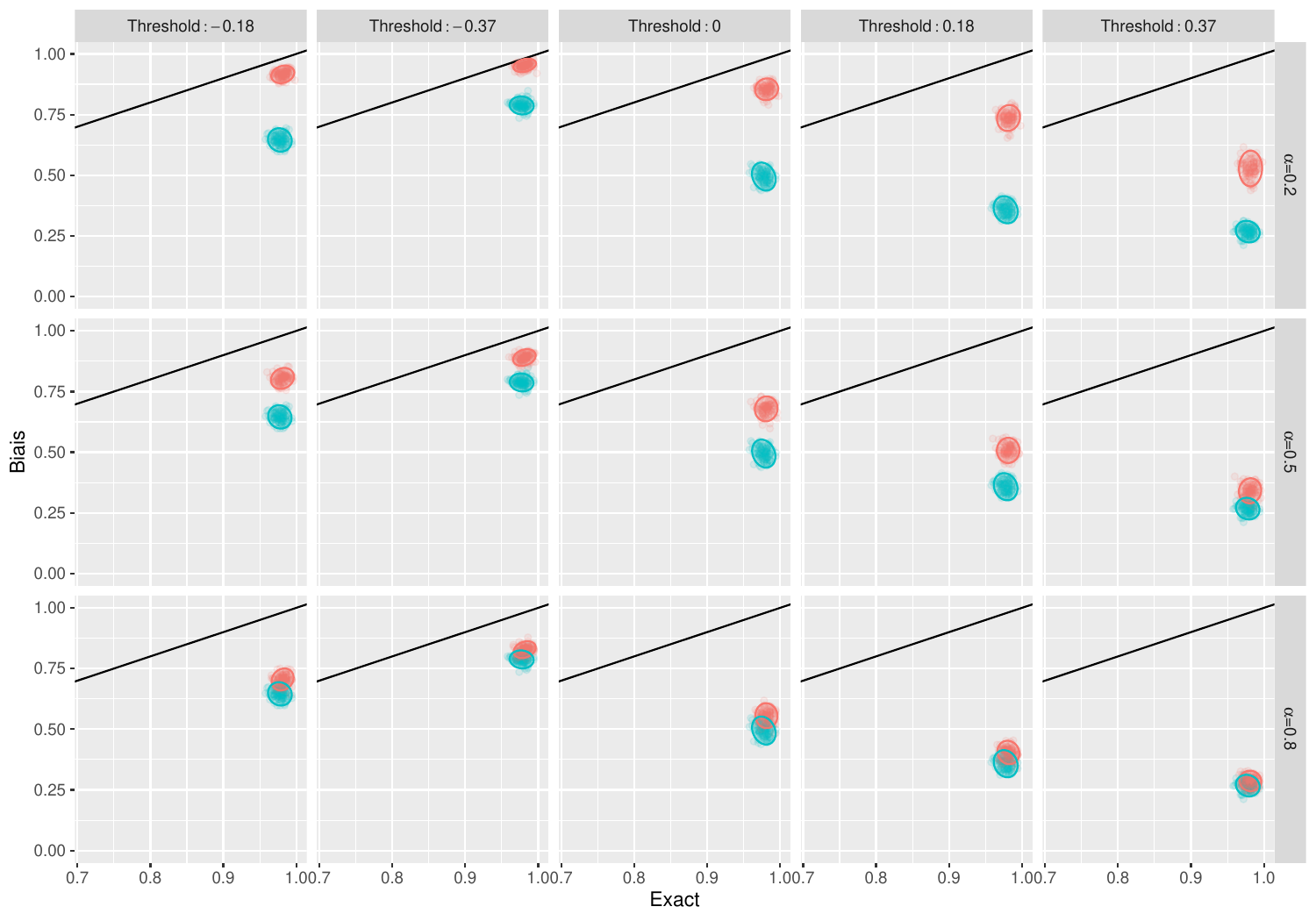}\\\hline
    \rotatebox{90}{Multi-layer perceptron}&\includegraphics[width=0.28\linewidth]{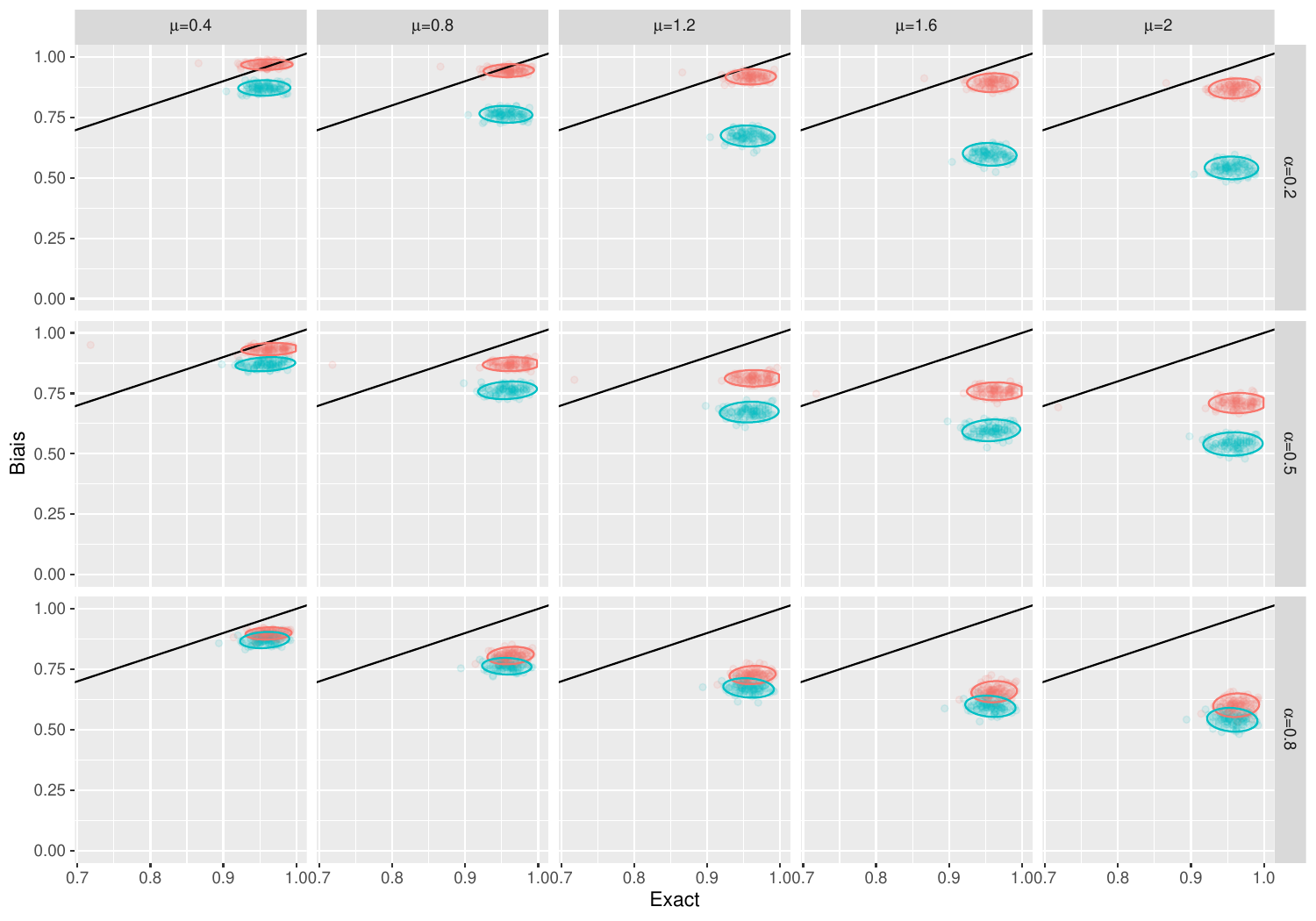}&\includegraphics[width=0.28\linewidth]{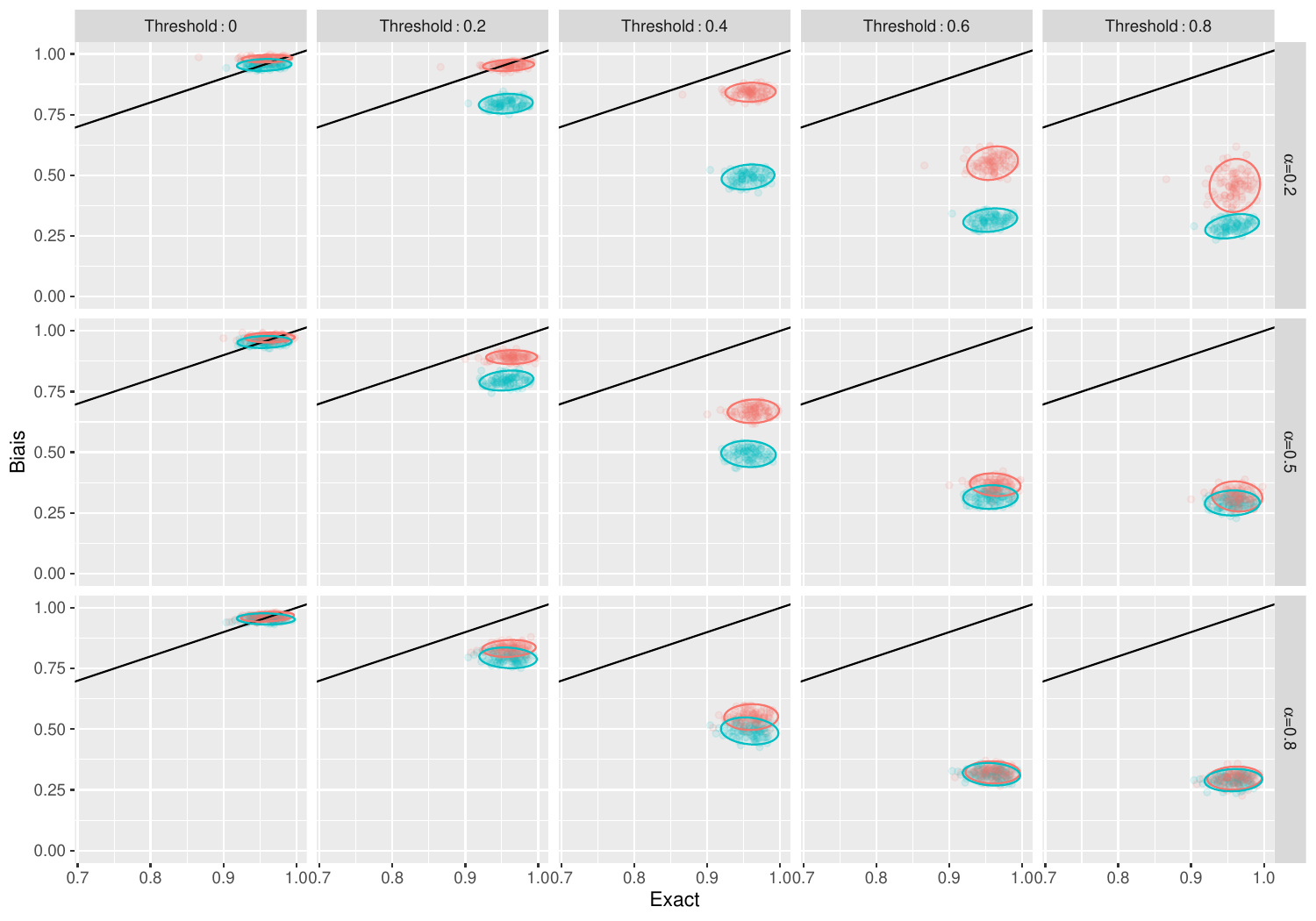}&\includegraphics[width=0.28\linewidth]{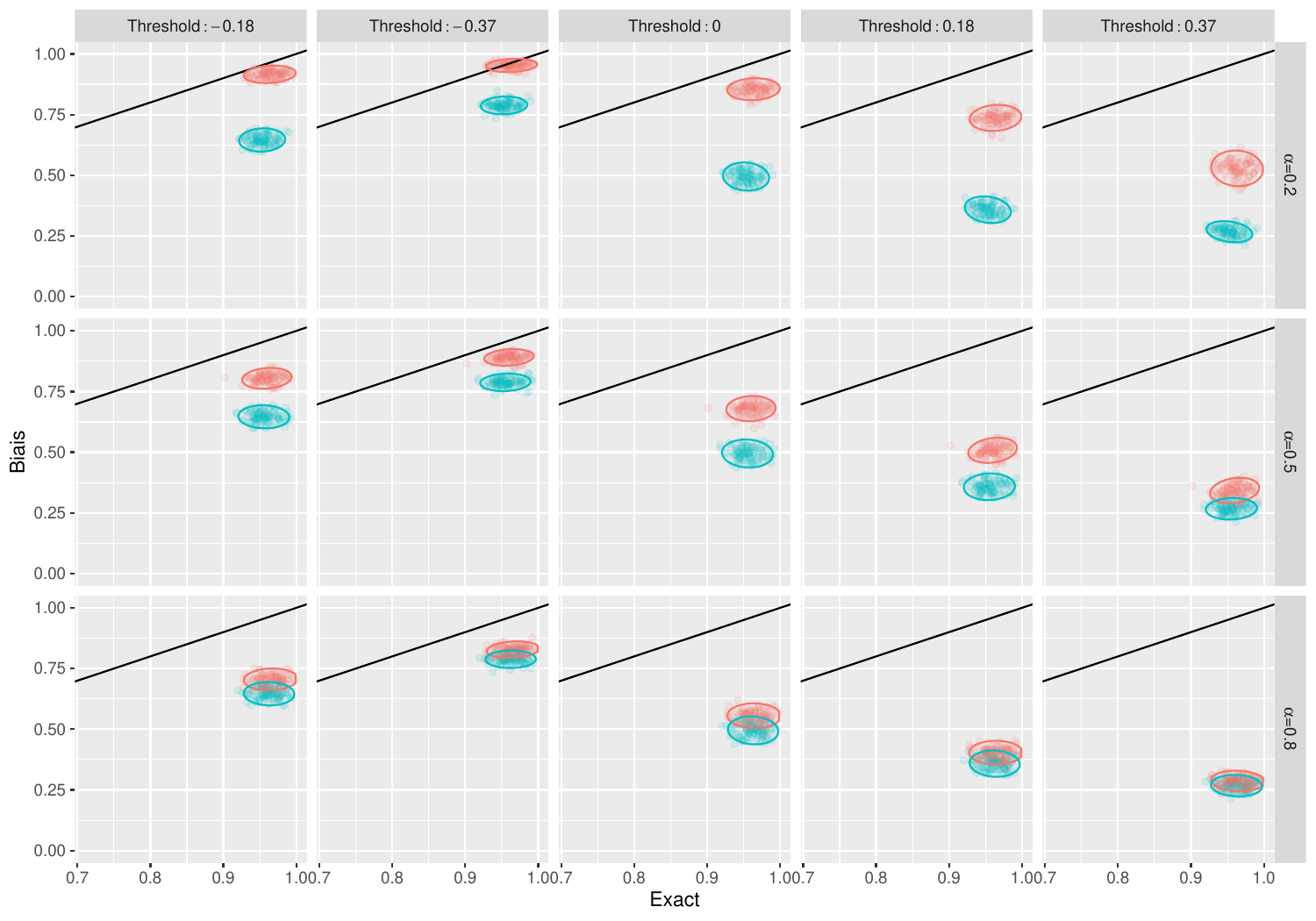}\\\hline
    \rotatebox{90}{$L$-nearest neighbors}&\includegraphics[width=0.28\linewidth]{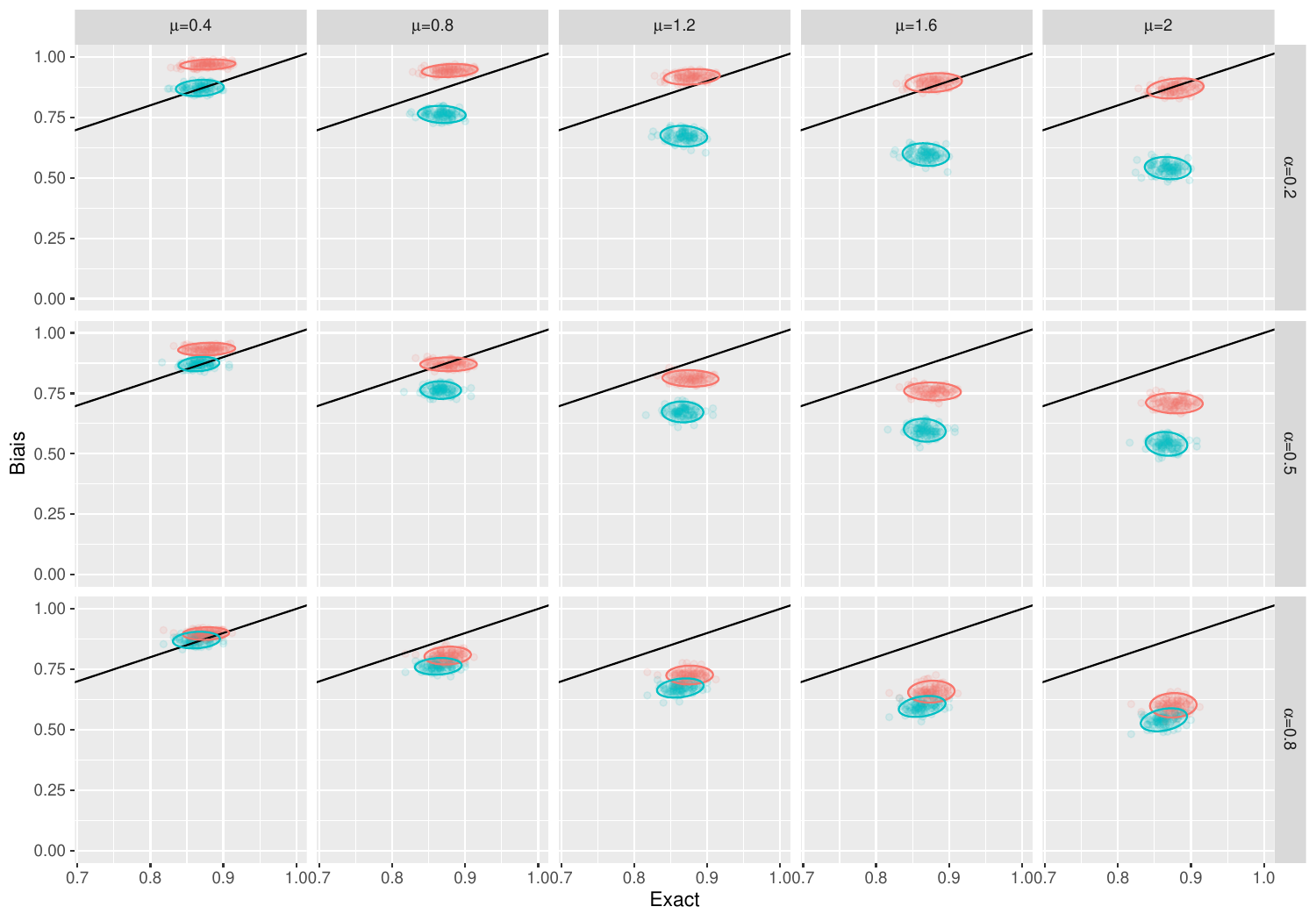}&\includegraphics[width=0.28\linewidth]{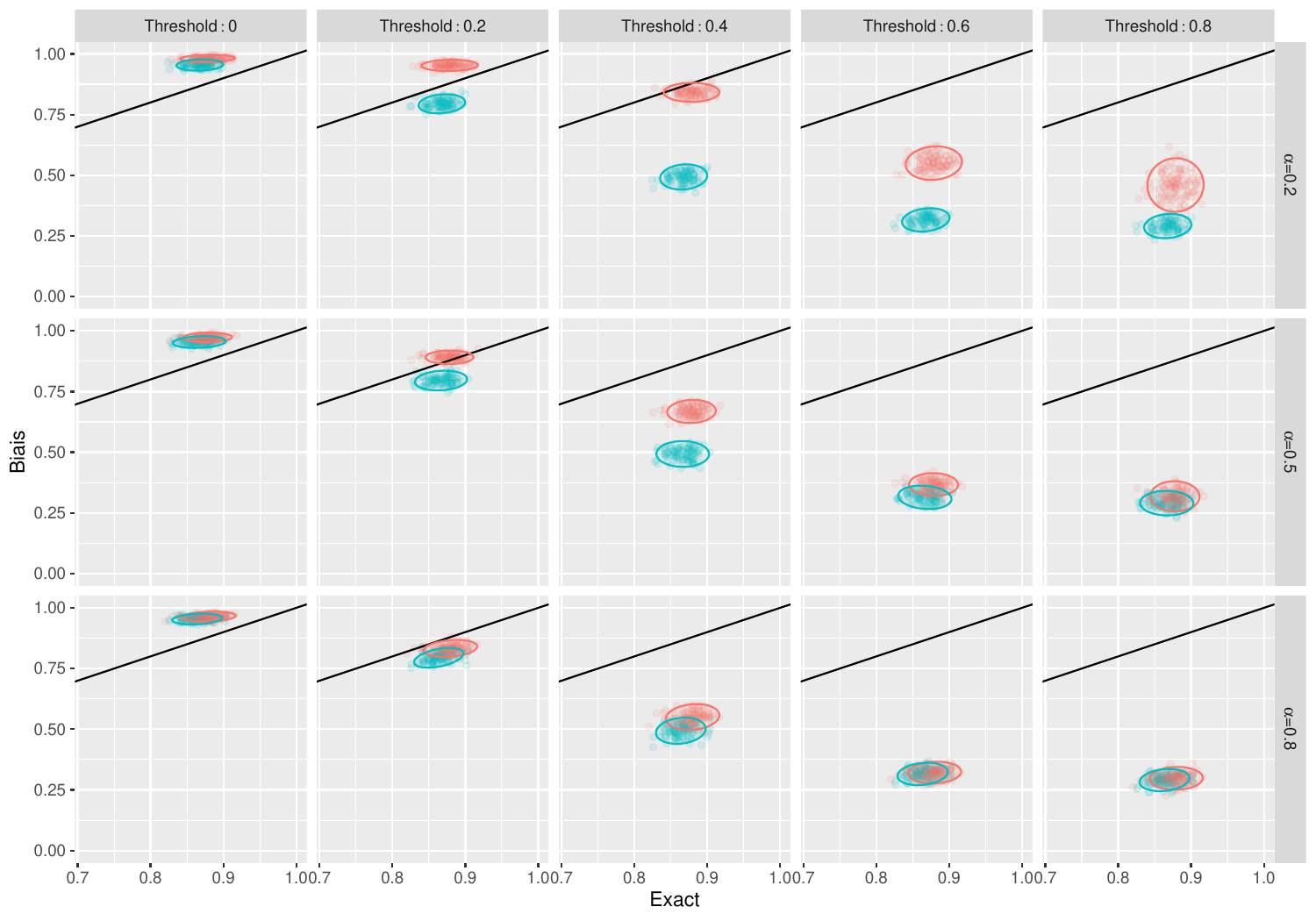}&\includegraphics[width=0.28\linewidth]{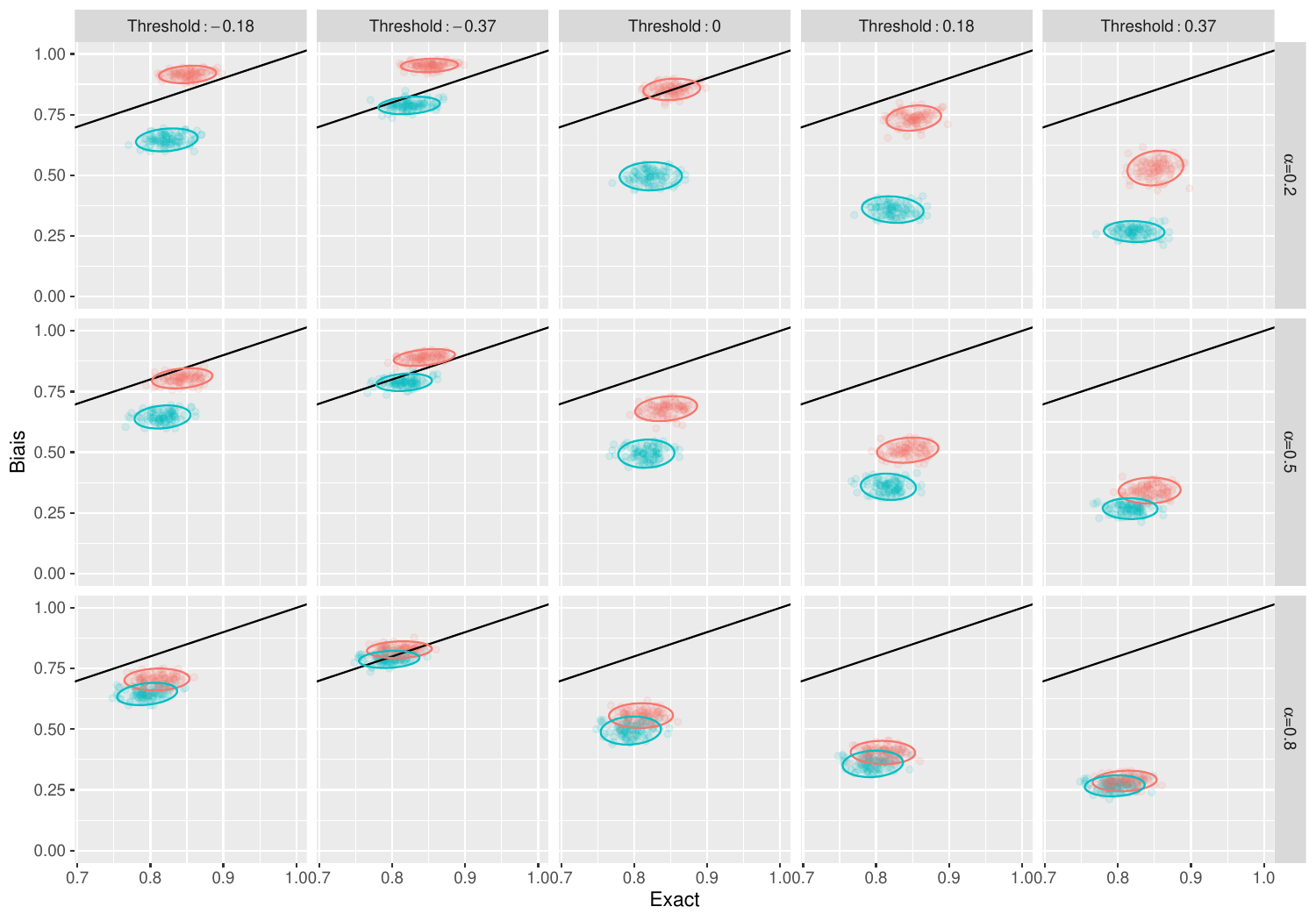}\\
    \end{tabular}
\end{table}

From the table~\ref{tab:sumarise}, it can be seen that logistic regression (with or without AIC criteria), the SVM method and multi-layer perceptrons have globally the same behaviour. In figure~\ref{fig:binom_Auto_Reg_Log}, the scatterplots are generally below the black line, which shows a worse prediction of the algorithms that were trained on the biased basis than those trained with a perfect classification. In the worst cases, the prediction goes from about 97\% on the exact basis (on the abscissa) to about 55\% for the biased basis.

It appears that the use of sophisticated "black box" algorithms such as SVM did not produce more convincing results  than classic, transparent algorithms such as logistic regression. This result is interesting from a practical point of view, as it should be noted that the sophistication of matching algorithms is one of the main marketing arguments used by developers of software solutions to convince recruiters to adopt these systems. However, this result must be interpreted with caution, as we did not optimize the hyperparameters of the most complex models.

Note now the difference between complete data (in turquoise on the figures) and anonymous data (in salmon). In the case of low correlation (top row), the point scatter plots are separated with an average better prediction of around 25\% for the algorithms having learned on anonymous data for a 97\% biased base. However, the clouds are very close in the case of a very strong correlation (bottom line), whatever the degree of bias. In their article, \citet{besse2020detecter} observe that the anonymization of their database on housing loans had not improved the prediction of the algorithms; it would be interesting to repeat the study to see if there are any variables strongly correlated with discrimination, or even to repeat the study in this case by removing them.

\begin{figure}[!ht]
    \centering
    \includegraphics[width=\linewidth]{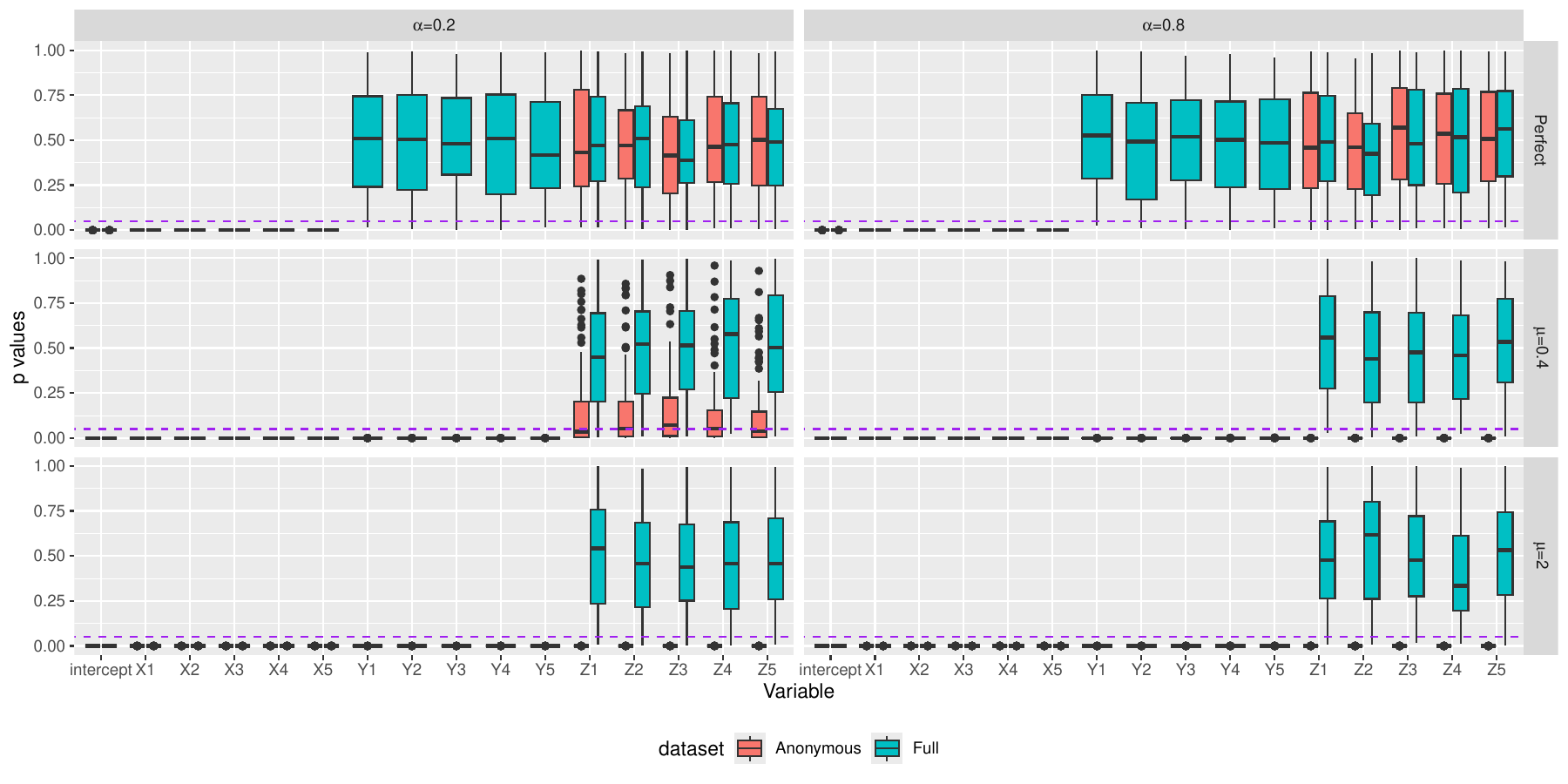}
    \caption{Boxplot representation of the $p$-value for each coefficient (on the x-axis) of the test assuming that it is zero according to whether the training dataset was complete or not (color of the boxplots), as a function of correlation (columns) and whether the data are unbiased (top), slightly biased (middle row) or highly biased (bottom row) for logistic regression estimation in the case of self-censoring. A \textcolor{purple}{purple} dashed line has been added to highlight the 5\% value.}
    \label{fig:glm:pvalue}
\end{figure}

Figure~\ref{fig:glm:pvalue} shows the $p$-values for each coefficient (on the x-axis) of the test assuming that it is zero (and therefore has no influence on prediction) according to whether the training dataset was complete or not (color of the boxplots), whether the data are unbiased (top row), weakly biased (middle row) or strongly biased (bottom row) and according to the correlation between $\bY$ and $\bZ$ (columns). These four scenarios correspond to the figures in the four corners of figure~\ref{fig:binom_Auto_Reg_Log}. First, we observe that there are no $p$ values for the $Y$ variables in the case of anonymized data, since they were not supplied. In the case where the algorithm is trained on the perfect base (top line), we can see that the values of the y-intercept $p$ and the variables $\bX$ are all close to $0$, which means that they were found to be useful for predicting the award of an recruitment method, whatever the correlation between $\bY$ and $\bZ$; on the other hand, only a few values of the variables $\bY$ and $\bZ$ are below 5\% (purple dotted line), which means that, in general, the algorithm does not attach any importance to these variables (this is the normal error rate for tests).

In the case of training on biased data with all variables (boxplots in turquoise), the algorithm retains the intercept and coefficients of the $\bX$ and $\bY$ variables whether the bias is large (bottom line) or small (middle line). On the other hand, the $p$-values of the coefficients of the $\bZ$ variables are globally above the 5\% value, showing that the algorithm does not consider these variables to be involved in the choice of whether or not to pass the recruitment method. Finally, in the case of biased and anonymous data (salmon-colored boxplots), the algorithm always retains the intercept and the $\bX$ variables in the final decision. On the other hand, as the $\bY$ variables are not present, we see that the algorithm restores importance to the $\bZ$ variables. In the case of weakly correlated data with low bias (center left graph), each variable $Z_k$ has $p$-values below $5\%$ in just over 50\% of simulations, leaving this variable with some influence. On the other hand, in the other cases (a strong correlation or a highly biased database), the $p$-values are all close to $0$: the algorithm compensates for the anonymization by fetching the biased information from the correlated data.

\begin{figure}[!ht]
    \centering
    \includegraphics[width=\linewidth]{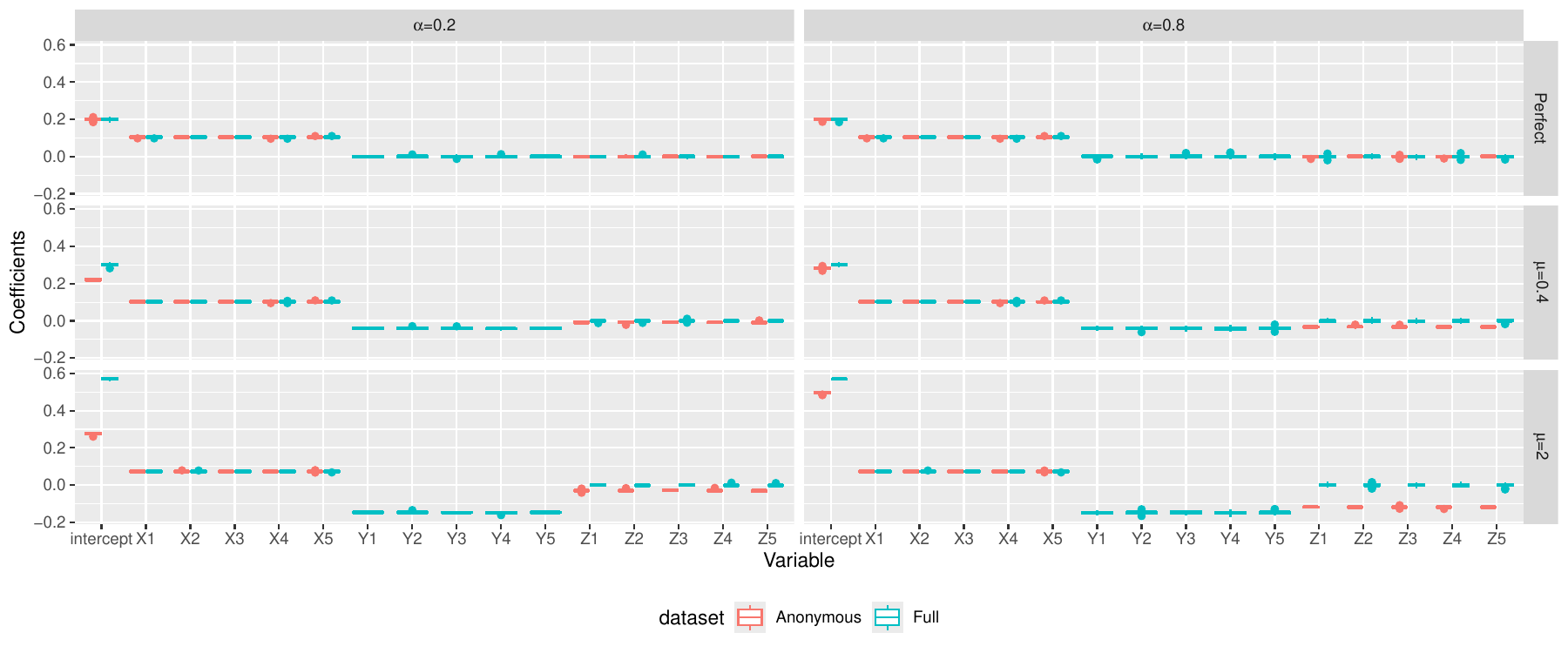}
    \caption{Boxplot representation of the coefficients for each variable (on the x-axis) of the test assuming that it is zero according to whether the training dataset was complete or not (color of the boxplots), as a function of correlation (columns) and whether the data are unbiased (top), slightly biased (middle row) or highly biased (bottom row) for logistic regression estimation in the case of self-censoring.}
    \label{fig:glm:coefficient}
\end{figure}

Figure~\ref{fig:glm:coefficient} shows the coefficient estimates whatever the associated $p$-value. We can see that the coefficients of the intercept and the $\bX$ variables are all positive, while those of the $\bY$ and $\bZ$ variables are negative. In the case of a weakly biased database (central line), the coefficients of $\bY$ are close to -0.2, compared with values closer to -1.5 for the strongly biased database (bottom line). Finally, even when the $p$-values are close to $0$ for the $\bZ$ variables, the coefficients are lower than for the $\bY$ variables, or even virtually zero when correlation and bias are low (figure center left).

\begin{figure}[!ht]
    \centering
    \includegraphics[width=\linewidth]{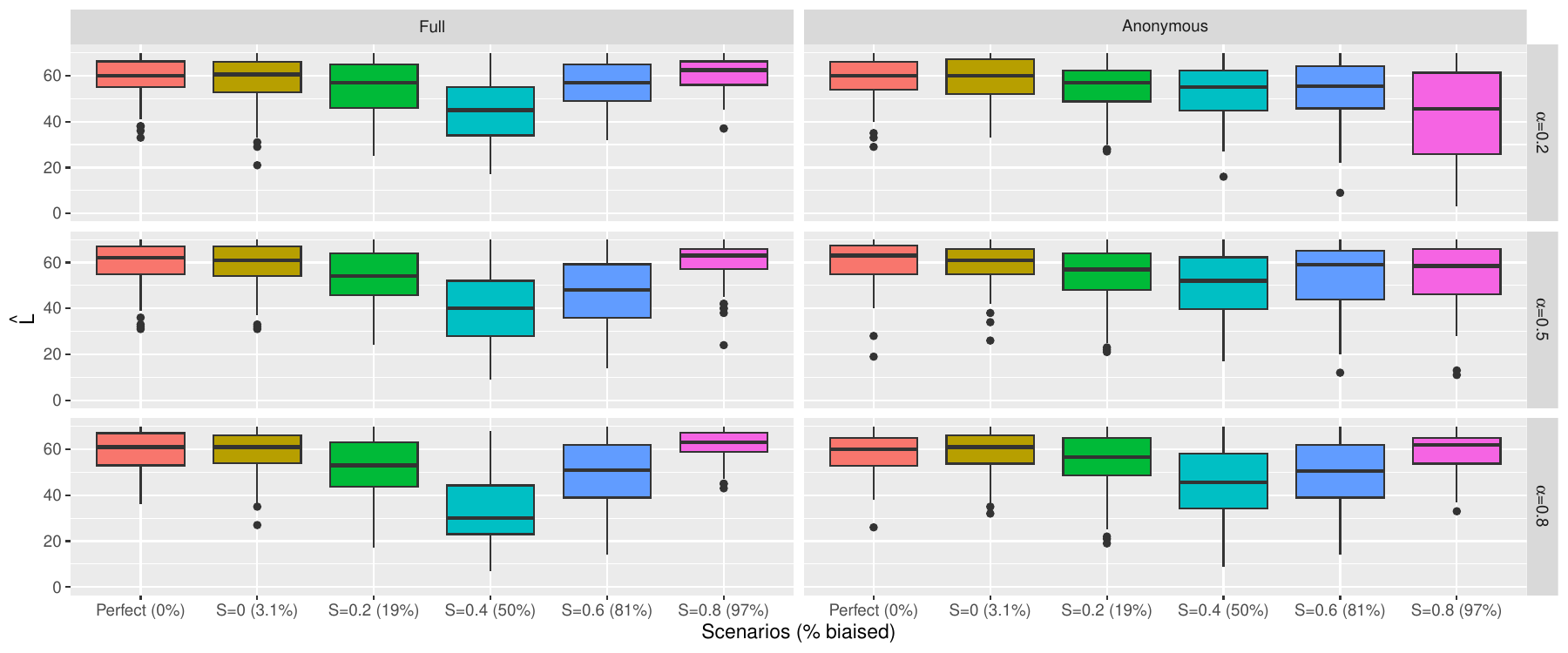}
    \caption{Boxplot representation of the $L$ values estimated by the different procedures in section~\ref{sec:Lnn} according to whether the files are complete (left) or anonymous (right) as a function of the percentage of files rejected by the threshold (x-axis).}
    \label{fig:knn}
\end{figure}

Finally, note the difference of the $L$ nearest neighbors procedure (last row of table~\ref{tab:sumarise}), as it is the only one where training on the biased basis sometimes gives better predictions than training with the perfect basis. Indeed, when the bias is low, whatever the correlation, the clouds lie above the $y=x$ line (Left-hand columns in graphs). This is particularly interesting in the case of threshold-biased data (figures bottom center and right), where the only difference at the time of prediction is the value used for $L$, since neighbors are defined by the same values for $\bX$, $\bY$ and $\bZ$. On figure~\ref{fig:knn} are represented the boxplots of the estimates of $L$ according to the various scenarios for the biased binomial data (bottom figure in the center of the table~\ref{tab:sumarise}). It would be interesting to extend the study to see whether, by improving predictions in the case of unbiased data (since this is the worst algorithm in terms of prediction when the database is perfect), this behaviour remains and so understand how to use this bias to improve prediction.

\section{Discussions}

In conclusion, in this article we propose a protocol for simulating recruitment methods and test the main methods of supervised classification by training the algorithms on perfect or biased data. We note that the algorithms can make a good prediction (of the order of 97\%) when the learning bases are said to be \textit{perfect} and there is no bias. 

On the other hand, predictions deteriorate as soon as the learning bases are biased. In the area of recruitment, training data may be biased because they are built mainly from available data on applicants, whose characteristics are compared with those of the employees actually hired. Biases can appear at this stage if the structure of the training base does not reflect the structure of the general population: the biases then lie in the over- or under-representation of certain categories of people \citep{hall2023systematic}.
If the training data is biased, two mechanisms can lead to discrimination \citep{barocas2016big}:
\begin{itemize}
    \item If the "good candidates" used as a reference in the algorithm reflect a stereotype (e.g., assertiveness and leadership skills associated with "masculine" traits), these stereotypes are likely to be reproduced in hiring recommendations.
    \item If predictive hiring tools draw probabilistic inferences from a biased sample of the population (e.g., a sample in which women are underrepresented), any hiring decision based on these inferences will systematically disadvantage candidates who are underrepresented in the training data (the proportion of women hired will be low).
\end{itemize}

We also note that the influence diminishes when the discriminating information is removed from the base and the other variables are not too correlated with this discrimination. This result constitutes empirical evidence in favour of the anonymization of CVs and in favour of the use of tests, since these methods are based solely on job-related information and eliminate non-job-related information likely to lead to discrimination such as an ethnic-sounding name \citep{derous2019your,lacroux2020anonymous}.

On the other hand, if there are still variables that are too correlated with the discriminating variables, the algorithms will give predictions close to those proposed by the same algorithms trained on complete databases; the algorithms will thus go looking for the slightest correlation to perpetuate the bias. This situation has already been highlighted in several studies, showing that despite the anonymization of the résumé , ethnicity cues contained in résumés (associative affiliations, professional experience) can lead to discrimination \citep{kang2016whitened}.

Note that the plans propose biased bases where resumes are not supplied to recruiters as soon as the bias is too great, or bases where the bias is intrinsic to the person (for example, people who would lose their nerve in front of a math test if it were specified that it was math). 

Looking ahead, it would be interesting to be able to propose procedures for evaluating the algorithms we bring to market. We can also envisage new scenarios closer to the profiles of the people recruited, in order to understand which profiles would be most discriminated against by the algorithms. In particular, it would be interesting to see how the variables deemed to be discriminating are correlated with the other variables; or whether we can predict a discriminant variable using the other variables, for example. The question may arise of how to anonymise a CV to guarantee a correlation below 0.2 with the discriminating variables, in order to counter external bias. We also need to understand how self-censorship manifests itself in order to propose scenarios that are as close as possible to reality.

\newpage

\begin{center}
{\large\sc Supplementary information}
\end{center}

\begin{appendices}

\section{Proof of correlation between $\Yijk$ and $\Zijk$}

In this section, the proof of the correlation in each case is detailed.

\subsection{Binary case}

As the variables $\Bijk$ are independent, independent of the variables $\Xijk$ and $\Yijk$ and follow the same Bernoulli distribution $\mathcal{B}\left(1/2\right)$ and $\Uijk$ are independent, independent of the variables $\Xijk$, $\Yijk$ and $\Bijk$ and follow the same Bernoulli distribution $\mathcal{B}\left(\alpha\right)$, for each triplet $(i,j,k)$, the covariance between $\Yijk$ and $\Zijk$ is:
\begin{eqnarray*}
    \Cov{\Yijk}{\Zijk}&=&\Cov{\Yijk}{\Uijk\Yijk+(1-\Uijk)\Bijk}\\
    &=&\Cov{\Yijk}{\Uijk\Yijk}+\underbrace{\Cov{\Yijk}{(1-\Uijk)\Bijk}}_{=0}\\
    &=&\Esp{\Yijk\times \Uijk\Yijk}-\Esp{\Yijk}\Esp{\Uijk\Yijk}\\
    &=&\Esp{\Yijk^2\Uijk}-1/2\times \Esp{\Uijk}\Esp{\Yijk}\\
    &=&\Esp{\Yijk}\Esp{\Uijk}-1/2\times \alpha\times 1/2\quad\text{because }\Yijk^2=\Yijk,\\
    &=&1/2\alpha - 1/4\alpha\\
    &=&1/4\alpha.\\
\end{eqnarray*}
Then, for the variance:
\begin{eqnarray*}
    \Var{\Yijk}&=&1/4\\
    \text{and }\Var{\Zijk}&=&\Var{\Uijk\Yijk+(1-\Uijk)\Bijk}\\
    &=&\Var{\Uijk\Yijk}+2\Cov{\Uijk\Yijk}{(1-\Uijk)\Bijk}+\Var{(1-\Uijk)\Bijk}\\
    &=&\Var{\Uijk}\Var{\Yijk}+2\left[\Esp{\Uijk\Yijk(1-\Uijk)\Bijk}\right.\\
    &&\quad\left.-\Esp{\Uijk\Yijk}\Esp{(1-\Uijk)\Bijk}\right]+\Var{\Bijk}+\Var{\Uijk\Bijk}\\
    &=&\alpha(1-\alpha)/4+2\times\left[\Esp{\Uijk\Yijk\Bijk}-\Esp{\Uijk^2\Yijk\Bijk}\right.\\
    &&\quad\left.-\Esp{\Uijk}\Esp{\Yijk}\Esp{(1-\Uijk)}\Esp{\Bijk}\right]+1/4+\Var{\Uijk}\Var{\Bijk}\\
    &=&\alpha(1-\alpha)/4+2\times\big[\underbrace{\Esp{\Uijk\Yijk\Bijk}-\Esp{\Uijk\Yijk\Bijk}}_{=0}\\
    &&\quad\left.-\alpha/4(1-\alpha)/4\right]+1/4+\alpha(1-\alpha)/4\\
    &=&2\alpha(1-\alpha)/4-2\alpha(1-\alpha)/4+1/4\\
    &=&1/4.\\
\end{eqnarray*}

Finaly, the correlation is equal to:
\[\Corre{\Yijk}{\Zijk}=\frac{\Cov{\Yijk}{\Zijk}}{\sqrt{\Var{\Yijk}\times\Var{\Zijk}}}=\frac{1/4\alpha}{\sqrt{1/4\times1/4}}=\alpha.\]

\subsection{Continuous case}

As the variables $\epsijk$ are independent, independent of the variables $\Xijk$ and $\Yijk$ and follow the same Bernoulli distribution $\mathcal{N}\left(0,1\right)$, for each triplet $(i,j,k)$, the covariance between $\Yijk$ and $\Zijk$ is:
\begin{eqnarray*}
    \Cov{\Yijk}{\Zijk}&=&\Cov{\Yijk}{\frac{\alpha}{\sqrt{1-\alpha^2}}\Yijk+\epsijk}\\
    &=&\Cov{\Yijk}{\frac{\alpha}{\sqrt{1-\alpha^2}}\Yijk}+\underbrace{\Cov{\Yijk}{\epsijk}}_{=0}\\
    &=&\frac{\alpha}{\sqrt{1-\alpha^2}}\Var{\Yijk}.\\
    &=&\frac{\alpha}{\sqrt{1-\alpha^2}}.\\
\end{eqnarray*}

Then, for the variance:
\begin{eqnarray*}
    \Var{\Yijk}&=&1\\
    \text{and }\Var{\Zijk}&=&\Var{\frac{\alpha}{\sqrt{1-\alpha^2}}\Yijk+\epsijk}\\
    &=&\Var{\frac{\alpha}{\sqrt{1-\alpha^2}}\Yijk}+2\Cov{\frac{\alpha}{\sqrt{1-\alpha^2}}\Yijk}{\epsijk}+\Var{\epsijk}\\
    &=&\frac{\alpha^2}{1-\alpha^2}\Var{\Yijk}+2\frac{\alpha}{\sqrt{1-\alpha^2}}\underbrace{\Cov{\Yijk}{\epsijk}}_{=0}+1\\
    &=&\frac{\alpha^2}{1-\alpha^2}\Var{\Yijk}+\frac{1-\alpha^2}{1-\alpha^2}\\
    &=&\frac{1}{1-\alpha^2}.\\
\end{eqnarray*}

Finally, the correlation is equal to:
\[\Corre{\Yijk}{\Zijk}=\frac{\Cov{\Yijk}{\Zijk}}{\sqrt{\Var{\Yijk}\times\Var{\Zijk}}}=\frac{\frac{\alpha}{\sqrt{1-\alpha^2}}}{\sqrt{1\times\frac{1}{1-\alpha^2}}}=\frac{\alpha}{\sqrt{1-\alpha^2}\times\frac{1}{\sqrt{1-\alpha^2}}}=\alpha.\]

\section{Estimation of $L$ in the $L$-nearest neighbors}

\begin{figure}[!ht]
    \centering
    \begin{tabular}{cc}
         \begin{minipage}{0.45\linewidth}
         \includegraphics[width=\linewidth]{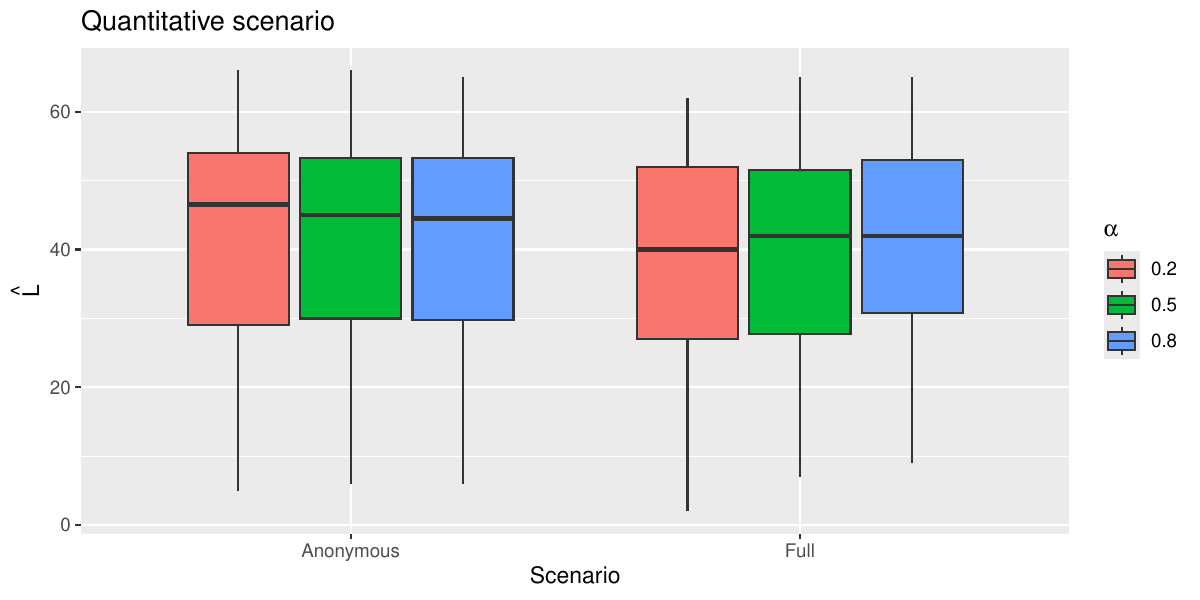}
         \end{minipage}&
         \begin{minipage}{0.45\linewidth}
         \includegraphics[width=\linewidth]{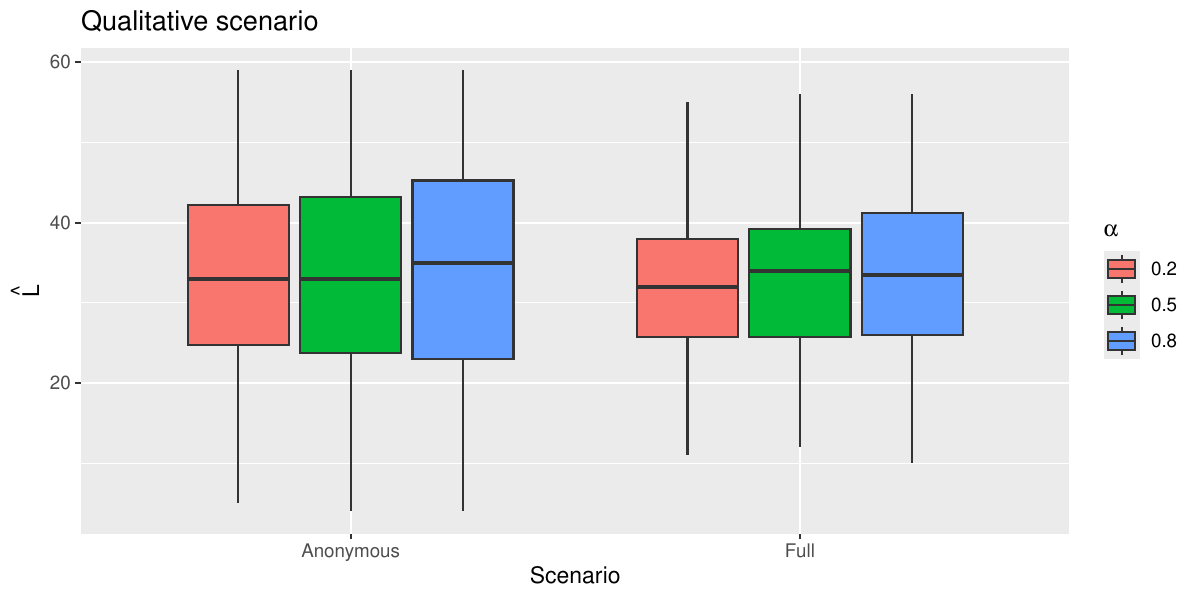}
         \end{minipage}  \\
    \end{tabular}
    \caption{Boxplot representation of the optimal $L$ chosen in the procedure according to the type of scenario (quantitative on the left and qualitative on the right), whether the file is anonymous or not (x-axis) and the value of $\alpha$ (in color).}
    \label{fig:appendix:knn}
\end{figure}

To select the range of $L$ values in the $L$ nearest neighbors procedure, the plan in section 4 has been adapted. To do this, the procedure\footnote{Implemented in the function \texttt{knn} (see \cite{venables2002modern})} was run using only perfect ranks, with a range between 1 and 5 for L at the start. The procedure is run on each matrix and, if the maximum value of the range is obtained, the range is enlarged and the procedure restarted. In the end, the optimal $L$ is retained (see figure~\ref{fig:appendix:knn}). For this experimental design, 100 matrices were simulated.

We can see from figure~\ref{fig:appendix:knn} that the range is for the both case approximately between 1 and 70.

\section{Estimation of the size and the parameter \texttt{decay} in multilayer perceptron}

\begin{figure}[!ht]
    \centering
    \begin{tabular}{ccc}
    &Quantitative&Qualitative\\
         \rotatebox{90}{\texttt{size}}&\begin{minipage}{0.45\linewidth}
         \includegraphics[width=\linewidth]{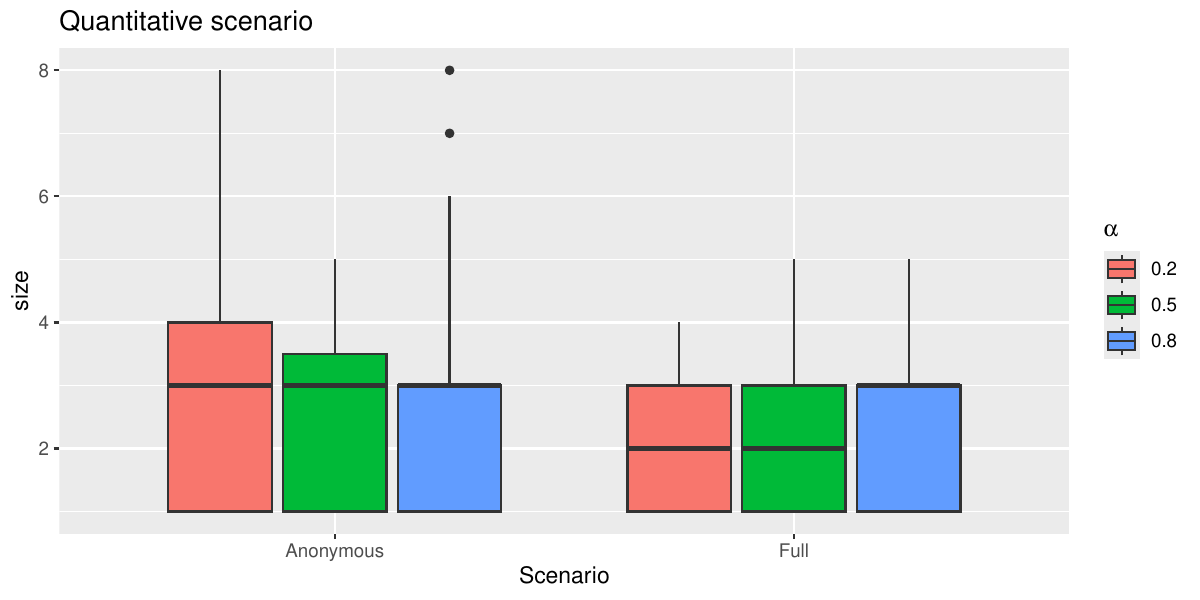}
         \end{minipage}&
         \begin{minipage}{0.45\linewidth}
         \includegraphics[width=\linewidth]{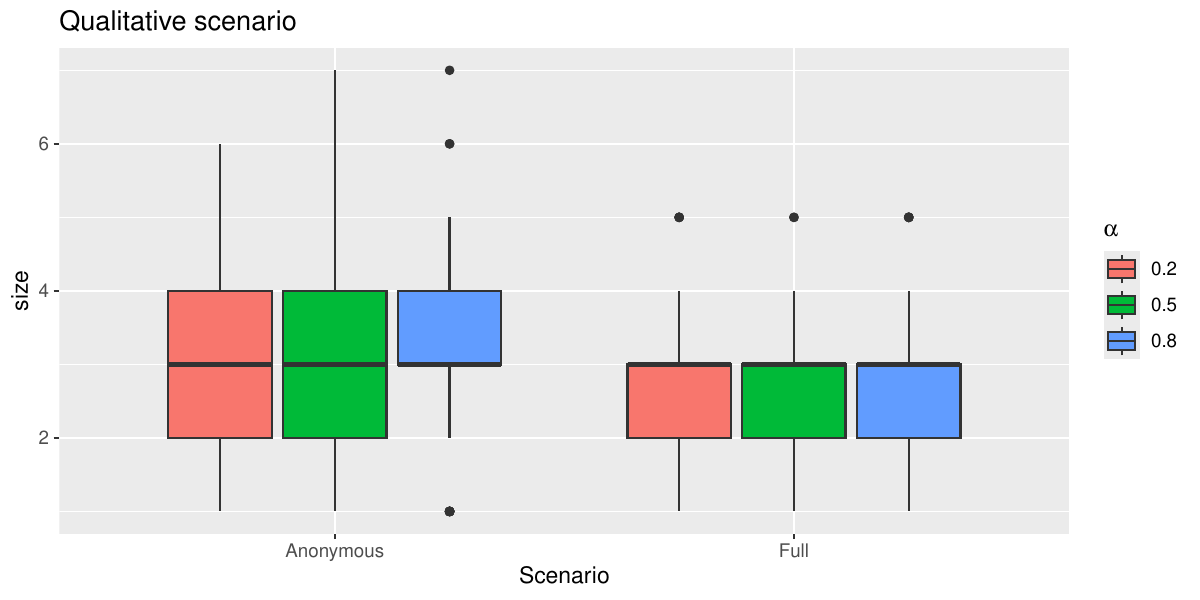}
         \end{minipage}  \\
         \rotatebox{90}{\texttt{decay}}&\begin{minipage}{0.45\linewidth}
         \includegraphics[width=\linewidth]{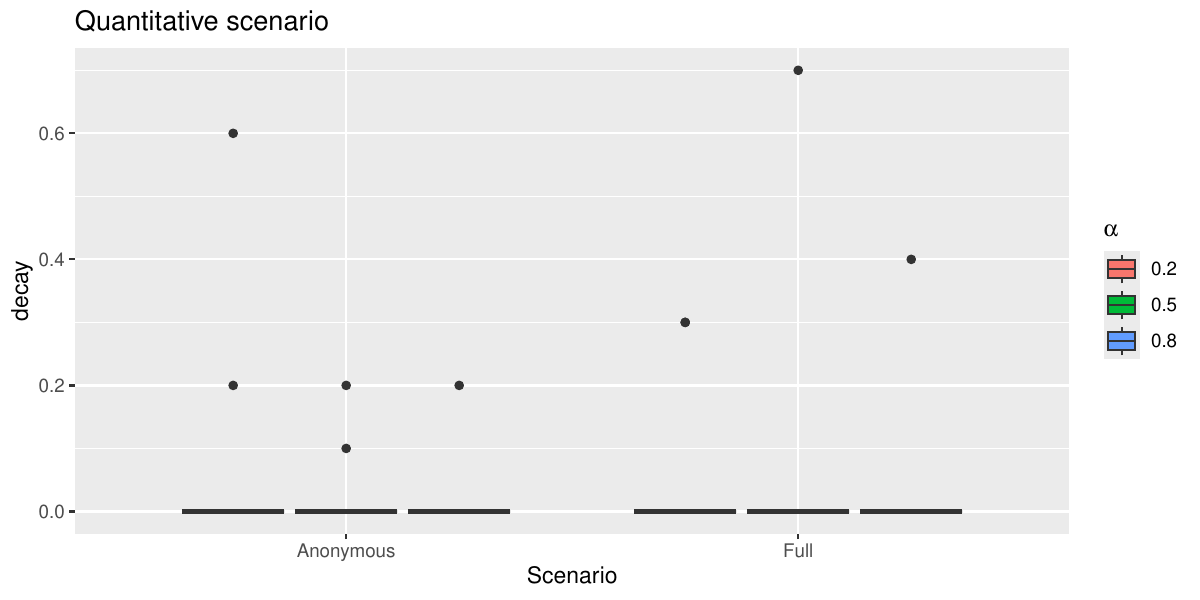}
         \end{minipage}&
         \begin{minipage}{0.45\linewidth}
         \includegraphics[width=\linewidth]{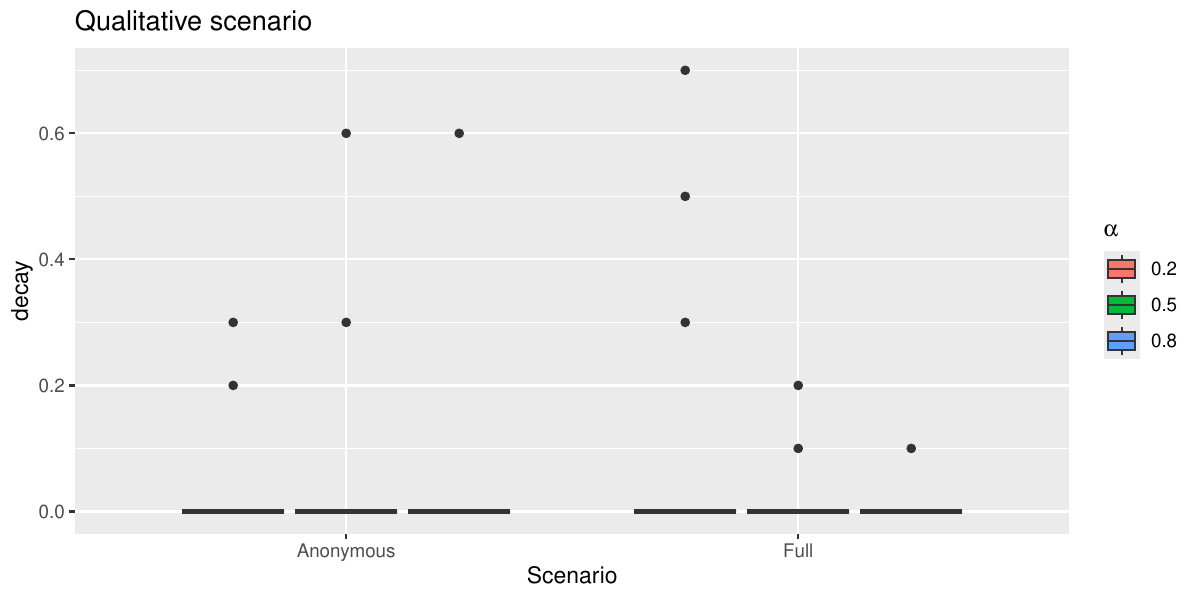}
         \end{minipage}  \\
    \end{tabular}
    \caption{Boxplot representation of the optimal number of nodes (\texttt{size} on the top) and the parameter \texttt{decay} (bottom) chosen in the procedure according to the type of scenario (quantitative on the left and qualitative on the right), whether the file is anonymous or not (x-axis) and the value of $\alpha$ (in color).}
    \label{fig:appendix:perceptron}
\end{figure}

To select the number of nodes and the penalty \texttt{decay}, we use the same procedure as for the k nearest neighbors starting with a number of nodes between 1 and 3 and \texttt{decay} in $\{0,0.1,\ldots,0.8\}$. The function used is \texttt{tune.nnet} from the \texttt{e1071} package (see \cite{meyer2023e1071}). The values obtained are shown in the figure~\ref{fig:appendix:perceptron}. Once again, 100 matrices were used.

We can see from the top of the figure~\ref{fig:appendix:perceptron} that the number of nodes is more or less the same whatever the scenario; between 1 and 4 for most of the time, with a maximum of 10. On the other hand, most of the \texttt{decay} parameters are set to 0.

To keep the time as short as possible while offering a variety of choices, we're going to propose a zero \texttt{decay} and nodes ranging from 1 to 10.

\section{Estimation of the kernel and the optimisation of this kernel in support vector machine}

As previously, we use the same procedure as for the k nearest neighbors and multilayer perceptron to chose the best kernel. For this, we compare the four kernels implemented in the function \texttt{tune.svm} from the \texttt{e1071} package (see \cite{meyer2023e1071}): \textit{linear}, \textit{polynomial}, \textit{radial} and \textit{sigmoïd}. For each simulation, we keep the kernel with the less error.

\section{All simulation results}

\begin{figure}[!ht]
    \centering
    \includegraphics[width=\linewidth]{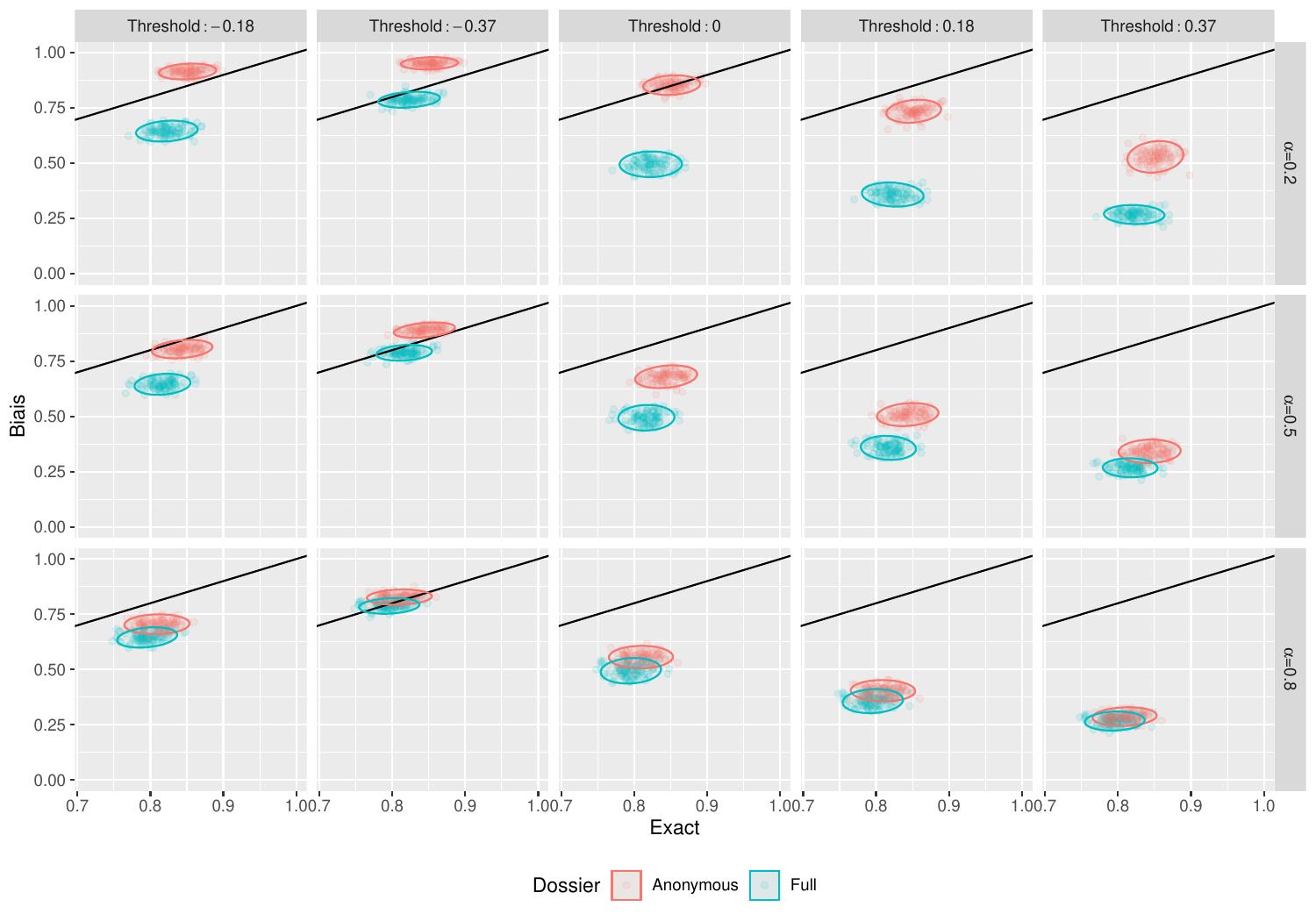}
    \caption{Representation of the $L$-nearest neighbors method results with the threshold scenario and binary $Y$, where each point has as its $x$-axis the average rate of correct classifications if the algorithm were to train on perfect classification, and as its $y$-axis the biased classification according to discrimination case (columns), $\alpha$ correlation (rows) and file type (in \textcolor{TURQUOISE}{turquoise} for complete files and \textcolor{SALMON}{saumon} for anonymized files). The black line represents $y=x$ and ellipses at 95\% have been added.}
    \label{fig:norm_Seuil_knn}
\end{figure}

In this section, all simulations are presented (Figures~\ref{fig:binom_Auto_Reg_Log}-\ref{fig:norm_Seuil_knn}). The table~\ref{tab:sumarise} displays all the figures for a global view and indicates the position of each enlarged figure.

\begin{table}[!ht]
    \centering
    \caption{Summary table of all figures by method (in rows) and scenario (in columns). For each figure, each point has as its $x$-axis the average rate of correct classifications if the algorithm were to train on perfect classification, and as its $y$-axis the biased classification according to discrimination case (columns), $\alpha$ correlation (rows) and file type (in \textcolor{TURQUOISE}{turquoise} for complete files and \textcolor{SALMON}{saumon} for anonymized files). The black line represents $y=x$ and ellipses at 95\% have been added.}
    \label{tab:sumarise}
    \begin{tabular}{c||c||c|c}
    \multicolumn{1}{c}{}&\multicolumn{3}{c}{Scenario}\\\cmidrule{2-4}
    &Self-censorship&Threshold and binary&Threshold and continuous\\\hline\hline
    &Figure~\ref{fig:binom_Auto_Reg_Log}&Figure~\ref{fig:binom_Seuil_Reg_Log}&Figure~\ref{fig:norm_Seuil_Reg_Log}\\
    \rotatebox{90}{Logistic regression}&\includegraphics[width=0.28\linewidth]{res_grille_RSS_zoom_nolegend_binom_Auto_Reg_Log.pdf}&\includegraphics[width=0.28\linewidth]{res_grille_RSS_zoom_nolegend_binom_Seuil_Reg_Log.pdf}&\includegraphics[width=0.28\linewidth]{res_grille_RSS_zoom_nolegend_norm_Seuil_Reg_Log.pdf}\\\hline
    &Figure~\ref{fig:binom_Auto_log_AIC}&Figure~\ref{fig:binom_Seuil_log_AIC}&Figure~\ref{fig:norm_Seuil_log_AIC}\\
    \rotatebox{90}{Logistic reg + AIC}&\includegraphics[width=0.28\linewidth]{res_grille_RSS_zoom_nolegend_binom_Auto_log_AIC.pdf}&\includegraphics[width=0.28\linewidth]{res_grille_RSS_zoom_nolegend_binom_Seuil_log_AIC.pdf}&\includegraphics[width=0.28\linewidth]{res_grille_RSS_zoom_nolegend_norm_Seuil_log_AIC.pdf}\\\hline
    &Figure~\ref{fig:binom_Auto_SVM}&Figure~\ref{fig:binom_Seuil_SVM}&Figure~\ref{fig:norm_Seuil_SVM}\\
    \rotatebox{90}{\textit{SVM} method}&\includegraphics[width=0.28\linewidth]{res_grille_RSS_zoom_nolegend_binom_Auto_SVM.pdf}&\includegraphics[width=0.28\linewidth]{res_grille_RSS_zoom_nolegend_binom_Seuil_SVM.pdf}&\includegraphics[width=0.28\linewidth]{res_grille_RSS_zoom_nolegend_norm_Seuil_SVM.pdf}\\\hline
    &Figure~\ref{fig:binom_Auto_Neural_Network}&Figure~\ref{fig:binom_Seuil_Neural_Network}&Figure~\ref{fig:norm_Seuil_Neural_Network}\\
    \rotatebox{90}{Multi-layer perceptron}&\includegraphics[width=0.28\linewidth]{res_grille_RSS_zoom_nolegend_binom_Auto_Neural_Network.pdf}&\includegraphics[width=0.28\linewidth]{res_grille_RSS_zoom_nolegend_binom_Seuil_Neural_Network.pdf}&\includegraphics[width=0.28\linewidth]{res_grille_RSS_zoom_nolegend_norm_Seuil_Neural_Network.pdf}\\\hline
    &Figure~\ref{fig:binom_Auto_knn}&Figure~\ref{fig:binom_Seuil_knn}&Figure~\ref{fig:norm_Seuil_knn}\\
    \rotatebox{90}{$L$-nearest neighbors}&\includegraphics[width=0.28\linewidth]{res_grille_RSS_zoom_nolegend_binom_Auto_knn.pdf}&\includegraphics[width=0.28\linewidth]{res_grille_RSS_zoom_nolegend_binom_Seuil_knn.pdf}&\includegraphics[width=0.28\linewidth]{res_grille_RSS_zoom_nolegend_norm_Seuil_knn.pdf}\\
    \end{tabular}
\end{table}

%%%%%%% Self-censorship

\begin{figure}[!ht]
    \centering
    \includegraphics[width=\linewidth]{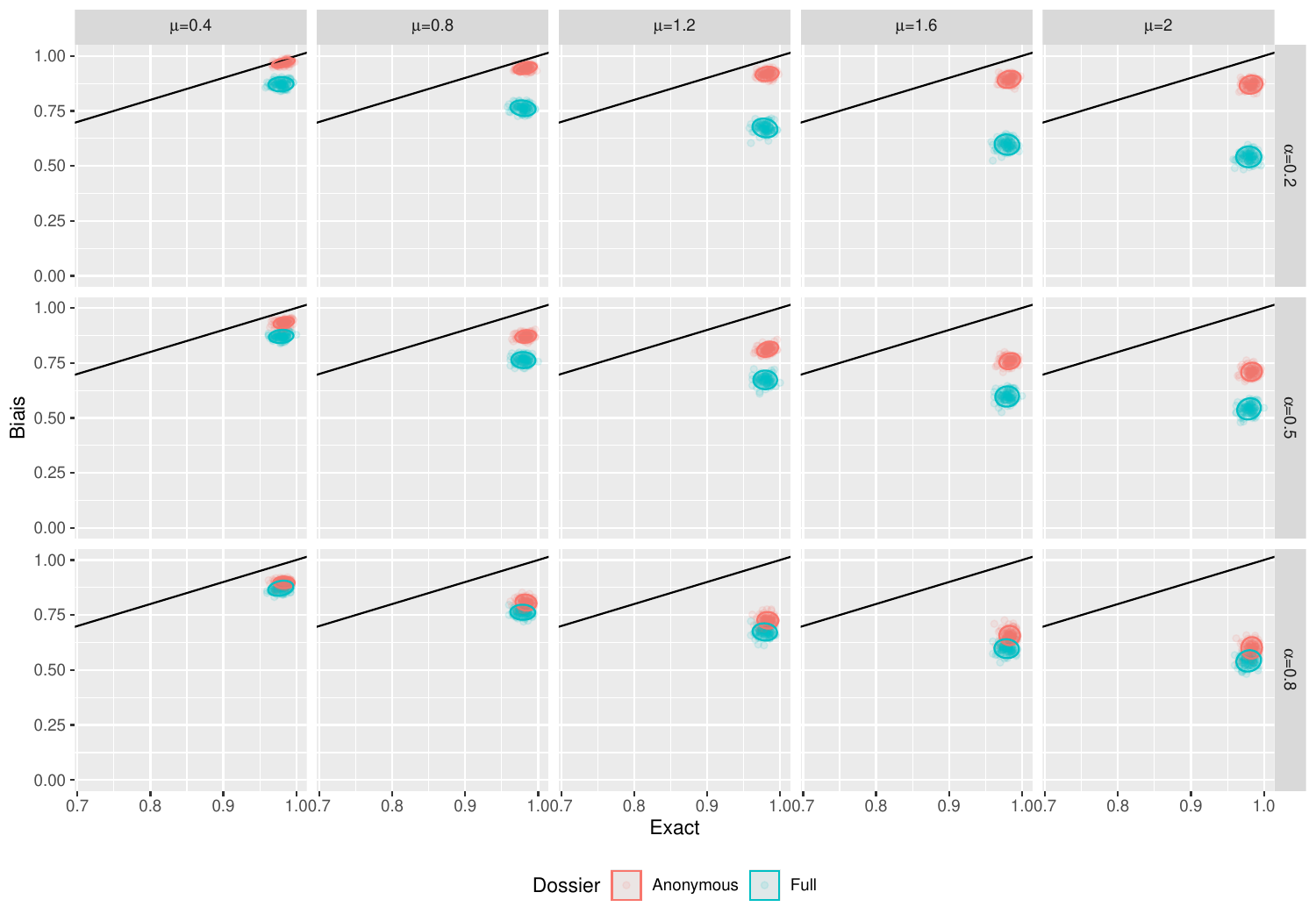}
    \caption{Representation of logistic regression results for the self-censorship scenario, where each point has as its $x$-axis the average rate of correct classifications if the algorithm were to train on perfect classification, and as its $y$-axis the biased classification according to discrimination case (columns), $\alpha$ correlation (rows) and file type (in \textcolor{TURQUOISE}{turquoise} for complete files and \textcolor{SALMON}{saumon} for anonymized files). The black line represents $y=x$ and ellipses at 95\% have been added.}
    \label{fig:binom_Auto_Reg_Log}
\end{figure}

\begin{figure}[!ht]
    \centering
    \includegraphics[width=\linewidth]{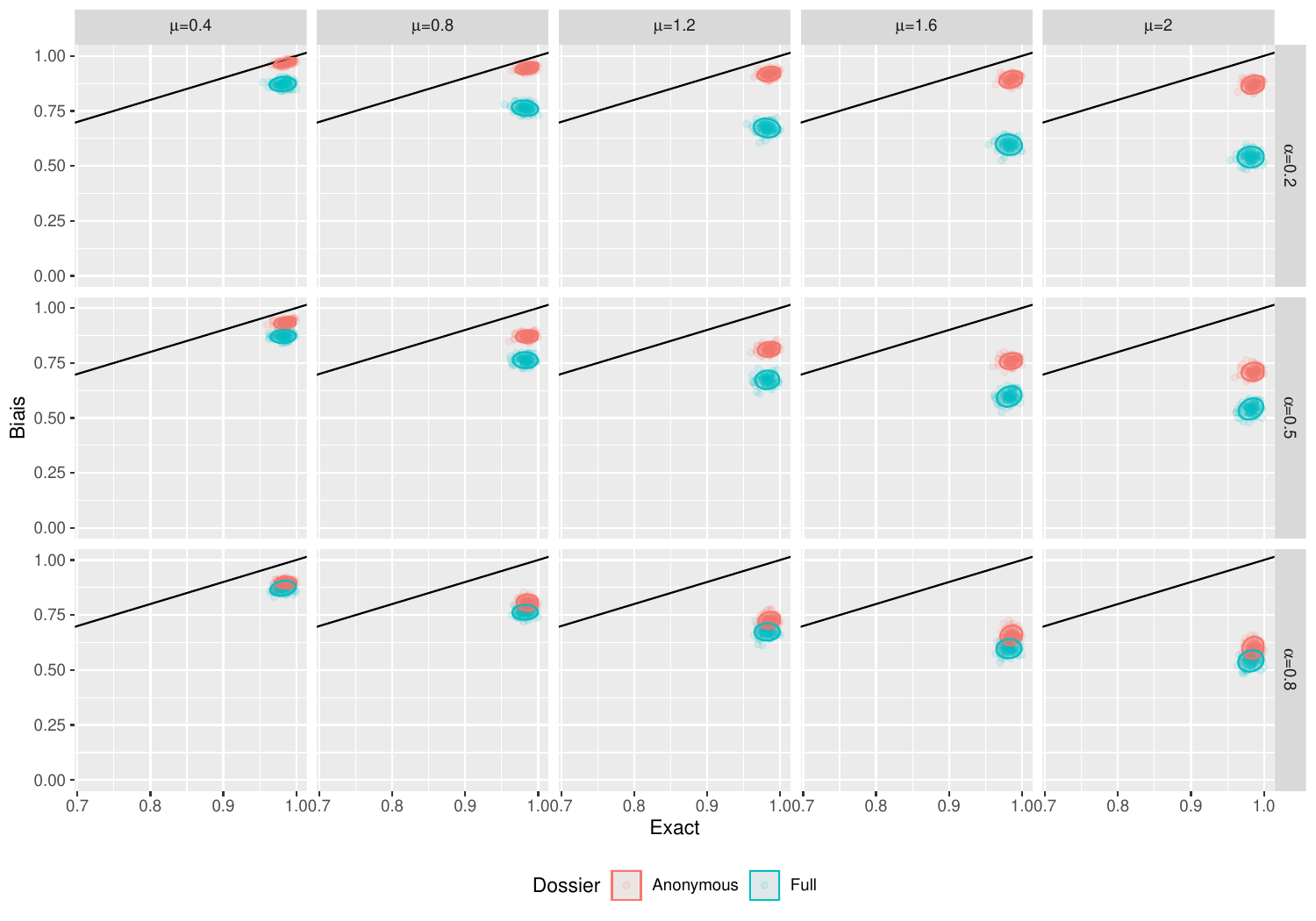}
    \caption{Representation of the logistic regression and AIC criterion results with the self-censorship scenario, where each point has as its $x$-axis the average rate of correct classifications if the algorithm were to train on perfect classification, and as its $y$-axis the biased classification according to discrimination case (columns), $\alpha$ correlation (rows) and file type (in \textcolor{TURQUOISE}{turquoise} for complete files and \textcolor{SALMON}{saumon} for anonymized files). The black line represents $y=x$ and ellipses at 95\% have been added.}
    \label{fig:binom_Auto_log_AIC}
\end{figure}

\begin{figure}[!ht]
    \centering
    \includegraphics[width=\linewidth]{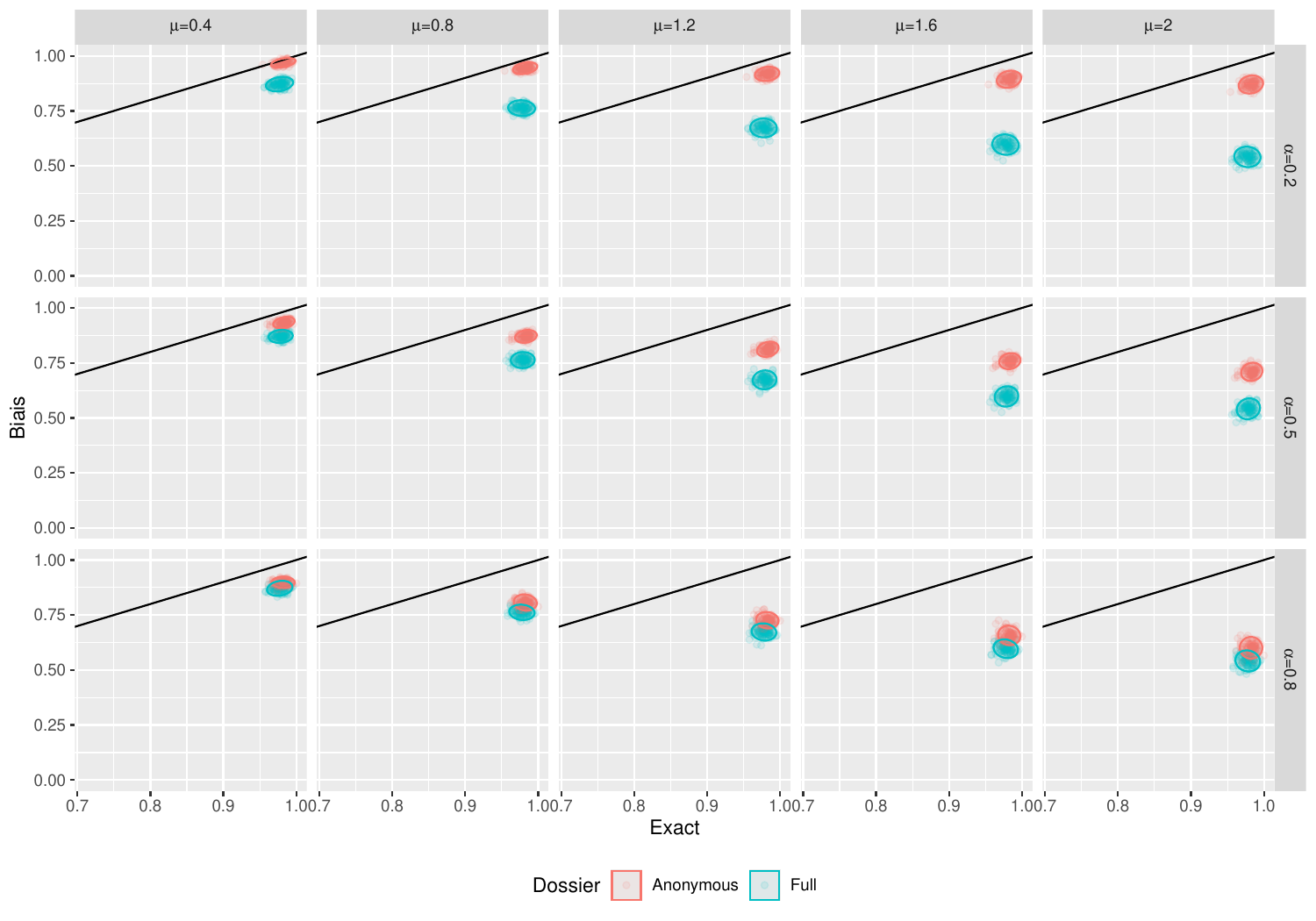}
    \caption{Representation of the SVM results for the self-censorship scenario, where each point has as its $x$-axis the average rate of correct classifications if the algorithm were to train on perfect classification, and as its $y$-axis the biased classification according to discrimination case (columns), $\alpha$ correlation (rows) and file type (in \textcolor{TURQUOISE}{turquoise} for complete files and \textcolor{SALMON}{saumon} for anonymized files). The black line represents $y=x$ and ellipses at 95\% have been added.}
    \label{fig:binom_Auto_SVM}
\end{figure}

\begin{figure}[!ht]
    \centering
    \includegraphics[width=\linewidth]{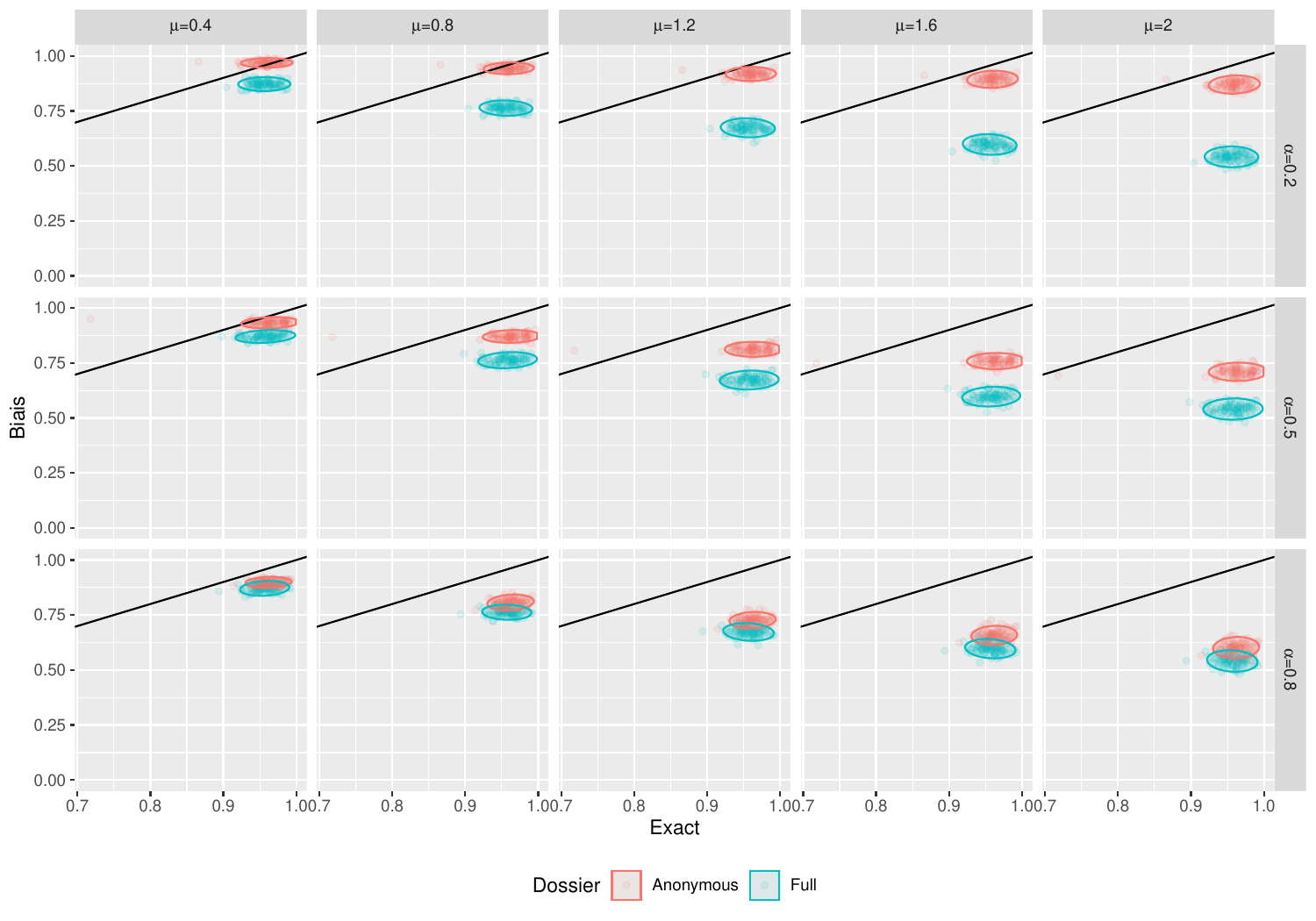}
    \caption{Representation of the multi-layer perceptron results for the self-censorship scenario, where each point has as its $x$-axis the average rate of correct classifications if the algorithm were to train on perfect classification, and as its $y$-axis the biased classification according to discrimination case (columns), $\alpha$ correlation (rows) and file type (in \textcolor{TURQUOISE}{turquoise} for complete files and \textcolor{SALMON}{saumon} for anonymized files). The black line represents $y=x$ and ellipses at 95\% have been added.}
    \label{fig:binom_Auto_Neural_Network}
\end{figure}

\begin{figure}[!ht]
    \centering
    \includegraphics[width=\linewidth]{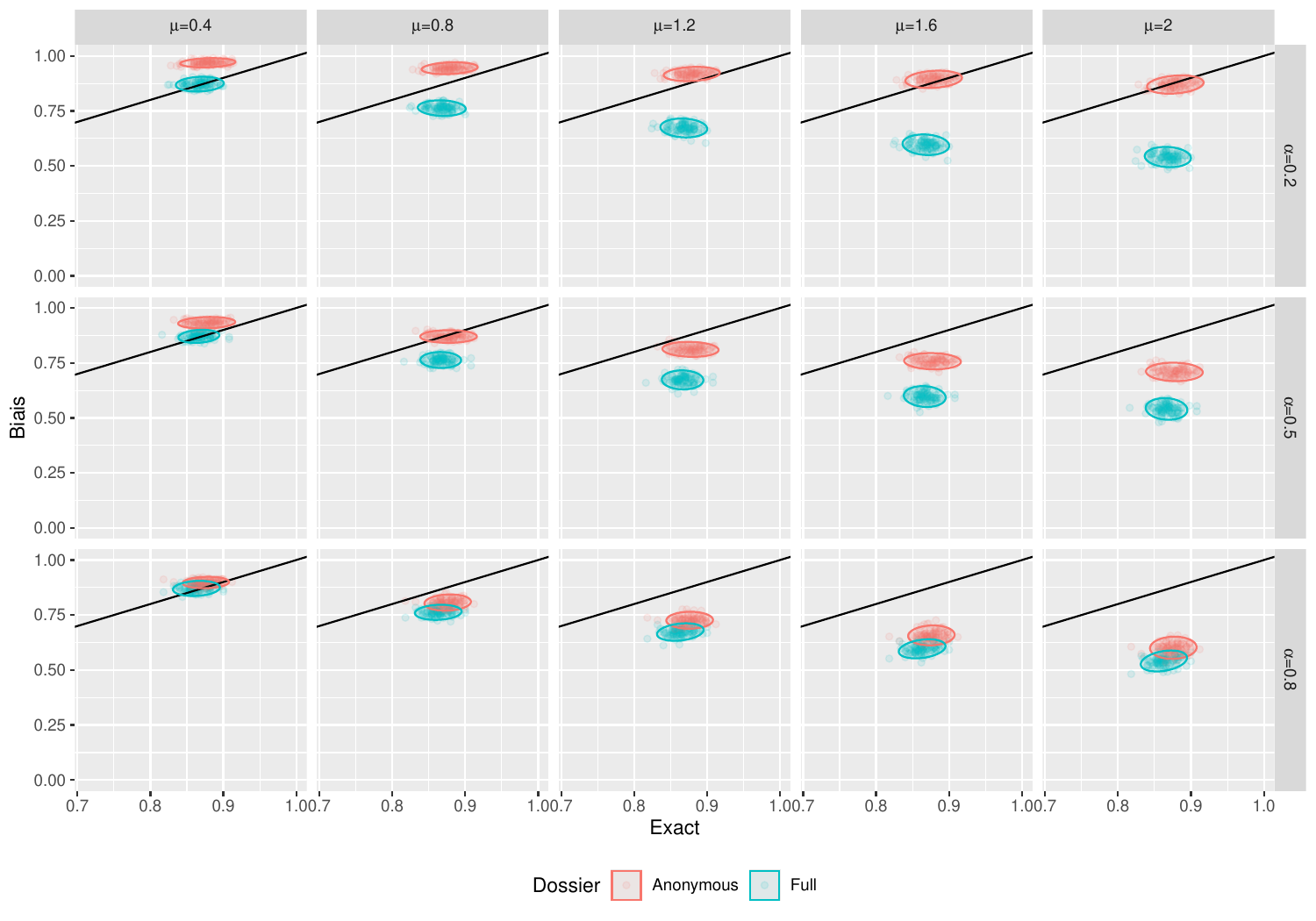}
    \caption{Representation of the $L$-nearest neighbors method results with the self-censorship scenario, where each point has as its $x$-axis the average rate of correct classifications if the algorithm were to train on perfect classification, and as its $y$-axis the biased classification according to discrimination case (columns), $\alpha$ correlation (rows) and file type (in \textcolor{TURQUOISE}{turquoise} for complete files and \textcolor{SALMON}{saumon} for anonymized files). The black line represents $y=x$ and ellipses at 95\% have been added.}
    \label{fig:binom_Auto_knn}
\end{figure}

%%%%%%% Binom threshold

\begin{figure}[!ht]
    \centering
    \includegraphics[width=\linewidth]{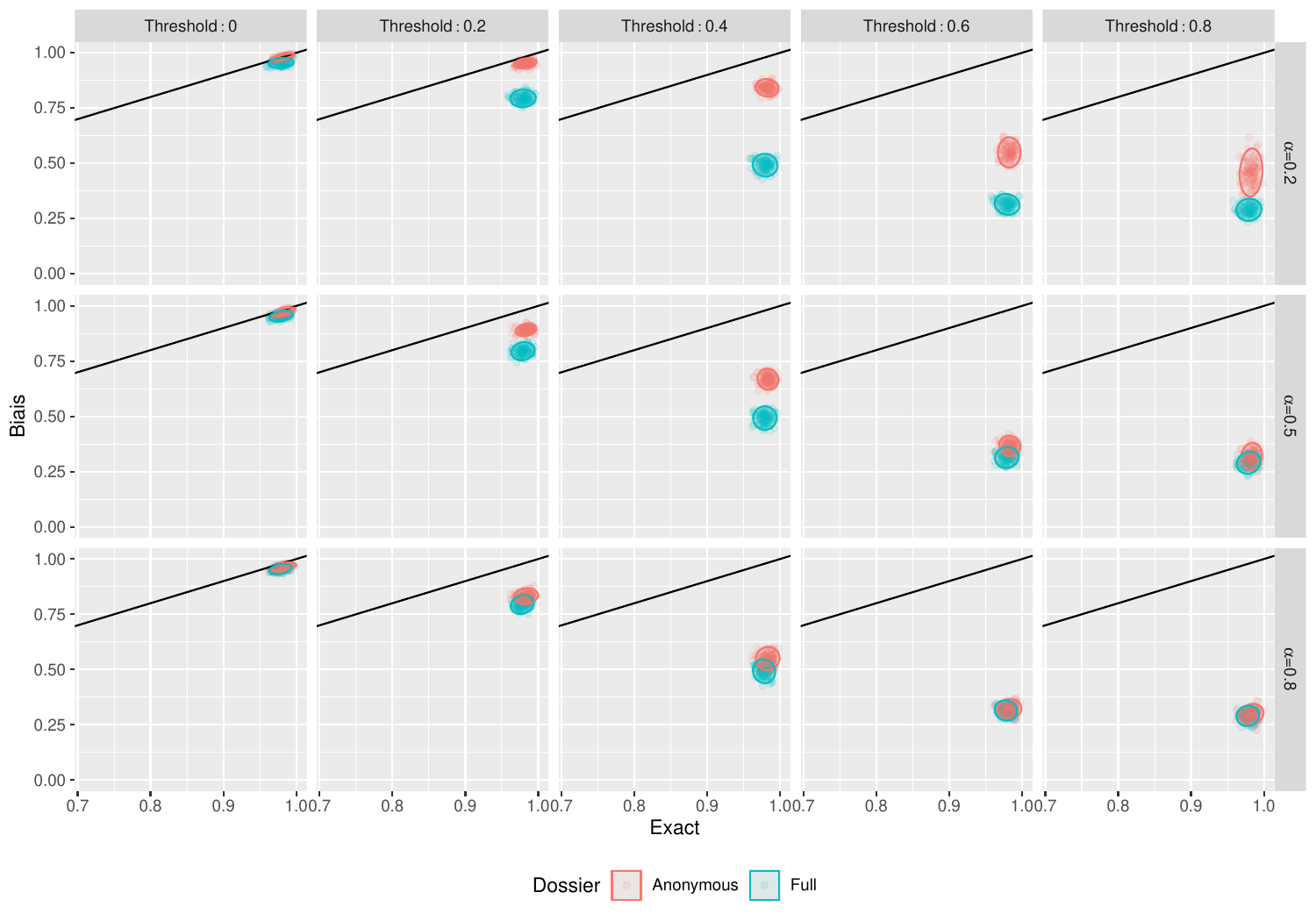}
    \caption{Representation of logistic regression results for the threshold scenario and binary $Y$, where each point has as its $x$-axis the average rate of correct classifications if the algorithm were to train on perfect classification, and as its $y$-axis the biased classification according to discrimination case (columns), $\alpha$ correlation (rows) and file type (in \textcolor{TURQUOISE}{turquoise} for complete files and \textcolor{SALMON}{saumon} for anonymized files). The black line represents $y=x$ and ellipses at 95\% have been added.}
    \label{fig:binom_Seuil_Reg_Log}
\end{figure}

\begin{figure}[!ht]
    \centering
    \includegraphics[width=\linewidth]{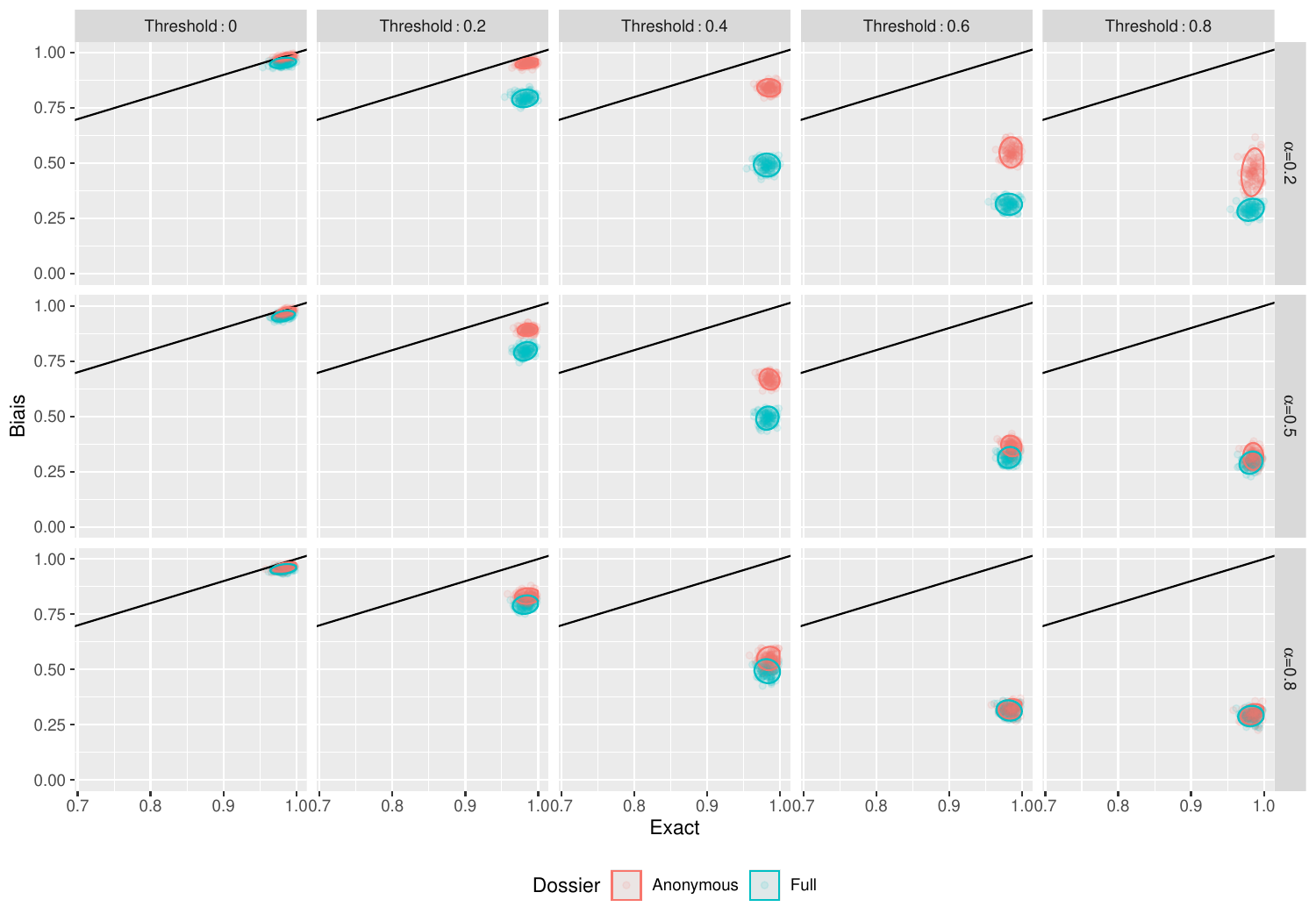}
    \caption{Representation of the logistic regression and AIC criterion results with the threshold scenario and binary $Y$, where each point has as its $x$-axis the average rate of correct classifications if the algorithm were to train on perfect classification, and as its $y$-axis the biased classification according to discrimination case (columns), $\alpha$ correlation (rows) and file type (in \textcolor{TURQUOISE}{turquoise} for complete files and \textcolor{SALMON}{saumon} for anonymized files). The black line represents $y=x$ and ellipses at 95\% have been added.}
    \label{fig:binom_Seuil_log_AIC}
\end{figure}

\begin{figure}[!ht]
    \centering
    \includegraphics[width=\linewidth]{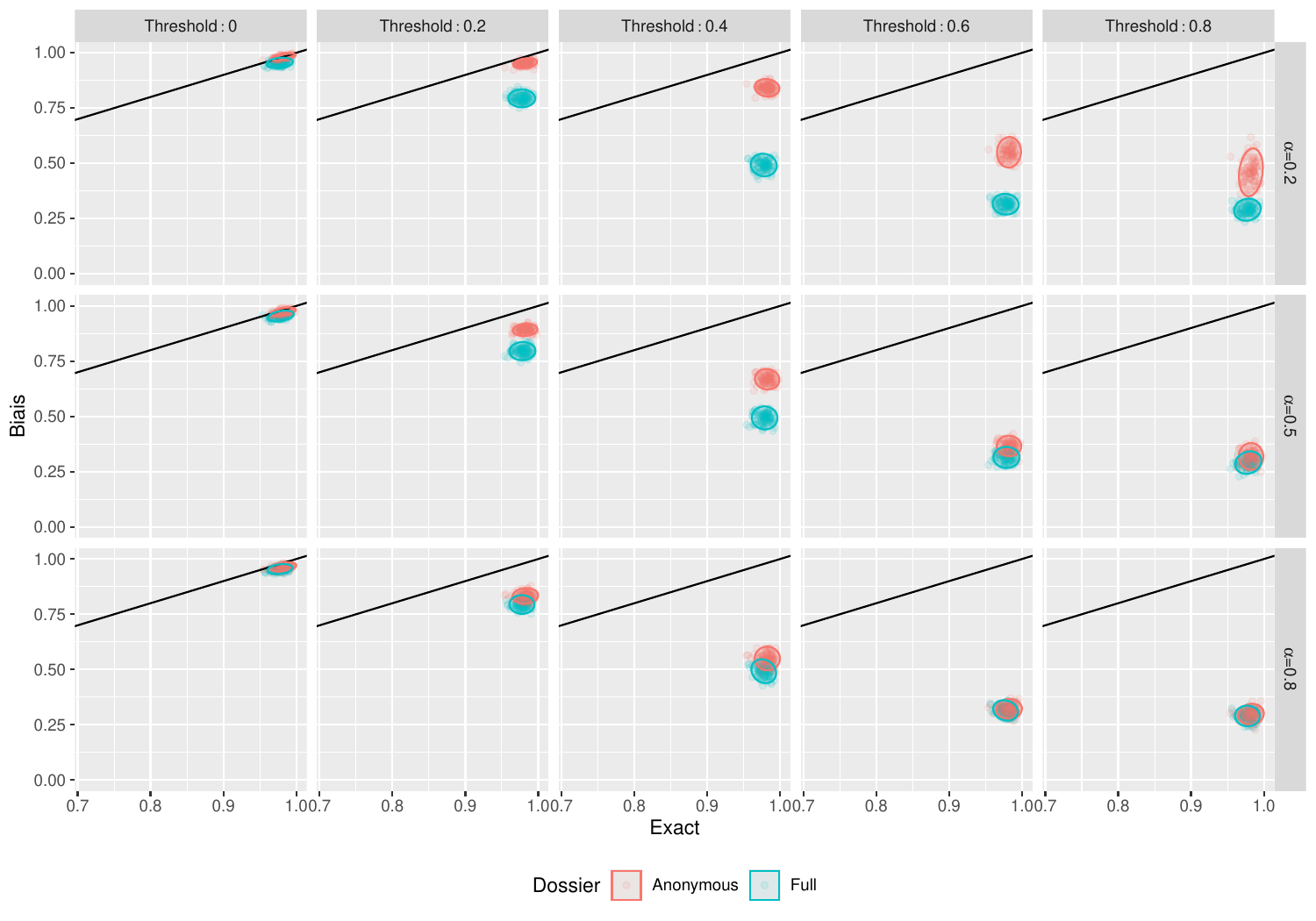}
    \caption{Representation of the SVM results for the threshold scenario and binary $Y$, where each point has as its $x$-axis the average rate of correct classifications if the algorithm were to train on perfect classification, and as its $y$-axis the biased classification according to discrimination case (columns), $\alpha$ correlation (rows) and file type (in \textcolor{TURQUOISE}{turquoise} for complete files and \textcolor{SALMON}{saumon} for anonymized files). The black line represents $y=x$ and ellipses at 95\% have been added.}
    \label{fig:binom_Seuil_SVM}
\end{figure}

\begin{figure}[!ht]
    \centering
    \includegraphics[width=\linewidth]{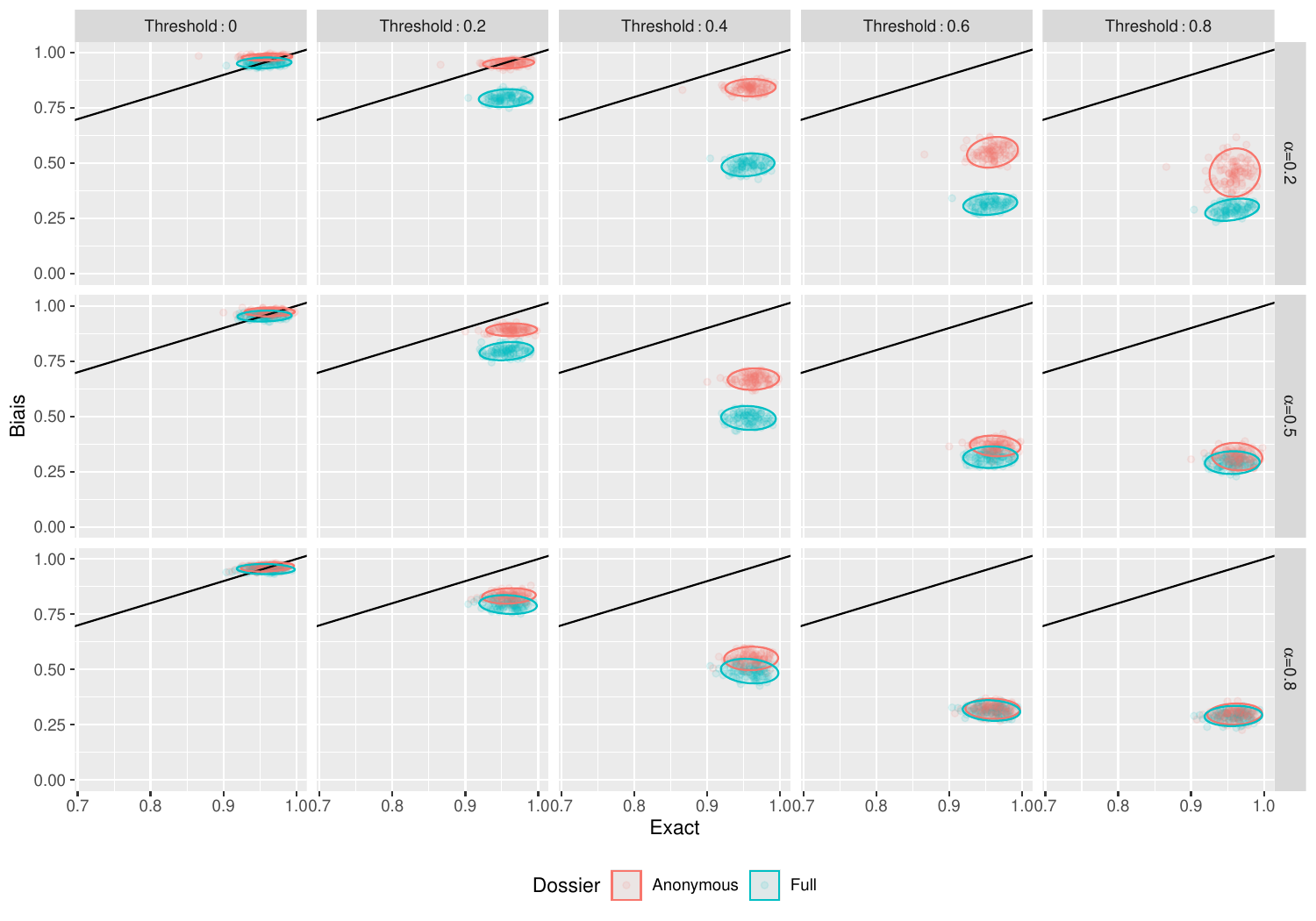}
    \caption{Representation of the multi-layer perceptron results for the threshold scenario and binary $Y$, where each point has as its $x$-axis the average rate of correct classifications if the algorithm were to train on perfect classification, and as its $y$-axis the biased classification according to discrimination case (columns), $\alpha$ correlation (rows) and file type (in \textcolor{TURQUOISE}{turquoise} for complete files and \textcolor{SALMON}{saumon} for anonymized files). The black line represents $y=x$ and ellipses at 95\% have been added.}
    \label{fig:binom_Seuil_Neural_Network}
\end{figure}

\begin{figure}[!ht]
    \centering
    \includegraphics[width=\linewidth]{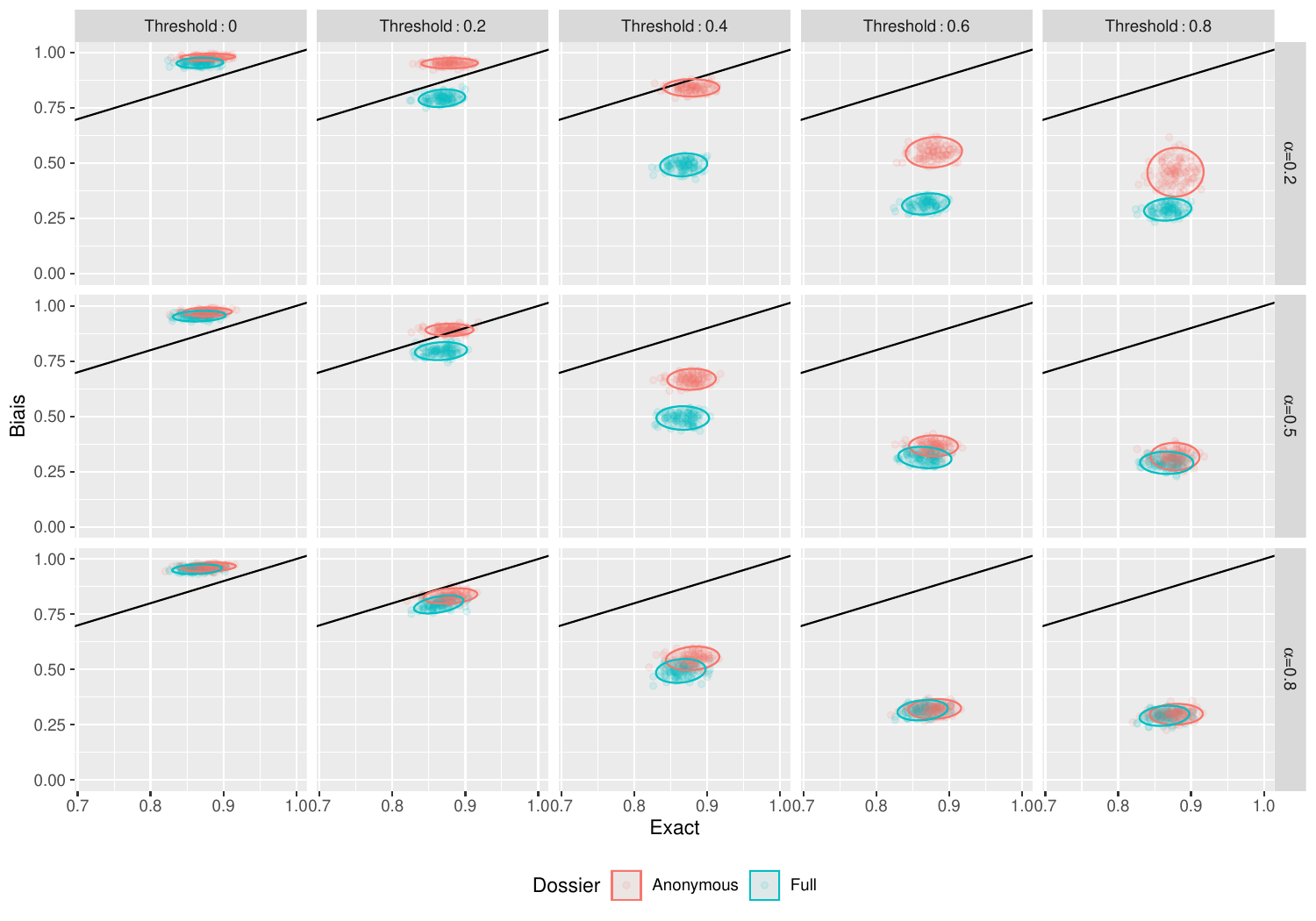}
    \caption{Representation of the $L$-nearest neighbors method results with the threshold scenario and binary $Y$, where each point has as its $x$-axis the average rate of correct classifications if the algorithm were to train on perfect classification, and as its $y$-axis the biased classification according to discrimination case (columns), $\alpha$ correlation (rows) and file type (in \textcolor{TURQUOISE}{turquoise} for complete files and \textcolor{SALMON}{saumon} for anonymized files). The black line represents $y=x$ and ellipses at 95\% have been added.}
    \label{fig:binom_Seuil_knn}
\end{figure}

%%%%%%% Continous threshold

\begin{figure}[!ht]
    \centering
    \includegraphics[width=\linewidth]{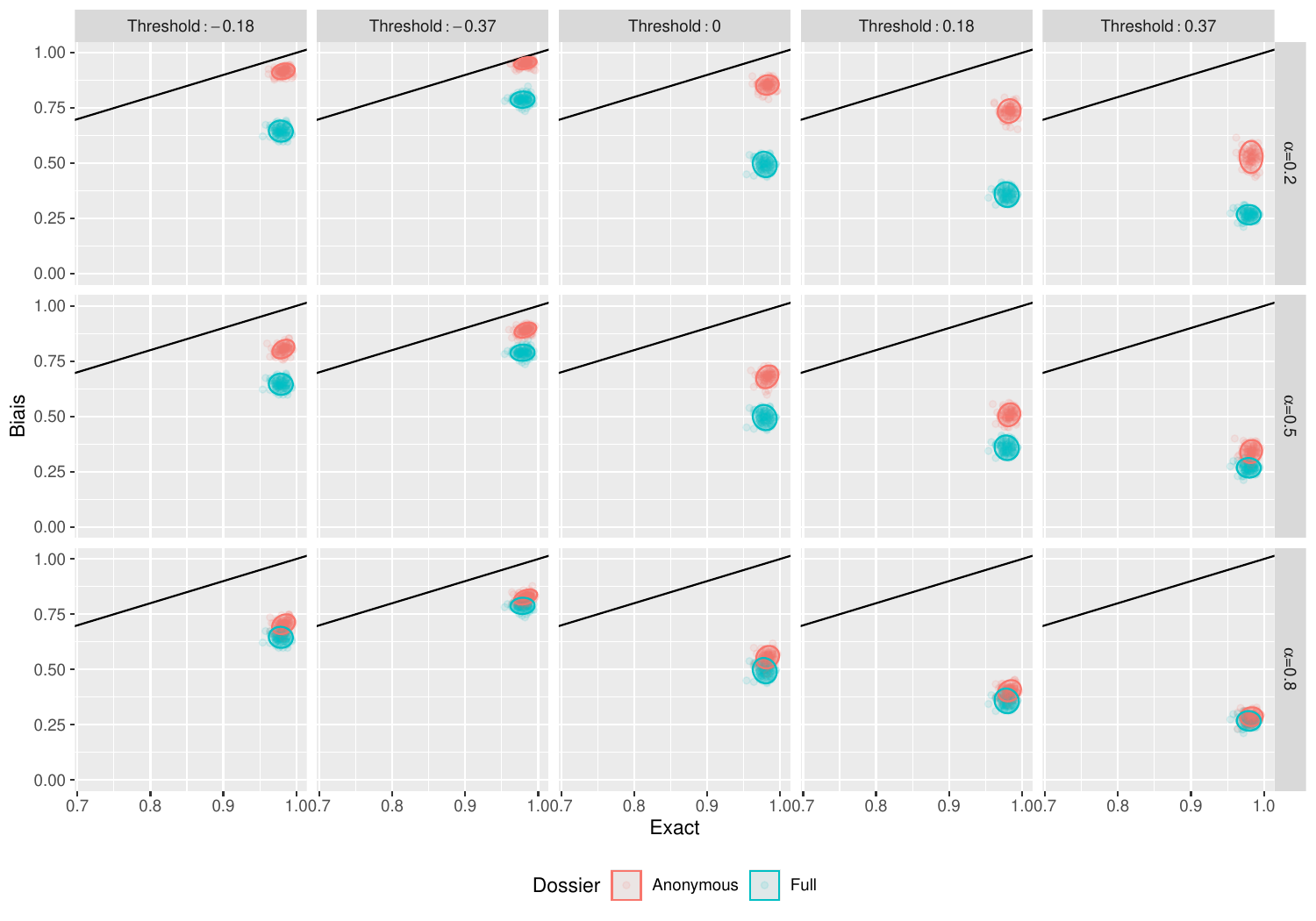}
    \caption{Representation of logistic regression results for the threshold scenario and binary $Y$, where each point has as its $x$-axis the average rate of correct classifications if the algorithm were to train on perfect classification, and as its $y$-axis the biased classification according to discrimination case (columns), $\alpha$ correlation (rows) and file type (in \textcolor{TURQUOISE}{turquoise} for complete files and \textcolor{SALMON}{saumon} for anonymized files). The black line represents $y=x$ and ellipses at 95\% have been added.}
    \label{fig:norm_Seuil_Reg_Log}
\end{figure}

\begin{figure}[!ht]
    \centering
    \includegraphics[width=\linewidth]{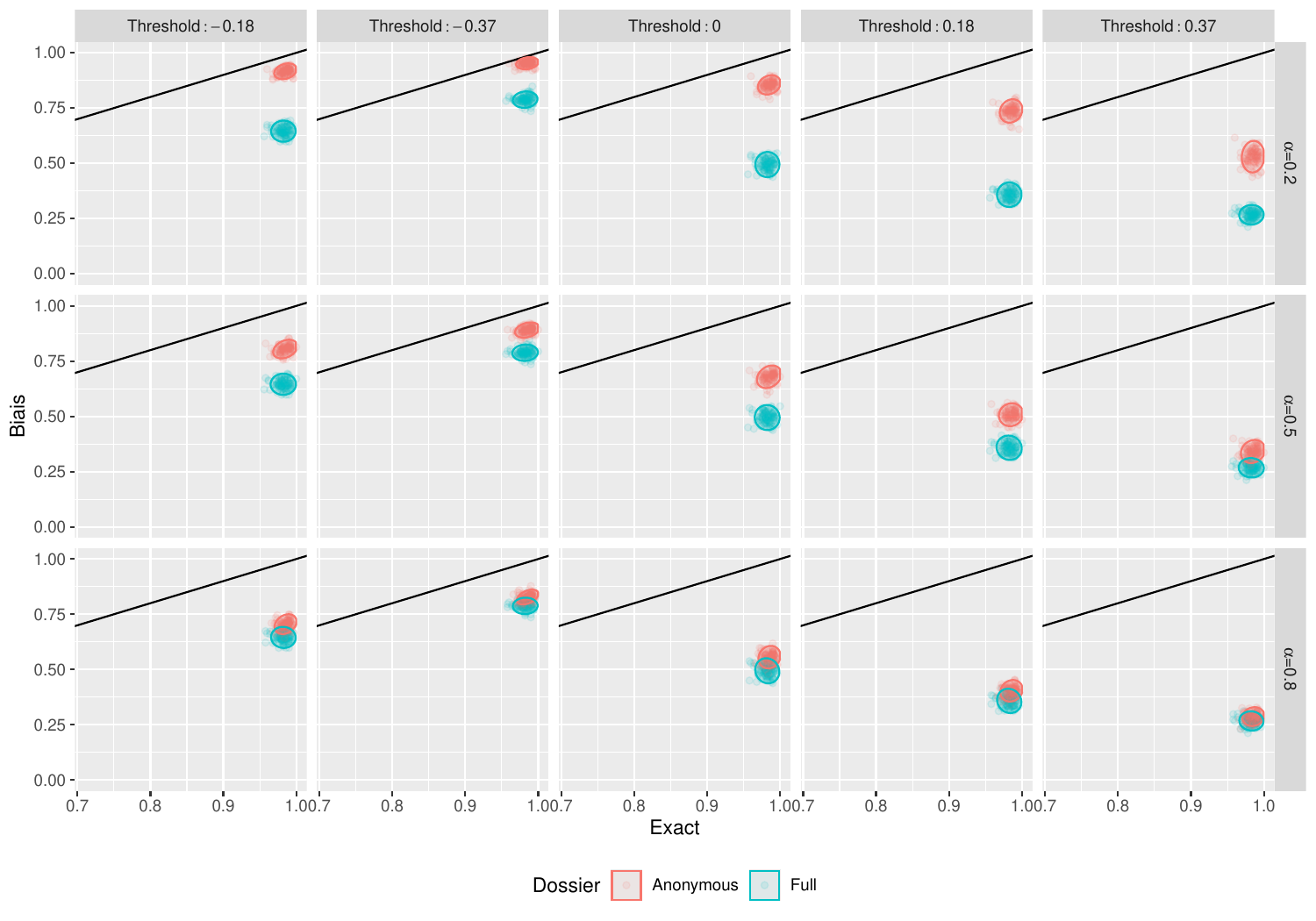}
    \caption{Representation of the logistic regression and AIC criterion results with the threshold scenario and binary $Y$, where each point has as its $x$-axis the average rate of correct classifications if the algorithm were to train on perfect classification, and as its $y$-axis the biased classification according to discrimination case (columns), $\alpha$ correlation (rows) and file type (in \textcolor{TURQUOISE}{turquoise} for complete files and \textcolor{SALMON}{saumon} for anonymized files). The black line represents $y=x$ and ellipses at 95\% have been added.}
    \label{fig:norm_Seuil_log_AIC}
\end{figure}

\begin{figure}[!ht]
    \centering
    \includegraphics[width=\linewidth]{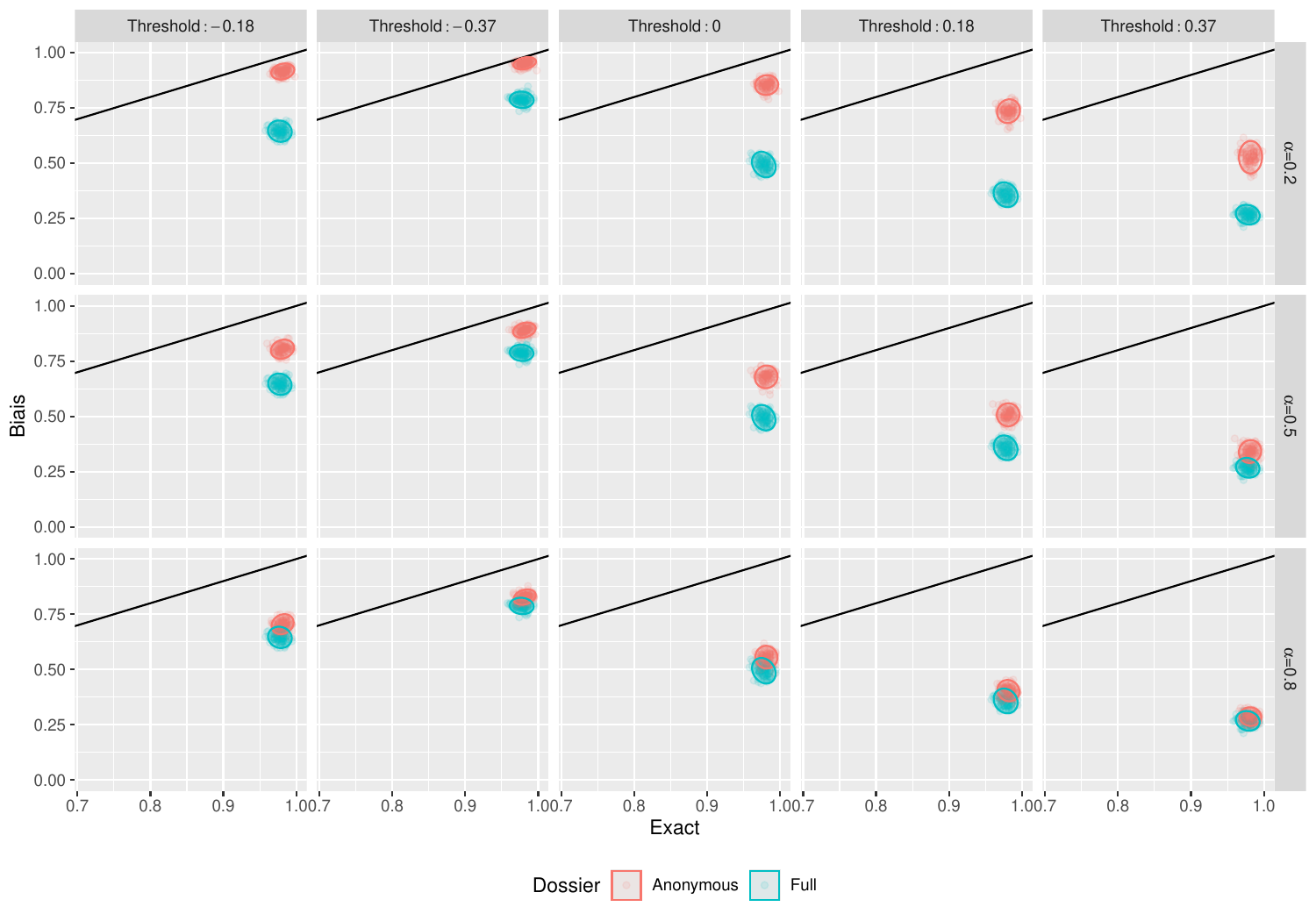}
    \caption{Representation of the SVM results for the threshold scenario and binary $Y$, where each point has as its $x$-axis the average rate of correct classifications if the algorithm were to train on perfect classification, and as its $y$-axis the biased classification according to discrimination case (columns), $\alpha$ correlation (rows) and file type (in \textcolor{TURQUOISE}{turquoise} for complete files and \textcolor{SALMON}{saumon} for anonymized files). The black line represents $y=x$ and ellipses at 95\% have been added.}
    \label{fig:norm_Seuil_SVM}
\end{figure}

\begin{figure}[!ht]
    \centering
    \includegraphics[width=\linewidth]{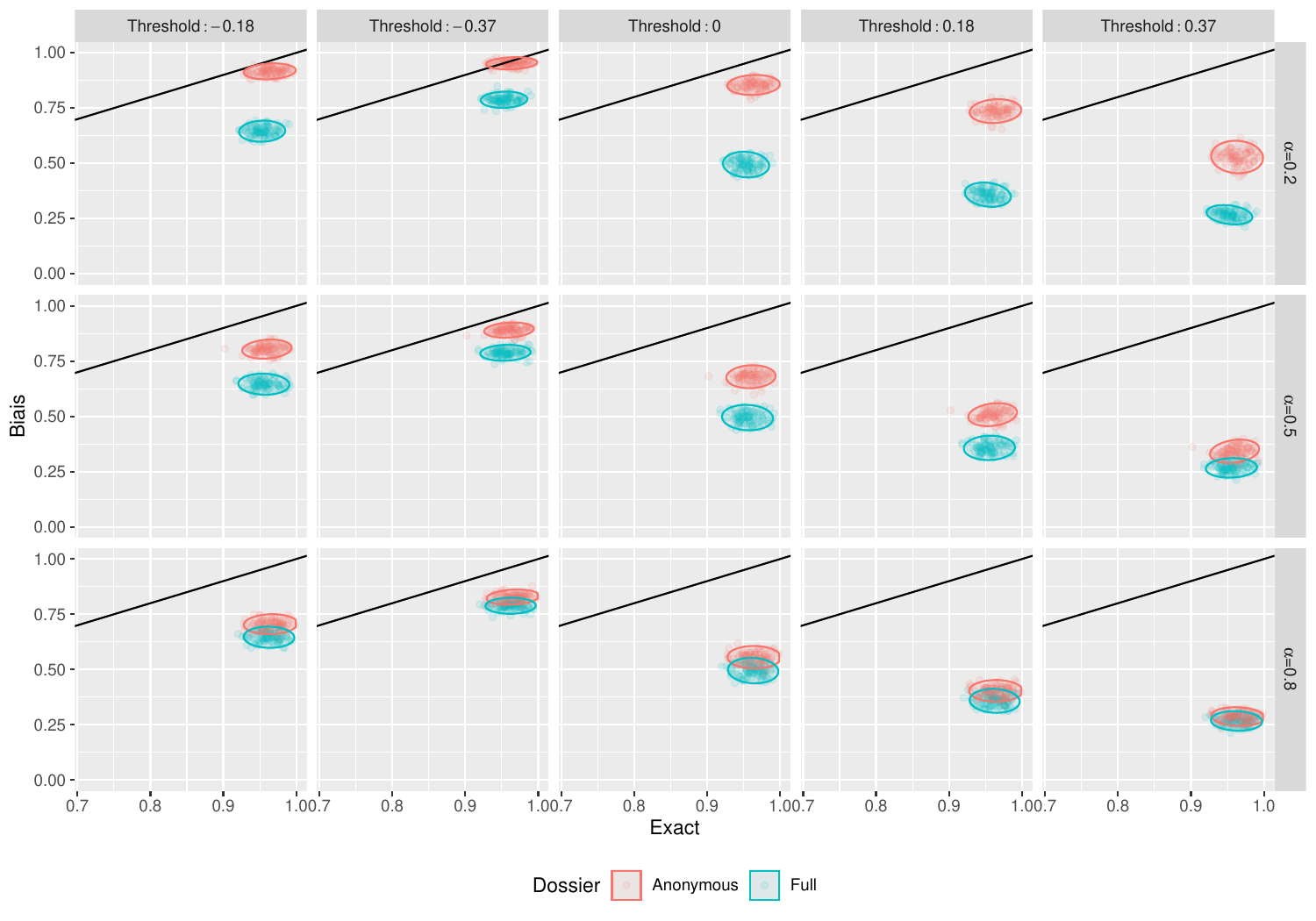}
    \caption{Representation of the multi-layer perceptron results for the threshold scenario and binary $Y$, where each point has as its $x$-axis the average rate of correct classifications if the algorithm were to train on perfect classification, and as its $y$-axis the biased classification according to discrimination case (columns), $\alpha$ correlation (rows) and file type (in \textcolor{TURQUOISE}{turquoise} for complete files and \textcolor{SALMON}{saumon} for anonymized files). The black line represents $y=x$ and ellipses at 95\% have been added.}
    \label{fig:norm_Seuil_Neural_Network}
\end{figure}

\end{appendices}

\FloatBarrier

\bibliographystyle{abbrvnat}
\bibliography{Biblio}% common bib file

\end{document}